\ifdefined\pdfobjcompresslevel\pdfobjcompresslevel=0\fi
\documentclass[]{uiuc_template}

\usepackage[utf8]{inputenc}
\usepackage[T1]{fontenc}
\usepackage{hyperref}
\usepackage{url}
\usepackage{amsfonts}
\usepackage{nicefrac}
\usepackage{amsmath}
\usepackage{amssymb}
\usepackage{mathtools}
\usepackage{natbib}
\usepackage{enumitem}
\usepackage{graphicx}
\usepackage{xcolor}
\usepackage{colortbl}
\usepackage{threeparttable}
\usepackage{amsmath}

\definecolor{airforceblue}{rgb}{0.36, 0.54, 0.66}
\definecolor{bluegray}{rgb}{0.4, 0.6, 0.8}
\definecolor{bleudefrance}{rgb}{0.19, 0.55, 0.91}
\hypersetup{colorlinks,linkcolor={airforceblue},citecolor={bleudefrance},urlcolor={bluegray}}

\definecolor{DeepBlue}{RGB}{36, 82, 168}
\definecolor{SoftRed}{RGB}{180, 60, 60}
\definecolor{DarkGreen}{RGB}{0, 120, 80}
\definecolor{WarmOrange}{RGB}{210, 120, 40}
\definecolor{MutedPurple}{RGB}{120, 90, 160}

\definecolor{DarkGray}{RGB}{80, 80, 80}
\definecolor{MidGray}{RGB}{120, 120, 120}
\definecolor{LightGray}{RGB}{180, 180, 180}
\definecolor{ShadowBlue}{RGB}{242, 245, 250}

\usepackage{xspace}
\usepackage{multirow}
\usepackage{listings}
\definecolor{bluegray}{rgb}{0.4,0.6,0.8}
\usepackage{wrapfig}
\usepackage{booktabs}
\usepackage{subcaption}
\DeclareCaptionFont{white}{\color{white}}
\DeclareCaptionFormat{listing}{\colorbox{bluegray}{\parbox{\textwidth}{#1#2#3}}}
\definecolor{no}{rgb}{0.851, 0.196, 0.196}
\definecolor{yes}{rgb}{0.271, 0.769, 0.690}
\definecolor{partial}{rgb}{0.95, 0.77, 0.36}

\definecolor{groupgray}{gray}{0.92}

\newcommand{\pmark}{\textcolor{partial}{\ding{51}\rotatebox[origin=c]{-6.2}{\kern-0.7em\ding{55}}}}
\usepackage[capitalize]{cleveref}
\crefname{figure}{Fig.}{Figs.}
\Crefname{figure}{Fig.}{Figs.}
\crefname{section}{Sec.}{Secs.}
\Crefname{section}{Sec.}{Secs.}
\crefname{table}{Table}{Tables}
\Crefname{table}{Table}{Tables}

\definecolor{affilblue}{RGB}{90, 170, 230}
\newcommand{\affilnum}[1]{\textsuperscript{\textcolor{affilblue}{#1}}}
\fancypagestyle{firststyle}{
    \fancyhead{}
    \fancyhead[L]{
        \vskip 4mm
    \includegraphics[height=34pt]{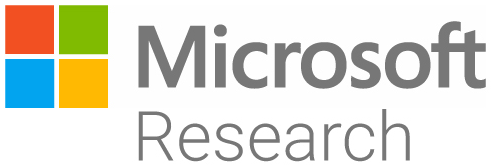}
    \hspace{4.5mm}
    \includegraphics[height=28.5pt]{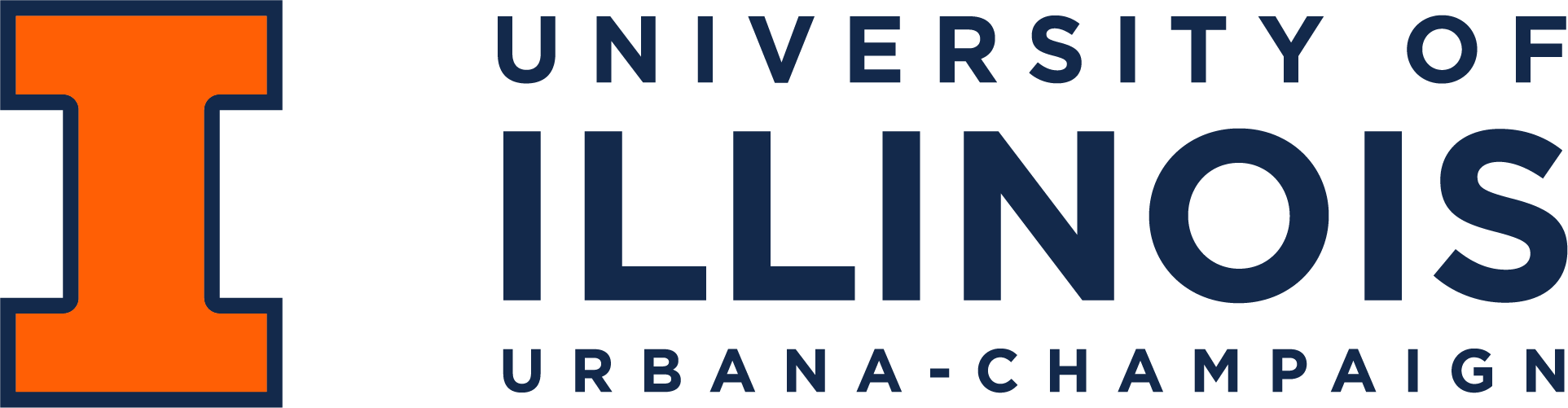}
    }
    \fancyfoot{}
    \fancyfoot[L]{%
        \footnotesize
        \begin{minipage}{0.62\textwidth}
        \rule{12pc}{0.4pt}\\[-0.2ex]
        \textsuperscript{*}Work done during an internship at Microsoft Research.
        \end{minipage}
    }
}

\title{StudentSim: Training LLM-based Student Simulators}

\author[]{%
\normalsize
Ke Yang\affilnum{1,2}\textsuperscript{*} \quad
Chenglong Wang\affilnum{1} \quad
Michel Galley\affilnum{1} \quad
Chandan Singh\affilnum{1} \quad
Jeevana Priya Inala\affilnum{1}

\vspace{0.12in}

ChengXiang Zhai\affilnum{2} \quad
Jianfeng Gao\affilnum{1}
}

\affiliation{
\normalsize
\affilnum{1}Microsoft Research
\quad
\affilnum{2}University of Illinois Urbana-Champaign
}

\abstract{AI tutors are most useful when they adaptively respond to each student's strengths, weaknesses, and preferred kinds of guidance, but which guidance works for which student is a sparse signal, slow and costly to collect from real students. 
\emph{Student simulators} can supply that signal as a proxy, yet existing ones cover only part of what this requires: state-tracking models fit how a student behaves but cannot digest a tutor's explanations or corrections well, while LLMs prompted to role-play a target student follow a tutor's guidance fluently but do not reliably reproduce the competence of the student they imitate. 
We present \textsc{StudentSim}, a training framework that turns sparse per-student data into an individualized simulator for each student through a two-stage pipeline of pooled training followed by per-student specialization, so that the simulator both mirrors the student's own responses and updates them under tutor guidance.
To measure these two abilities fairly, we build \textsc{StudentSimEval}, a standardized protocol spanning 60 students across chess, second-language English writing, and mathematics, drawn from public learner datasets whose de-identified student records are shared for research.
It scores every method on \emph{behavioral fidelity} ($\mathcal{F}\uparrow$), how well a simulator matches a student's own responses, and \emph{guidance responsiveness} ($\mathcal{R}\uparrow$), how readily it updates its response under a tutor's guidance, fitting each method on the same records and scoring it on the same held-out records so results are directly comparable; we release our construction and evaluation code so others can score new methods on the same benchmark and extend it. 
Across all three domains, our per-student simulators outperform GPT-5.4 on both metrics. 
In chess, for example, \textsc{StudentSim} reaches $\mathcal{F}=0.51$ and $\mathcal{R}=0.91$, compared with $0.23$ and $0.72$ for GPT-5.4 and $0.45$ and $0.27$ for Maia2, a skill-conditioned chess move prediction model. 
As a proof of concept that the framework also supports AI tutor improvement, a trained \textsc{StudentSim} used as the reward for tutor model reinforcement learning yields a chess tutor that expert humans rate as more accurate, better-guided, and more personalized than both a no-RL baseline and a tutor RL-trained against a GPT-5.4 simulator reward.
Our code is available at \href{https://github.com/microsoft/StudentSim}{https://github.com/microsoft/StudentSim}.
}

\begin{document}

\maketitle

\section{Introduction}
\label{sec:intro}

An AI assistant that helps human students learn (e.g., a math tutor, a foreign-language coach, a chess trainer) is more useful when it is trained on real-world interactions across a diverse student population: it must see varied backgrounds, baseline skills, and learning styles to understand the typical proficiency and common error patterns of students in that subject, and it must collect feedback on each individual's responses to different teaching materials to learn which teaching approach is most effective for a student with particular traits and learning habits. Recruiting and training a tutor against such a population, however, is prohibitively expensive and time-consuming. Therefore, the evaluation and improvement of adaptive AI tutors lag behind the rapid advancement of the underlying AI models themselves.

\begin{wrapfigure}{r}{0.6\linewidth}
  \vspace{-1.2\baselineskip}
  \centering
  \includegraphics[width=\linewidth]{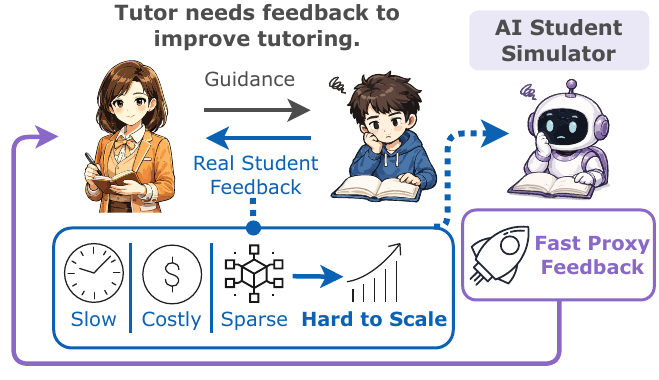}
  \caption{\textbf{Real-student feedback is hard to scale; a student simulator supplies it at machine timescales.}}
  \label{fig:motivation}
  \vspace{-0.6\baselineskip}
\end{wrapfigure}

\begin{figure}[!b]
  \centering
  \begin{minipage}[b]{0.563\linewidth}\centering
    \includegraphics[width=\linewidth]{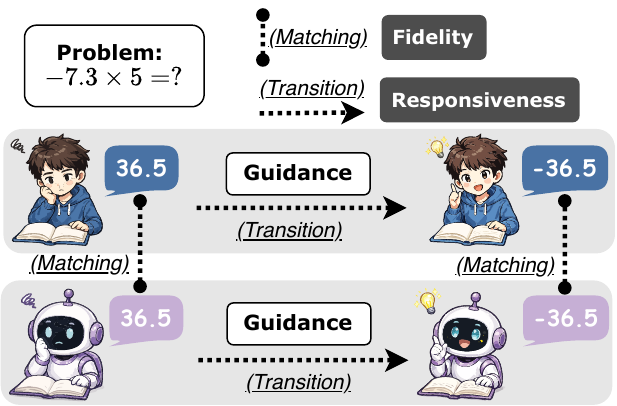}\\[3pt]
    \footnotesize (a) What a simulator is scored on
  \end{minipage}\hfill
  \begin{minipage}[b]{0.427\linewidth}\centering
    \includegraphics[width=\linewidth]{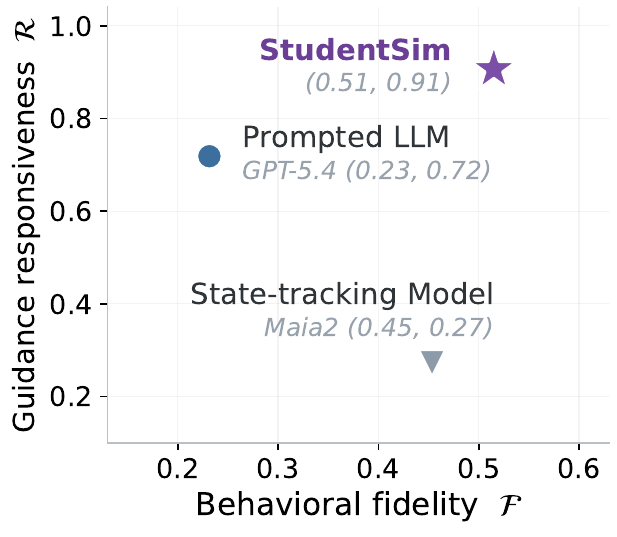}\\[3pt]
    \footnotesize (b) Where each method lands, on chess
  \end{minipage}
  \caption{\textbf{Scoring a student simulator.} Education tracks a learner's progress through what they produce independently and how far they advance once support arrives, the zone of proximal development \citep{VYGOTSKY_1980} that dynamic assessment operationalizes \citep{Grigorenko_1998}. We score whether a student simulator can serve as a feedback source for a tutor, which rests on two abilities mirroring those signals: \emph{behavioral fidelity}, behaving like the student it stands for, and \emph{guidance responsiveness}, moving where the tutor's guidance indicates (a). On chess (b), \textsc{StudentSim} leads on both, the prompted LLM on responsiveness alone, and the state-tracking model on fidelity alone, and the same ranking holds in L2 and math (\cref{tab:fidelity_main} and~\cref{tab:guidance_main}).}
  \label{fig:two_axes}
\end{figure}

One alternative is to use \emph{student simulators} that produce proxy feedback at machine timescales (\cref{fig:motivation}). Educational measurement assesses a learner's improvement along two targets: what the learner produces independently, and how far the learner progresses with support, which is the zone of proximal development \citep{VYGOTSKY_1980} that dynamic assessment operationalizes \citep{Grigorenko_1998}. Motivated by these, a useful simulator must satisfy two requirements (\cref{fig:two_axes}). First, \textbf{behavioral fidelity}: how well the simulator reproduces a student's own response to a problem, including that student's characteristic mistakes and strengths. Second, \textbf{guidance responsiveness}: given a problem, the student's initial response, and a tutor's guidance addressing it, how readily the simulator updates its response in the direction the guidance is steering. These two axes are the two properties tutor training needs from a simulated student, i.e., a personalized starting point and the capacity to be taught. A pool of such simulators then stands in for a diverse population of teachable students for AI tutor training.

Existing student simulator approaches usually satisfy only one of these two requirements. \emph{State-tracking student models}, including knowledge-tracing approaches \citep{Corbett_1995,piech2015deepknowledgetracing,ghosh2020contextawareattentiveknowledgetracing} and human-style behavior prediction models \citep{mcilroyyoung2020aligningsuperhumanaihuman,tang2024maia2unifiedmodelhumanai}, are built from real human data and capture aspects of behavioral fidelity, but they have no input pathway for tutor explanations or corrections. Conversely, \emph{LLM-prompted user simulators} \citep{owoicho2023exploitingsimulateduserfeedback,terragni2023incontextlearningusersimulators} respond to tutor guidance fluently, but even given a precise textual description of the student's cognitive state, the LLM does not reliably produce behavior consistent with that state. Therefore, the central challenge is to build per-student simulators that are jointly faithful to individual student competence (e.g., their strengths and weaknesses) and responsive to tutor guidance.

We address this gap with \textsc{StudentSim}, a training framework that builds an individualized simulator for each student even though any one student contributes only sparse records. It works in two stages (\cref{fig:training_pipeline}): first it pools records across all students to pretrain a base simulator that captures what is shared in the domain, such as common mistakes and how students revise their responses after guidance; then it specializes that base on a single student's own records, producing a simulator that mirrors that student's behavior and updates the way that student would after tutor guidance. To evaluate such simulators on the same footing, we build \textsc{StudentSimEval}, a standardized protocol spanning 60 students across chess, second-language English writing (L2), and mathematics (math): it turns public learner corpora into a fixed per-student record schema with per-student train/held-out splits, and scores every method on two metrics, behavioral fidelity ($\mathcal{F}$) and guidance responsiveness ($\mathcal{R}$). Because every method is fit on the same per-student records and scored on the same held-out records under the same metrics, \textsc{StudentSimEval} makes per-student simulators directly comparable, whether they come from our framework or from prior families. Applying our training framework yields a reference family of 60 \textsc{StudentSim} simulators. The learner corpora are public, and we release the construction pipeline, per-student splits, and evaluation code so others can score new methods on \textsc{StudentSimEval} and extend it.

Across all three domains the reference simulators outperform GPT-5.4 \citep{openai2026gpt54}, a strong closed-sourced LLM prompted to role-play students, on both behavioral fidelity and guidance responsiveness, both on average and on most individual students. For example, in chess, \textsc{StudentSim} reaches $\mathcal{F}=0.51$ and $\mathcal{R}=0.91$, compared with $\mathcal{F}=0.23$ and $\mathcal{R}=0.72$ for GPT-5.4 and $\mathcal{F}=0.45$ and $\mathcal{R}=0.27$ for Maia2 \citep{tang2024maia2unifiedmodelhumanai}, a skill-conditioned chess move-prediction model that stands in for the state-tracking family. The results show that each baseline is weak on a different axis: GPT-5.4 follows tutor guidance fluently, which earns it a fair $\mathcal{R}$, but prompting alone cannot make it reproduce a particular student's competence and mistakes, so its $\mathcal{F}$ stays low. Maia2 is trained on real human games and tracks a player's move distribution reasonably well, earning a moderate $\mathcal{F}$, but it has no input pathway for natural-language guidance and stays near its zero-instruction floor, so its $\mathcal{R}$ collapses. \textsc{StudentSim} is the only model strong on both axes. The same pattern holds in L2 and math, with full per-domain numbers in Sections~\ref{sec:fidelity_results} and \ref{sec:guidance_results}.

We also give a proof of concept that connects student simulation back to AI tutor improvement. Using a trained \textsc{StudentSim} as the reward model for chess tutor reinforcement learning produces an AI tutor that expert human raters score as more accurate, better-guided, and more personalized than both the supervised-finetuned starting tutor with no RL and a tutor RL-trained against a GPT-5.4 student-simulator reward (\cref{sec:tutor_rl}). We do not present this as a best-tutor claim, but as evidence that trained student simulators can provide useful, practical proxy feedback for AI tutor optimization.

In sum, our contributions are: \textit{i)} we present the first framework to cast personalized student simulation as a concrete target for AI tutor optimization, jointly formalizing the two properties a simulator must have, behavioral fidelity ($\mathcal{F}$) and guidance responsiveness ($\mathcal{R}$); this $\mathcal{F}$/$\mathcal{R}$ decomposition is what turns student simulation into an optimizable objective; \textit{ii)} \textsc{StudentSim}, which overcomes sparse per-student data through pooled training and per-student specialization to realize both properties; \textit{iii)} \textsc{StudentSimEval}, a standardized protocol over $60$ students in chess, L2 English writing, and math, with released code, under which our reference family of simulators outperforms state-tracking and prompted-LLM baselines; and \textit{iv)} a chess tutor reinforcement-learning proof of concept in which a trained \textsc{StudentSim} reward yields the best-rated AI tutor in an expert human study.

\section{Related Work}
\label{sec:related_work}

We organize prior work on student simulation and AI tutor optimization into three lines. First, cognitive-state and behavior-prediction models, including knowledge tracing and human-play move predictors, match learner behavior at the population level but have no input channel for natural-language guidance \citep{Corbett_1995,piech2015deepknowledgetracing,ghosh2020contextawareattentiveknowledgetracing,mcilroyyoung2020aligningsuperhumanaihuman,tang2024maia2unifiedmodelhumanai}. Second, LLM-prompted student and user simulators are conditioned on textual descriptions and engage with conversational text \citep{owoicho2023exploitingsimulateduserfeedback,terragni2023incontextlearningusersimulators}; this recipe has been instantiated across educational settings as dialogue-tutoring corpora, multi-agent classrooms, application-specific simulators, and learner-data generation \citep{macina2023mathdialdialoguetutoringdataset, zhang2024simulatingclassroomeducationllmempowered, gao2026agent4edugeneratinglearnerresponse, xu2025classroomsimulacrabuildingcontextual, xu2024eduagentgenerativestudentagents, lu2024generativestudentsusingllmsimulated, Markel_2023}. Competence description alone is, however, only a thin conditioning channel, and a growing validity-focused literature raises concerns about LLM-prompted simulators from architectural, fidelity-benchmark, and teacher-facing angles \citep{yuan2026validstudentsimulationlarge, scarlatos2026simulatedstudentstutoringdialogues, Wu_2025, Martynova_2025, mannekote2024llmsreliablysimulatehuman, louie2024roleplaydohenablingdomainexpertscreate, peng2024quantifyingoptimizingglobalfaithfulness, samuel2025personagymevaluatingpersonaagents, wang2024incharacterevaluatingpersonalityfidelity}. Third, tutor-side work supervises tutor explanations on expert-curated dialogues \citep{chevalier2024languagemodelssciencetutors} or queries a generic LLM judge against a quality rubric \citep{zheng2023judgingllmasajudgemtbenchchatbot}; recent attempts to ground the tutor's reward in rubric-judge feedback or simulated-student use a knowledge-tracing model or a prompted LLM student rather than a simulator trained on real learner data \citep{scarlatos2025trainingllmbasedtutorsimprove, dinucujianu2025problemsolvingteachingproblemsolvingaligning, maurya2025unifyingaitutorevaluation}, while a parallel tutor-side line builds pedagogical evaluation benchmarks and runs classroom or expert-preference studies of tutor models \citep{macina2025mathtutorbenchbenchmarkmeasuringopenended, wang2025tutorcopilothumanaiapproach, wang2024bridgingnoviceexpertgapmodels, learnlmteam2025learnlmimprovinggeminilearning}.

To our knowledge, \textsc{StudentSim} is the first to \textit{i)} formalize per-student simulation as a measurable target defined jointly by behavioral fidelity and guidance responsiveness; \textit{ii)} train per-student simulators on real per-individual learner data that both match a student's own behavior and respond to natural-language tutor guidance; and \textit{iii)} use such a trained simulator as the reward source for optimizing an AI tutor (\cref{sec:tutor_rl}). \cref{app:related_work} gives the per-strand survey with methodological detail.

\section{Problem Formulation}
\label{sec:problem_formulation}

Let $\{\pi_1, \dots, \pi_N\}$ be $N$ real students and $\{M_1, \dots, M_N\}$ be trained student simulators, where $M_i$ is meant to approximate $\pi_i$. For each student $i$, we hold out two kinds of recorded data. \emph{Single-turn records}, $S_i = \{(x, m)\}$, pair a problem $x$ with the student's actual response $m$; $S_i$ serves to measure whether the simulator's prediction matches the student's response. \emph{Multi-turn records}, $T_i = \{(x, m, \tau, m^*)\}$, additionally include the tutor guidance $\tau$ provided to address the student's wrong response $m$ on $x$ and the canonical corrected response $m^*$ that the guidance is steering toward (e.g., the engine-recommended move in chess, the corrected fragment in second-language English writing, or the correct answer in mathematics); $T_i$ serves to measure whether the simulator successfully updates toward $m^*$ after reading guidance. A useful population must answer two questions: does each simulator match its target student's behavior on $S_i$, and does it produce the corrected response on $T_i$ after reading guidance?

\paragraph{Behavioral fidelity.}
Per-student behavioral fidelity, $\mathcal{F}_i$, scores simulator $M_i$ on $S_i$: how well its response on a held-out problem matches student $\pi_i$'s recorded response. Each of three domains tested in \textsc{StudentSimEval} instantiates this match notion to its own response space. In chess, fidelity is move-prediction accuracy: how often the simulator's predicted move is the same move the student actually played. In second-language English writing, fidelity is error-profile match: whether the simulator's generated essay matches the student's overall error rate and issue-type profile. In mathematics, where problems are four-way multiple-choice, fidelity is how faithfully the simulator's prediction matches the option the student actually selected. The population score is $\mathcal{F} = \tfrac{1}{N} \sum_i \mathcal{F}_i$. Appendix~\ref{app:metric_definitions} gives the per-benchmark formula and operational recipe; Appendix~\ref{app:design_f_per_domain} gives the design rationale for the per-domain operationalization.

\paragraph{Guidance responsiveness.}
Per-student guidance responsiveness, $\mathcal{R}_i$, scores simulator $M_i$ on $T_i$ using multi-turn records $(x, m, \tau, m^*)$, where $x$ is the problem, $m$ is the student's initial response, $\tau$ is the tutor guidance addressing it, and $m^*$ is the response that guidance is steering toward. $\mathcal{R}_i$ asks: after reading $(x, m, \tau)$, does the simulator update to $m^*$? The population score is $\mathcal{R} = \tfrac{1}{N} \sum_i \mathcal{R}_i$. This isolates the simulator's update under guidance: behavioral fidelity already captures the student-specific answer patterns present before the intervention, and $\mathcal{R}$ captures whether the simulator takes up the guidance and follows it to the endpoint it steers toward. That is the operative property when the simulator serves as a reward model for tutor training (Section~\ref{sec:tutor_rl}), since a static imitator that ignores guidance offers no signal about whether an intervention moved the student. Appendix~\ref{app:design_g_semantics} details this target choice.

\paragraph{Orthogonality.}
$\mathcal{F}$ and $\mathcal{R}$ measure separable capabilities: a faithful but unresponsive simulator describes the student without responding to guidance, and a responsive but unfaithful one follows guidance from the wrong starting state. Therefore, the high-$\mathcal{F}$, high-$\mathcal{R}$ corner is the target.

\section{StudentSim Training Pipeline}
\label{sec:method}

\begin{figure}[t]
  \centering
  \includegraphics[width=\linewidth]{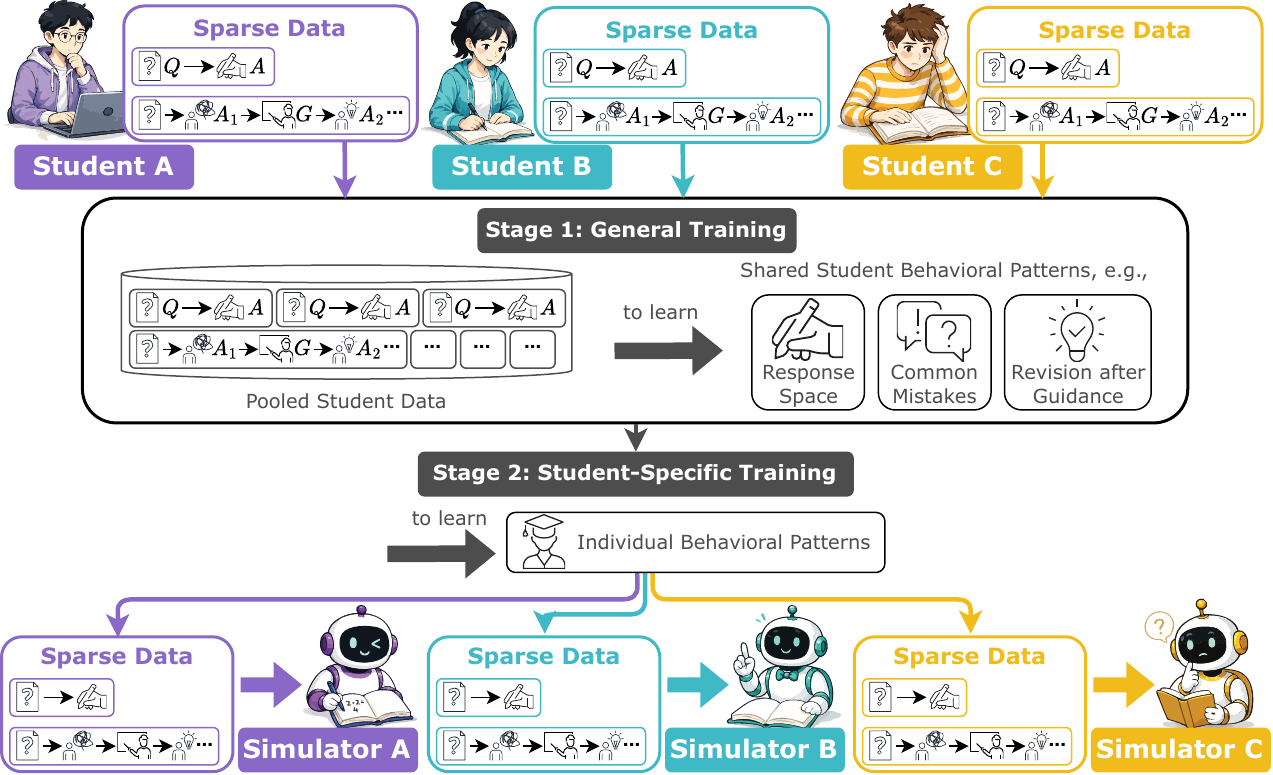}
  \caption{\textbf{Two-stage training pipeline for \textsc{StudentSim}.} Stage 1 pools the sparse records of many students in a domain and trains one base simulator on what they share. Stage 2 specializes that base to a single student on that student's own records, yielding one simulator per student.}
  \label{fig:training_pipeline}
\end{figure}

Each student in the populations we study contributes only a small amount of recorded data. For example, a learner in the second-language English writing corpus \citep{Oksuz_2025} writes three essays at the median, and more than two thirds write five or fewer (Appendix~\ref{app:benchmark_details}). This sparsity is not specific to our setup: collecting records from real users is slow and costly, and assembling per-user data on this order typically takes months to years of collection, so per-user data sparsity is a familiar problem in any system that learns from interactions with real users \citep{Schein_2002, lee2019melumetalearneduserpreference, bhattacharjee2025coldstartproblemexperimental, li2017usersimulatortaskcompletiondialogues}. It is also a problem that data-hungry large language models are particularly poorly equipped to solve directly: training one simulator end-to-end per student from a small per-student record would overfit, and would also have to relearn the shared structure of student behavior in the domain from scratch every time \citep{dodge2020finetuningpretrainedlanguagemodels, finn2017modelagnosticmetalearningfastadaptation}. We therefore train each per-student simulator in two stages (\cref{fig:training_pipeline}).

\paragraph{Stage 1: pooled training.}
Stage 1 trains one domain-specific base simulator on data pooled across all students in the domain. This stage learns what is shared across students: common mistake patterns, the response format expected in the domain, and the pathway from natural-language tutor guidance to a changed response. The output of Stage 1 is one base simulator per domain that already exhibits broad student-like behavior but has not yet been specialized to any individual student.

\paragraph{Stage 2: per-student specialization.}
Stage 2 initializes from the Stage 1 base simulator and adapts it to one specific student using only that student's own recorded responses and recorded tutor-guidance interactions. Each student receives an independent specialization, producing one simulator $M_i$ per student. Stage 2 captures what is specific to the individual student: for example, which mistakes a particular student tends to make, and how that student's behavior shifts in response to a given form of tutor guidance.

\paragraph{Why two stages.}
Stage 1 supplies what is shared across students and what cannot be learned reliably from one student's sparse data alone; for example, in chess, performance drops when pooled records are replaced with repeated records from one student (Appendix~\ref{app:ablation_cross_student_pooling}). Stage 2 supplies what is specific to each individual student.

\section{Results}
\label{sec:experiment}

This section instantiates the framework on chess, second-language English writing (L2), and mathematics (math). \cref{sec:experiment_setup} specifies data sources, tasks, guidance types, training setup, and baselines. \cref{sec:fidelity_results} reports behavioral fidelity $\mathcal{F}$. \cref{sec:guidance_results} reports guidance responsiveness $\mathcal{R}$.

\subsection{Experiment Setup}
\label{sec:experiment_setup}

\paragraph{Data sources, tasks, and student populations.}
Each domain provides single-turn records (problem $\to$ student response) used to learn a student's baseline behavior and multi-turn records (problem $\to$ wrong response $\to$ tutor guidance $\to$ canonical corrected response) used to learn how a student updates after tutor guidance, instantiating the $(S_i, T_i)$ schema from \cref{sec:problem_formulation}. All three corpora consist of records from real human students. For chess, the corpus is drawn from Lichess (May 2025), a fully open-source online chess platform on which players record real games against human opponents; each record pairs a board position with the player's actual move. The simulator task is to predict a specific player's next move and their response to natural-language coach guidance. For second-language English writing (L2), we use the EFCAMDAT corpus \citep{geertzen2013automatic}, a collection of essays by English-as-foreign-language learners alongside tutor corrections; the simulator task is to generate essays whose error patterns match a specific learner, and to predict the learner's response after a tutor correction. For mathematics (math), we use a foundational-assistance corpus \citep{worden2026foundationalassisteducationaldatasetfoundational} of open-ended problems annotated with student answers and correct answer; the simulator task is to predict a specific student's answer option and their response after a tutor explanation. In all three domains, Stage 1 pretrains on a larger pool of students and Stage 2 specializes per-student simulators on a smaller held-out subset; per-student evaluation records are disjoint from training data. \textsc{StudentSimEval} evaluates a fixed roster of $30$ chess players, $15$ L2 learners, and $15$ math students, each on a per-student held-out set that is frozen and scored identically for every method; these per-student held-out sizes are the last two columns of \cref{tab:training_records}, and the full evaluation protocol is in Appendix~\ref{app:studentsimeval_protocol}. \cref{tab:training_records} summarizes the per-domain Stage 1 and Stage 2 training data scale. Construction of the multi-turn corpora (real-teacher-annotated for L2, controllable LLM-tutor-generated for chess and math under fixed style templates) is detailed in Appendix~\ref{app:design_multiturn_corpus}; for chess and math, the LLM determines only the wording of the guidance, while the response it steers toward is fixed in advance by the engine or an audited answer key.

\begin{table}[t]
\centering
\small
\setlength{\tabcolsep}{3pt}
\caption{\textbf{Per-domain training data scale: Stage 1 (pooled training), Stage 2 (per-student specialization), and held-out evaluation per Stage-2 student.} Stage 1 / Stage 2 \emph{instances} count training records (single-turn plus multi-turn pooled); the multi-turn ratio is the fraction of multi-turn records ($T_i$) sampled into training batches, with the remainder single-turn ($S_i$). The held-out columns count per-student evaluation records: $S$ is the single-turn split, and $T$ the multi-turn split. Each count is the same for every student in the domain, except math $T$ (mean shown, range in parentheses): each student has too few wrong-answer problems for a uniform cap to leave enough per student, so we keep all of them, and the count varies with how many the student first answered incorrectly.}
\label{tab:training_records}
\begin{tabular}{@{}lrrrrcrr@{}}
\toprule
\multirow{2}{*}{\textbf{\boldmath Domain}} & \multicolumn{2}{c}{\textbf{\boldmath Stage 1 (pooled)}} & \multicolumn{2}{c}{\textbf{\boldmath Stage 2 (per-student)}} & \multirow{2}{*}{\textbf{\boldmath Multi-turn ratio}} & \multicolumn{2}{c}{\textbf{\boldmath Held-out (per-student)}} \\
\cmidrule(lr){2-3} \cmidrule(lr){4-5} \cmidrule(lr){7-8}
& \textbf{\boldmath students} & \textbf{\boldmath instances} & \textbf{\boldmath students} & \textbf{\boldmath instances} & & \textbf{\boldmath $S$} & \textbf{\boldmath $T$} \\
\midrule
chess & $100$ & $100{,}000$ & $30$ & $1{,}000$ & $0.20$ & $5{,}000$ & $4{,}000$ \\
L2    & $200$ & $7{,}800$   & $15$ & $73$       & $0.20$ & $26$      & $40$      \\
math  & $200$ & $23{,}400$  & $15$ & $153$      & $0.20$ & $66$      & $59$ ($21$--$99$) \\
\bottomrule
\end{tabular}
\end{table}

\paragraph{Guidance types.}
Effective tutoring adjusts the level of support to what the learner needs at that moment, giving more when the student is struggling and stepping back as the student takes over \citep{Wood_1976}. Each domain therefore includes guidance types that span that range of support: chess uses error remediation (diagnosing the error and explaining the correction), comparative (contrasting the played move with stronger alternatives), strategic (outlining the plan the chess position calls for), and Socratic (asking questions that guide the student's reasoning); L2 uses point-based (scoping feedback to one error occurrence) and rule-based (explaining the underlying grammatical rule) correction; math uses error remediation, Socratic, and conceptual explanation (teaching the principle behind the error). Error remediation covers the direct-correction end of that range, the appropriate tutor action when a learner is stuck or repeatedly failing \citep{Koedinger_2007}. A tutor trained against the simulator has to learn when to use that level of support, so the training environment must include students who respond to it. At the other end, Socratic and conceptual modes correspond to the case where the tutor asks questions or supplies a principle without stating which response is correct, so the learner must work the correction out for themselves. The simulated student must then reason through those hints to arrive at the correction, making these modes the hardest case for guidance responsiveness. Multi-turn records mix these guidance types uniformly during training so the simulator learns to follow any of them at evaluation time; per-mode breakdown is in Appendix~\ref{subsec:r_per_mode}.

\paragraph{Training setup.}
The base model in all three domains is Qwen3-4B-Instruct \citep{yang2025qwen3technicalreport}. Stage 1 trains one domain-specific LoRA adapter on records pooled across the Stage-1 student set, mixing single-turn and multi-turn records at a multi-turn ratio of $0.2$. Single-turn records train fidelity, multi-turn records train guidance responsiveness, and joint training develops both. Stage 2 continues training the Stage-1 adapter on each Stage-2 student's own records, producing one specialized adapter per student. This two-stage design separates the learning of population-level behavior patterns (Stage 1, pooled training across all students) from per-student specialization (Stage 2), so each student's adapter refines a well-initialized shared foundation rather than training from scratch on sparse individual data. An ablation of this two-stage design is given in Appendix~\ref{app:ablation_cross_student_pooling}. Both stages share the base model, LoRA architecture, and optimizer; the small number of per-domain differences in batch size and optimization length follow directly from each domain's dataset scale, with all other settings held identical across domains. Decoding is greedy ($T = 0$). Headline numbers in Sections~\ref{sec:fidelity_results}--\ref{sec:guidance_results} are the cross-seed mean over 3 independent training runs; run-to-run standard deviations are in Appendix~\ref{app:results_std}. Full hyperparameters are in Appendix~\ref{app:training_details}.

\paragraph{Baselines.}
We compare against three categories of reference baselines: \textit{i)} the closed-source LLM baselines GPT-4o \citep{openai2024gpt4o} and GPT-5.4 \citep{openai2026gpt54}, prompted in-context with each student's profile, problem, and guidance; \textit{ii)} the base Qwen3-4B-Instruct, prompted identically but without per-domain training, for L2 and math (we omit it for chess because an untrained Qwen3-4B-Instruct cannot reliably predict legal chess moves, so its near-zero $\mathcal{F}$ would not be informative); and \textit{iii)} for chess only, Maia2 \citep{tang2024maia2unifiedmodelhumanai}, a chess-specific human-style move predictor conditioned on FEN and player ELO. We do not have an analogous domain-specific behavior model for L2 or math: chess has a fixed, narrow task scope and an active community that has trained dedicated player-behavior engines, while knowledge-acquisition domains such as second-language writing and math problem-solving have no such specialized model. For L2 and math, the closed-source LLM baselines therefore stand in as the strongest available reference. All baselines are scored on the same held-out records as our trained simulators. Appendix~\ref{app:design_baselines} gives the full per-domain baseline details.

\subsection{Behavioral Fidelity Results}
\label{sec:fidelity_results}

\begin{table}[t]
\centering
\small
\caption{\textbf{Behavioral fidelity ($\mathcal{F}$) by domain.} $\mathcal{F}$ is how well the simulator's response to a problem matches the student's own recorded response. \emph{Naive baseline} is Maia2 for chess (a chess-specific human-style move predictor), and Qwen3-4B-Instruct without per-domain training for L2 and math; chess Qwen3-4B-Instruct is omitted because the untrained model cannot reliably predict chess moves and its near-zero score would not be informative. Run-to-run standard deviations across training seeds are reported in Appendix~\ref{app:results_std}.}
\setlength{\tabcolsep}{3pt}
\begin{tabular}{l l r r r r}
\toprule
\textbf{\boldmath Domain} & \textbf{\boldmath $\mathcal{F}$ Definition} & \textbf{\boldmath Naive Baseline} & \textbf{\boldmath GPT-4o} & \textbf{\boldmath GPT-5.4} & \textsc{\textbf{StudentSim}} \\
\midrule
Chess (30 players)  & top-1 accuracy ($\uparrow$)        & $0.4535$  & $0.2163$  & $0.2316$  & $\boldsymbol{0.5150}$ \\
L2 (15 learners)    & error-profile match ($\uparrow$)   & $0.5130$  & $0.4718$  & $0.5141$  & $\boldsymbol{0.5624}$ \\
Math (15 students)  & K=4 MC accuracy ($\uparrow$)       & $0.4919$  & $0.5121$  & $0.6121$  & $\boldsymbol{0.6384}$ \\
\bottomrule
\end{tabular}
\label{tab:fidelity_main}
\end{table}

\begin{figure}[t]
  \centering
  \begin{minipage}[b]{0.245\linewidth}\centering
    \includegraphics[width=\linewidth]{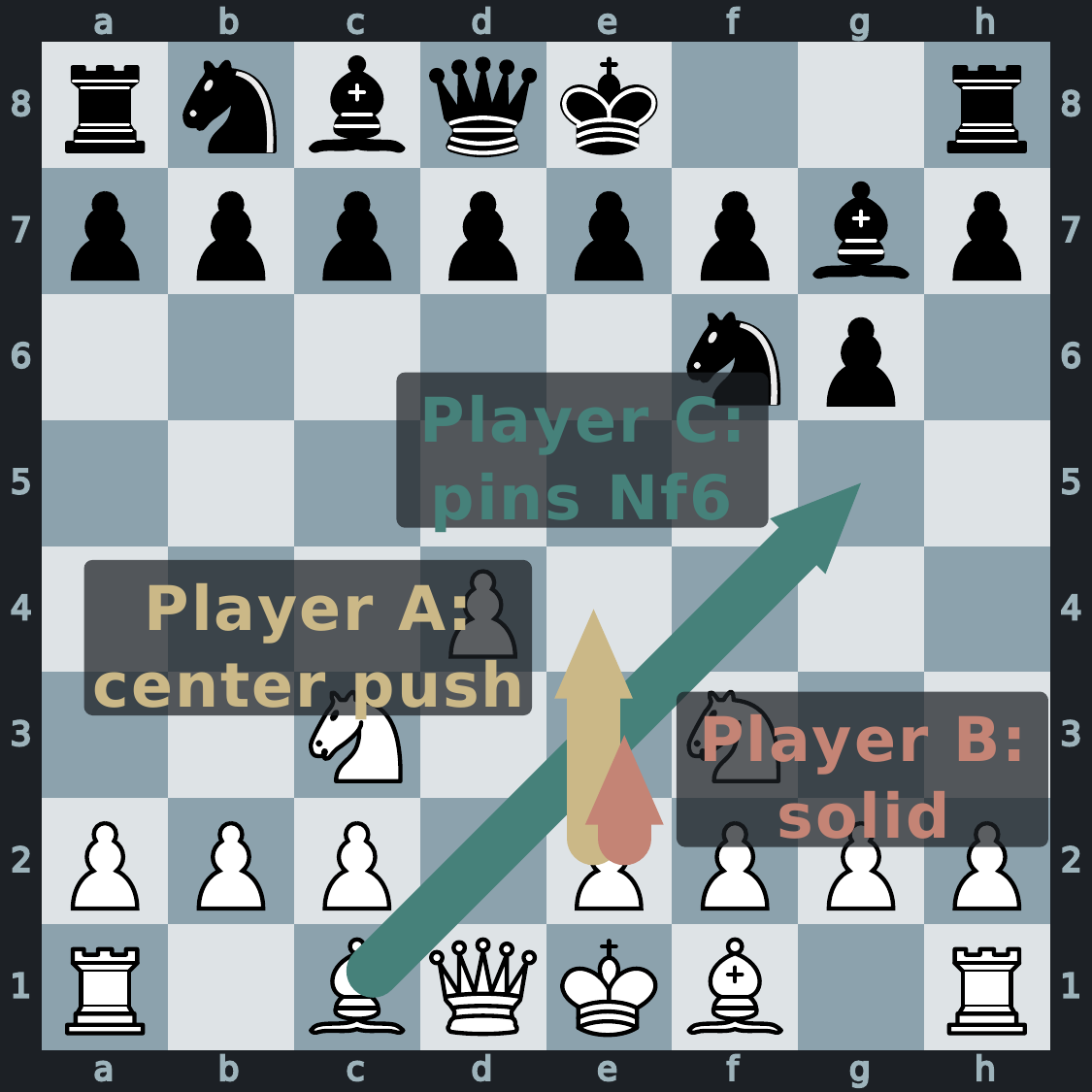}\\[2pt]
    \footnotesize (a) Ground truth
  \end{minipage}\hfill
  \begin{minipage}[b]{0.245\linewidth}\centering
    \includegraphics[width=\linewidth]{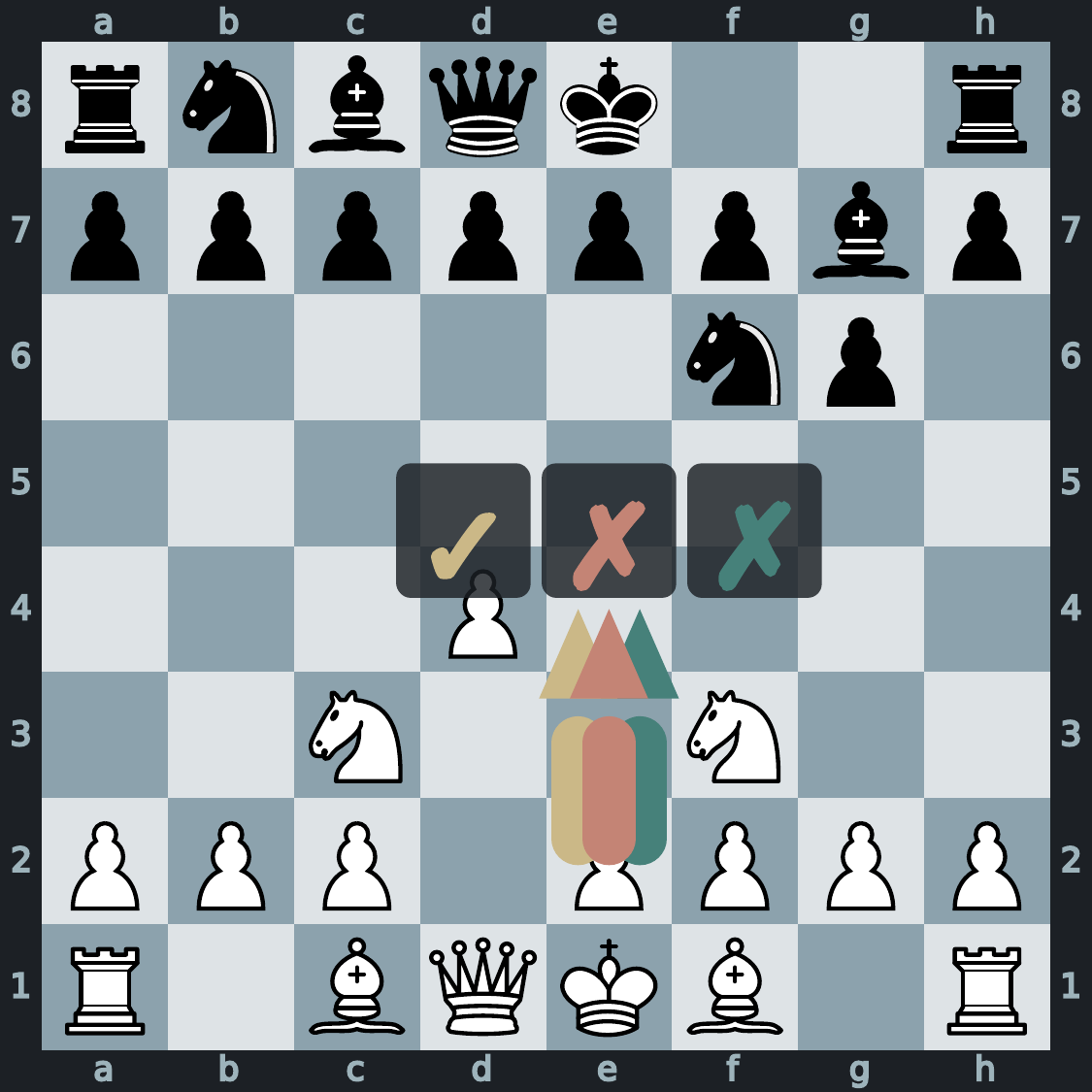}\\[2pt]
    \footnotesize (b) Maia2
  \end{minipage}\hfill
  \begin{minipage}[b]{0.245\linewidth}\centering
    \includegraphics[width=\linewidth]{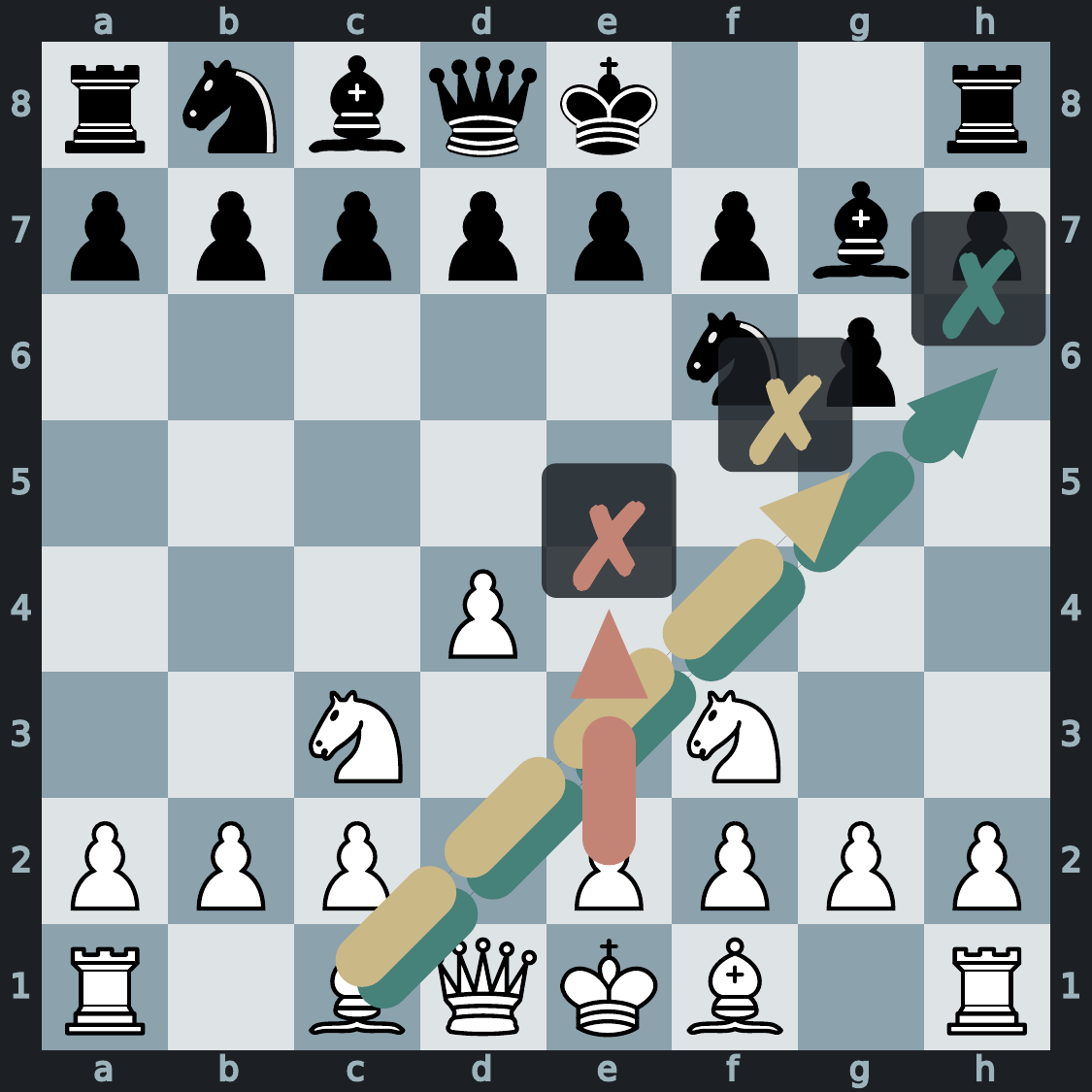}\\[2pt]
    \footnotesize (c) GPT-5.4
  \end{minipage}\hfill
  \begin{minipage}[b]{0.245\linewidth}\centering
    \includegraphics[width=\linewidth]{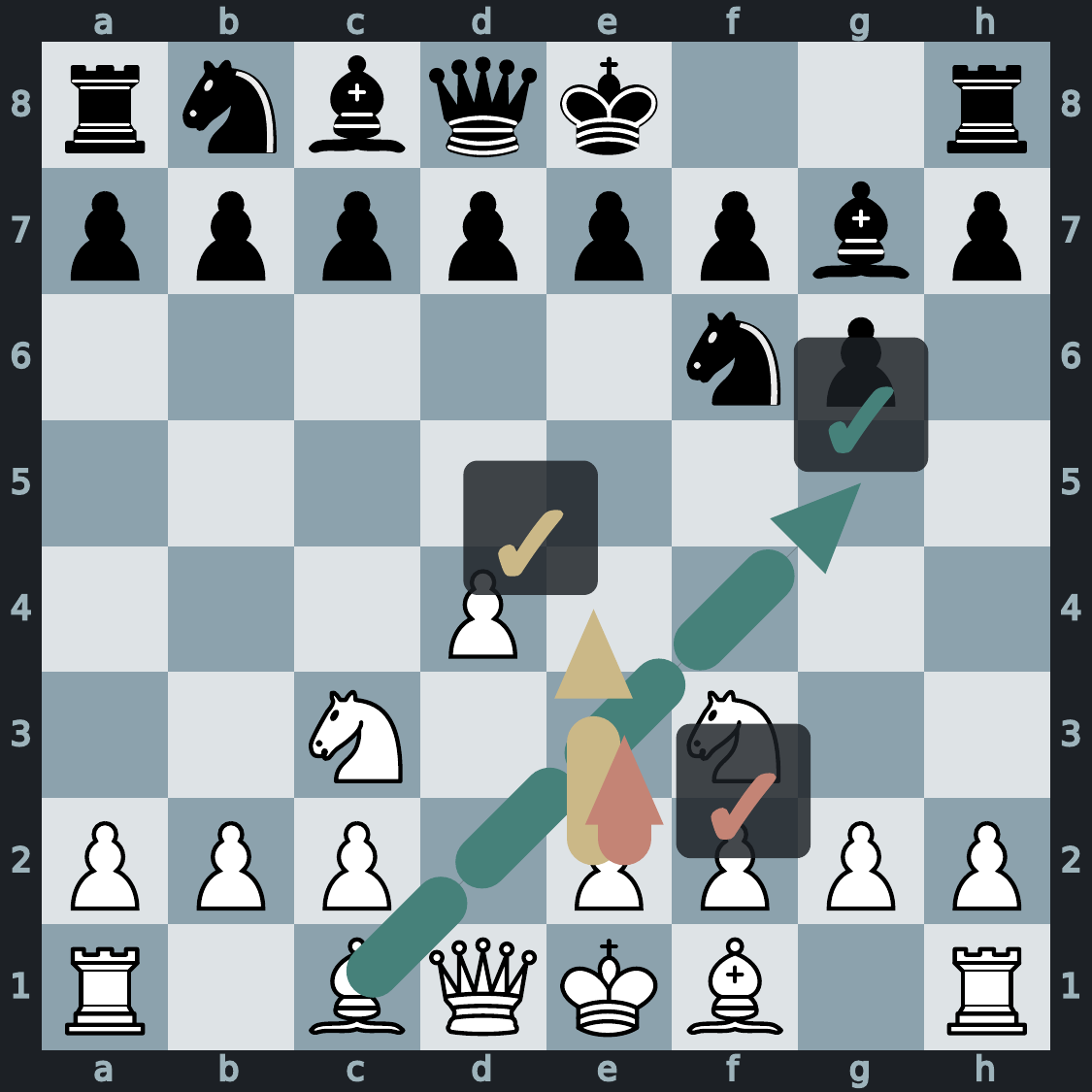}\\[2pt]
    \footnotesize (d) \textsc{StudentSim}
  \end{minipage}
  \caption{\textbf{Behavior fidelity on a held-out chess position.} Three real players in our held-out set choose three different moves at this board (panel a): Player A plays \texttt{e4} (gold, central push), Player B plays \texttt{e3} (rose, solid), Player C plays \texttt{Bg5} (teal, pin on \texttt{Nf6}). Maia2 (panel b) collapses all three onto the ELO-modal \texttt{e4}. GPT-5.4 (panel c), prompted in-context with each player's recent history, misses all three (\texttt{Bg5} / \texttt{e4} / \texttt{Bh6}). \textsc{StudentSim} (panel d) reproduces all three actual moves. Solid arrows are ground truth; dashed arrows are predictions; {\color[RGB]{60,120,75}$\checkmark$}\,/\,{\color{red}$\times$} mark per-player match. On the full $30$-player held-out set, \textsc{StudentSim} reaches per-player top-1 fidelity $\mathcal{F}=\boldsymbol{0.51}$ vs.\ Maia2 $\mathcal{F}=0.45$, GPT-5.4 $\mathcal{F}=0.23$, and GPT-4o $\mathcal{F}=0.22$.}
  \label{fig:fidelity_comparison}
\end{figure}

The trained simulators outperform every reference baseline on $\mathcal{F}$ across all three domains (\cref{tab:fidelity_main}). A higher $\mathcal{F}$ has a different concrete meaning in each benchmark: in chess it means the simulator predicts the player's actual next move more accurately; in L2 it means the simulator's essays match, more closely, the rate and category profile of the mistakes a particular learner still makes on a writing prompt; in math it means the simulator more often selects, among the answer choices, the specific answer (correct or not) that this student actually gave. We use chess as the case study below because per-student behavior distinction shows up most directly at a single board position.

\cref{fig:fidelity_comparison} shows one held-out position where three real players in our test set choose three different moves. Each move reflects a different style: center push, solid, and tactical pin. The trained simulators reproduce all three actual moves, one per player. This is the capability per-student training is meant to add: at a fixed input, the simulator's output reflects the specific student it was trained on rather than a population-level prediction. Neither baseline matches this resolution. Maia2 conditions on player ELO only and collapses all three players onto its ELO-modal \texttt{e4}. GPT-5.4 prompted with each player's textual history also misses all three, because a textual description of past games does not constrain move choice at per-student resolution.

\subsection{Guidance Responsiveness Results}
\label{sec:guidance_results}

\begin{table}[t]
\centering
\small
\caption{\textbf{Guidance responsiveness ($\mathcal{R}$) by domain.} $\mathcal{R}$ is how well the simulator revises its initial wrong response to the canonical correction after reading the tutor's guidance. \emph{Naive baseline} is Maia2 for chess (no natural-language input pathway, so its score is a zero-guidance lower bound); Qwen3-4B-Instruct without per-domain training for L2 and math. Run-to-run standard deviations across training seeds are reported in Appendix~\ref{app:results_std}.}
\setlength{\tabcolsep}{3pt}
\begin{tabular}{l l r r r r}
\toprule
\textbf{\boldmath Domain} & \textbf{\boldmath $\mathcal{R}$ Definition} & \textbf{\boldmath Naive Baseline} & \textbf{\boldmath GPT-4o} & \textbf{\boldmath GPT-5.4} & \textsc{\textbf{StudentSim}} \\
\midrule
Chess (30 players)  & corrected-move rate ($\uparrow$)   & $0.2721$ & $0.7655$ & $0.7186$ & $\boldsymbol{0.9067}$ \\
L2 (15 learners)    & fragment-rewrite match ($\uparrow$) & $0.0200$ & $0.3883$ & $0.5950$ & $\boldsymbol{0.6417}$ \\
Math (15 students)  & answer-correction rate ($\uparrow$) & $0.6132$ & $0.6940$ & $0.7099$ & $\boldsymbol{0.9181}$ \\
\bottomrule
\end{tabular}
\label{tab:guidance_main}
\end{table}

\begin{figure}[t]
  \centering
  \begin{minipage}[c]{0.35\textwidth}
    \centering
    \includegraphics[width=\linewidth]{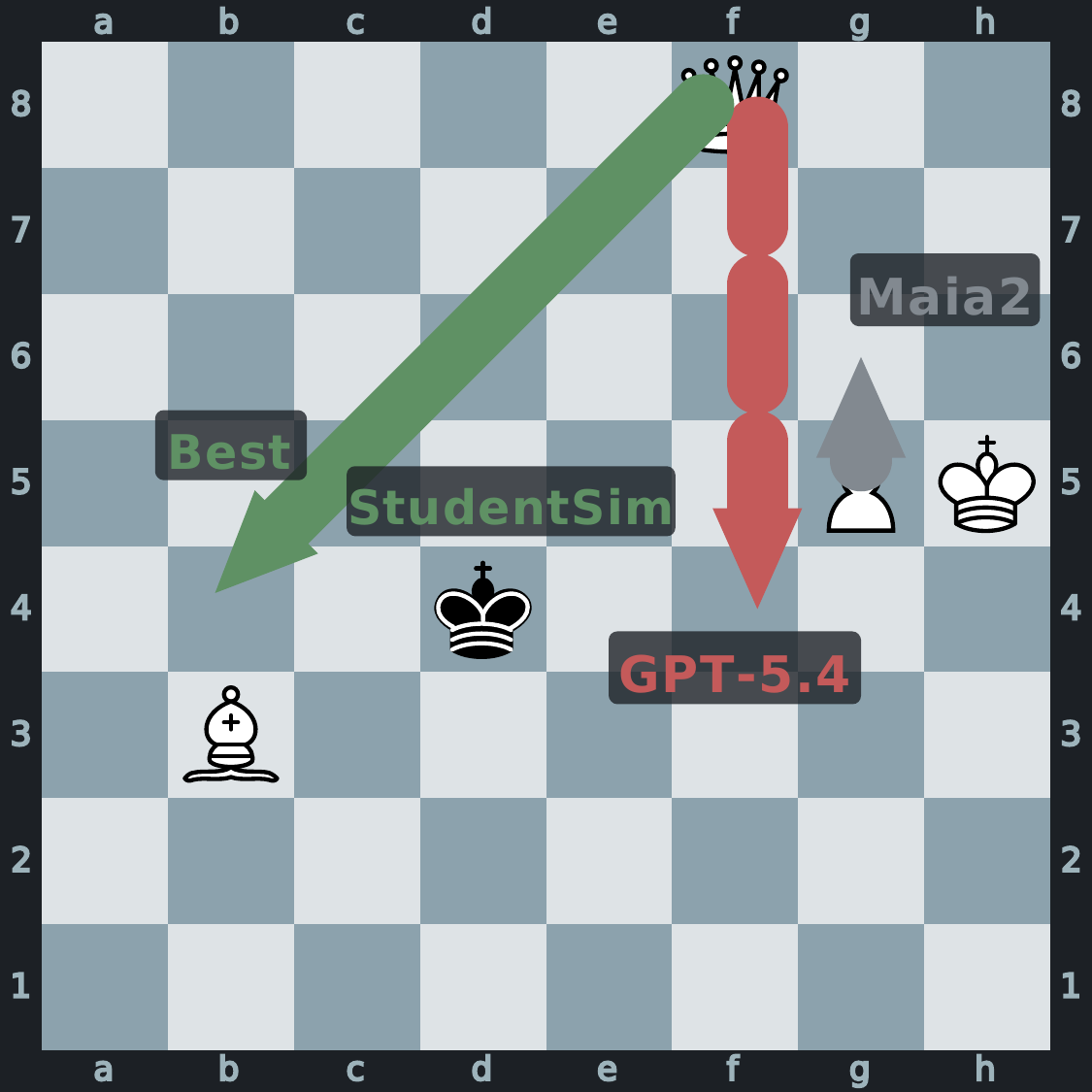}
    \\[2pt]
    \footnotesize White to move.\quad FEN \texttt{5Q2/8/8/6PK/3k4/1B6/8/8 w}
  \end{minipage}\hfill
  \begin{minipage}[c]{0.56\textwidth}
    \footnotesize
    Socratic guidance (abridged):\\[2pt]
    \textit{``Compare \texttt{1.g5g6} with a forcing queen move that gives check from the queenside. After such a check, where is Black's king forced to go, and do you gain a useful tempo? \dots Is it stronger to push the pawn immediately, or to cut off the enemy king with the queen first?''}\\[3pt]
    The guidance never names the destination square.\\[4pt]
    \begin{tabular}{@{}lcc@{}}
      \toprule
      \textbf{Source} & \textbf{Move} & \textbf{Follows guidance?} \\
      \midrule
      Player's actual move             & \textcolor[RGB]{105,112,120}{\texttt{g5g6}} & n/a \\
      Guidance target                  & \textcolor[RGB]{60,120,75}{\texttt{f8b4}} & n/a \\
      \midrule
      Maia2 (no guidance support)      & \textcolor[RGB]{105,112,120}{\texttt{g5g6}} & {\color{red}$\times$} \\
      GPT-5.4 (after guidance)         & \textcolor[RGB]{180,80,80}{\texttt{f8f4}}  & {\color{red}$\times$} \\
      \textbf{\textsc{StudentSim} (Ours)} (after guidance)   & \textcolor[RGB]{60,120,75}{\texttt{f8b4}} & {\color[RGB]{60,120,75}$\checkmark$} \\
      \bottomrule
    \end{tabular}
  \end{minipage}
  \caption{\textbf{Guidance-following capability on a held-out chess position.} The held-out player blundered with \texttt{1.g5g6} (centipawn loss $=8638$), missing a forced queen check on \texttt{f8b4}. The tutor's socratic prompt contrasts the pawn push with ``a forcing queen move that gives check from the queenside'' but never names a destination square. Arrows on the board show each source's predicted move. Maia2 cannot incorporate the prompt and reproduces the blunder. GPT-5.4 reads the prompt in-context yet outputs a different wrong queen move. Our per-player fine-tuned model follows the open-ended guidance and outputs the engine best move. Across the full guidance-following test set, \textsc{StudentSim} reaches $\mathcal{R}=\boldsymbol{0.91}$, vs.\ GPT-4o $\mathcal{R}=0.77$, GPT-5.4 $\mathcal{R}=0.72$, and Maia2 $\mathcal{R}=0.27$.}
  \label{fig:guidance_following_chess}
\end{figure}

Across all three domains, the trained simulators outperform every baseline on aggregate $\mathcal{R}$ (\cref{tab:guidance_main}); the gap is largest on the open-ended guidance modes, where the tutor prompts the student to reason toward the fix through questions or principles. A higher $\mathcal{R}$ also has a different concrete meaning per benchmark: in chess, after reading the tutor's textual guidance the simulator switches from its initial wrong move to a corrected move that the guidance is steering it toward; in L2, after reading the teacher's feedback, which flags the offending fragment of an essay, the simulator supplies the exact correction that fragment requires; in math, after the tutor walks the student through the reasoning behind their error, the simulator's ability to revise its initial wrong numerical answer to the correct one improves. The case study below illustrates this on the strictest of the chess modes.

\cref{fig:guidance_following_chess} examines the strictest of the four chess guidance modes, Socratic. Here the tutor leads the student toward the fix through a chain of questions and does not state the corrected move, so the simulator must infer the canonical correction from that reasoning. The position is held out from both pooled training and per-player training. The player's actual move is the pawn push \texttt{1.g5g6}, a serious blunder; the engine best move is the queen check \texttt{f8b4}. The Socratic prompt asks the student to compare the pawn push with ``a forcing queen move that gives check from the queenside''. The prompt never names the destination square; the model has to derive the answer from the chain of guiding questions.

The baselines fail in three different ways. Maia2 has no input pathway for natural-language guidance and reproduces the player's blunder. GPT-5.4 reads the prompt and selects a queen move, but answers a different wrong square (\texttt{f8f4}). Our per-player simulator, a $4$B Qwen3 base with a single per-player LoRA adapter, reasons through the same chain of questions and outputs the engine best move \texttt{f8b4}. That a $4$B trained simulator solves the Socratic case where prompted access to a much larger closed-source model fails is direct evidence that the multi-turn training stage adds a guidance-following capability, not merely a hint-copying shortcut on the easy modes where the prompt already contains the answer. Per-mode breakdowns by guidance type are in Appendix~\ref{subsec:r_per_mode}.

\section{Tutor RL with Student Simulator Feedback}
\label{sec:tutor_rl}

\begin{figure}[t]
  \centering
  \includegraphics[width=\linewidth]{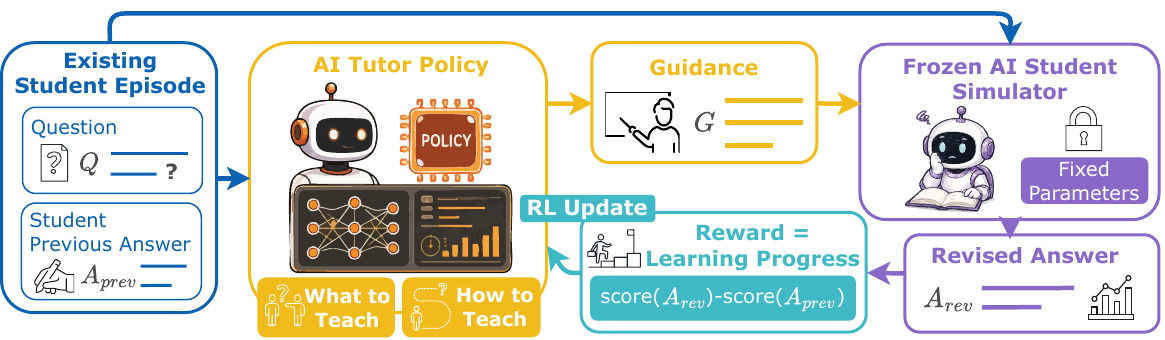}
  \caption{\textbf{Tutor RL with a frozen student simulator as the reward source.} Each episode replays a real student's wrong answer, the tutor policy proposes guidance, the frozen simulator revises, and the improvement in move quality drives the RL update.}
  \label{fig:rl_setup}
\end{figure}

This section gives a chess proof of concept that our trained \textsc{StudentSim}, used as the reward model for tutor reinforcement learning, produces a better tutor than two natural alternatives: the supervised-finetuned tutor with no RL, and a reward in which a frontier LLM (GPT-5.4) plays the student simulator. We keep this a proof of concept: the claim is that a faithful, responsive student simulator is a useful and practical reward signal for AI tutor optimization, not that we produce a best-in-class tutor. Chess is the right scope because it admits a precise per-position reward (Stockfish \citep{stockfish} centipawn evaluation) that decouples the validation signal from the simulator's behavior; transfer to other domains needs per-domain reward functions on free-form responses, outside the claim tested here (Appendix~\ref{app:design_poc_scope}). We evaluate all conditions with an expert human study (Appendix~\ref{app:training_rl}).

\paragraph{Setup.}
The loop is shown in \cref{fig:rl_setup}. Each episode draws a chess problem $Q$ and the student's recorded wrong move $A_\text{prev}$ from real student records; the tutor proposes guidance, a frozen student simulator emits a revised move $A_\text{rev}$, and the reward rewards guidance that improves move quality, derived from a precomputed Stockfish lookup of $A_\text{rev}$ relative to $A_\text{prev}$ (full formulation in Appendix~\ref{app:training_details}). All conditions share the same tutor policy (Qwen3-VL-8B \citep{bai2025qwen3vltechnicalreport}), which reads the position as a rendered board alongside the text, and the same SFT starting point, optimized with GRPO \citep{shao2024deepseekmathpushinglimitsmathematical}; they differ only in the reward model:
\begin{itemize}
\item \textbf{No RL}: the SFT tutor itself, a lower-bound control.
\item \textbf{GPT-5.4 simulator}: the reward uses GPT-5.4 prompted to play the student; GPT-5.4 emits the revised move after reading the tutor's guidance, and the shared move-quality reward scores that move. This is the heavyweight, closed, fixed-behavior baseline that depends on a frontier-model API at every rollout.
\item \textbf{\textsc{StudentSim} (Ours)}: the reward is our trained \textsc{StudentSim} together with two extensible reward heads mounted on the same simulator backbone, both reading the tutor's explanation. The frozen-simulator term is the shared move-quality improvement of the simulator's revised move; a personalization head scores whether the explanation follows the intended teaching style; and a perception head penalizes explanations that misdescribe the board, such as naming a wrong square or a hallucinated piece. The two heads act as multiplicative gates on the move-quality term (Appendix~\ref{app:training_rl}). The simulator and heads are small, open, and served locally, and the head set can be customized to the student role one wants to optimize against.
\end{itemize}

Our reward uses the pooled Stage-1 \textsc{StudentSim} simulator, which captures population-level student behavior, so the tutor is optimized to improve guidance for students in general and the reward is not tied to any single individual's idiosyncrasies.

\paragraph{Expert Human Study.}
After tutor model optimization, competitive chess players rate each tutor's generated guidance, blind to presentation order, on three axes: accuracy (the fraction of a tutor's responses with no misleading factual error), guidance quality (a $1$--$5$ rating of how well the response teaches), and personalization (a $1$--$5$ rating in the context of a Socratic-leaning student). Among the rewards we study, our \textsc{StudentSim} reward produces the best-rated tutor on all three axes (\cref{tab:tutor_rl_human}). The GPT-5.4-simulator reward stays below even the no-RL baseline on accuracy, held back by a markedly higher rate of severe factual errors. Full study details are in Appendix~\ref{app:training_rl}.

\begin{table}[h]
\centering
\small
\caption{\textbf{Human evaluation of the AI chess tutors.} \emph{Accuracy} is the percentage of responses with no actively misleading factual error; \emph{guidance} and \emph{personalization} are $1$--$5$ ratings. Scores are means over the study's $74$ annotations from $8$ annotators. The triply-annotated and $2000$+ elo expert subsets in Appendix~\ref{app:training_rl} give the same ordering.}
\label{tab:tutor_rl_human}
\vspace{1.5mm}
\renewcommand{\arraystretch}{1.2}
\begin{tabular}{@{}lrrr@{}}
\toprule
\textbf{\boldmath Condition} & \textbf{\boldmath Accuracy (\%)} & \textbf{\boldmath Guidance ($1$--$5$, $\uparrow$)} & \textbf{\boldmath Personalization ($1$--$5$, $\uparrow$)} \\
\midrule
No RL           & $75.7$ & $2.99$ & $2.80$ \\
GPT-5.4 reward  & $71.6$ & $3.08$ & $2.42$ \\
\textbf{\textsc{StudentSim} (Ours)}            & $\boldsymbol{90.5}$ & $\boldsymbol{3.31}$ & $\boldsymbol{3.93}$ \\
\bottomrule
\end{tabular}
\end{table}

\paragraph{Scope and interpretation.}
The comparison isolates the reward model: all three conditions share the tutor policy, the SFT start, and the GRPO setup, so the differences trace to the reward. The style and perception heads are only lightweight linear probes; they work because they read our trained simulator's own backbone, which already encodes student behavior and board state. A closed frontier-model API used as the simulator exposes no such backbone to probe, so it admits no comparable heads. The trained student simulator is thus the source of the signal in both roles: it produces the faithful, responsive move-level reward, and its backbone is what the heads extend to score the tutor's factual grounding and teaching style. The human-study gains across all three axes therefore follow from a single trained student simulator, evidence that a faithful, responsive simulator makes the reward practically effective for AI tutor optimization.

\section{Conclusion}
\label{sec:conclusion}

\textsc{StudentSim} is a training framework for per-student AI student simulation built around two requirements that must hold together: behavioral fidelity ($\mathcal{F}$), how well a simulator matches a student's own responses, and guidance responsiveness ($\mathcal{R}$), how reliably it updates towards where the tutor guidance leads. To measure both abilities on a common footing, \textsc{StudentSimEval} casts public learner corpora for research across chess, second-language English writing, and mathematics into a standardized per-student protocol on which any simulator, from our framework or a prior family, is fit and scored on the same held-out records. Under this protocol, \textsc{StudentSim}'s two-stage pooled-then-specialized pipeline yields a reference family of $60$ individualized simulators that are strong on both axes across all three domains, outperforming domain-specific state-tracking (weak on $\mathcal{R}$) and prompt-only LLM role-play (weak on $\mathcal{F}$). As a proof of concept that a trained simulator can also drive AI tutor improvement, a frozen \textsc{StudentSim} used as the reward in a chess tutor RL loop yields a tutor that expert humans rate as more accurate, better-guided, and more personalized than tutors trained with a frontier-LLM-simulator reward or the baseline model with no RL.

Looking ahead, behavioral fidelity and guidance responsiveness capture a student's state and its one-step update under guidance. Reproducing the fuller learning dynamics a student shows over time, how they acquire, retain, and forget knowledge across many student-tutor interactions or even through self-learning, is the deeper foundation of a simulator's value for AI tutor optimization and the direction \textsc{StudentSim} opens toward.

\section*{Acknowledgements}

We thank members of the \href{https://www.microsoft.com/en-us/research/group/deep-learning-group/}{Microsoft Research Deep Learning Group} for their helpful discussions and feedback during the early development of this work. We also thank members of the \href{https://timan.cs.illinois.edu/}{UIUC TIMAN Group} for their feedback and discussions as the project was being finalized. We are grateful to the contributors to the human studies for their time and effort. This work is supported in part by the National Science Foundation (NSF) and the Institute of Education Sciences (IES) under Grants 2229612 and 2433308.

\bibliographystyle{plainnat}
\bibliography{Sections/references}

@article{Corbett_1995, title={Knowledge tracing: Modeling the acquisition of procedural knowledge}, volume={4}, ISSN={1573-1391}, url={http://dx.doi.org/10.1007/bf01099821}, DOI={10.1007/bf01099821}, number={4}, journal={User Modelling and User-Adapted Interaction}, publisher={Springer Science and Business Media LLC}, author={Corbett, Albert T. and Anderson, John R.}, year={1995}, pages={253–278} }

@misc{piech2015deepknowledgetracing,
      title={Deep Knowledge Tracing}, 
      author={Chris Piech and Jonathan Spencer and Jonathan Huang and Surya Ganguli and Mehran Sahami and Leonidas Guibas and Jascha Sohl-Dickstein},
      year={2015},
      eprint={1506.05908},
      archivePrefix={arXiv},
      primaryClass={cs.AI},
      url={https://arxiv.org/abs/1506.05908}, 
}

@misc{ghosh2020contextawareattentiveknowledgetracing,
      title={Context-Aware Attentive Knowledge Tracing}, 
      author={Aritra Ghosh and Neil Heffernan and Andrew S. Lan},
      year={2020},
      eprint={2007.12324},
      archivePrefix={arXiv},
      primaryClass={cs.LG},
      url={https://arxiv.org/abs/2007.12324}, 
}

@misc{mcilroyyoung2020aligningsuperhumanaihuman,
      title={Aligning Superhuman AI with Human Behavior: Chess as a Model System}, 
      author={Reid McIlroy-Young and Siddhartha Sen and Jon Kleinberg and Ashton Anderson},
      year={2020},
      eprint={2006.01855},
      archivePrefix={arXiv},
      primaryClass={cs.AI},
      doi={https://doi.org/10.1145/3394486.3403219},
      url={https://arxiv.org/abs/2006.01855}, 
}

@misc{hu2021loralowrankadaptationlarge,
      title={LoRA: Low-Rank Adaptation of Large Language Models}, 
      author={Edward J. Hu and Yelong Shen and Phillip Wallis and Zeyuan Allen-Zhu and Yuanzhi Li and Shean Wang and Lu Wang and Weizhu Chen},
      year={2021},
      eprint={2106.09685},
      archivePrefix={arXiv},
      primaryClass={cs.CL},
      url={https://arxiv.org/abs/2106.09685}, 
}

@misc{chevalier2024languagemodelssciencetutors,
      title={Language Models as Science Tutors}, 
      author={Alexis Chevalier and Jiayi Geng and Alexander Wettig and Howard Chen and Sebastian Mizera and Toni Annala and Max Jameson Aragon and Arturo Rodríguez Fanlo and Simon Frieder and Simon Machado and Akshara Prabhakar and Ellie Thieu and Jiachen T. Wang and Zirui Wang and Xindi Wu and Mengzhou Xia and Wenhan Xia and Jiatong Yu and Jun-Jie Zhu and Zhiyong Jason Ren and Sanjeev Arora and Danqi Chen},
      year={2024},
      eprint={2402.11111},
      archivePrefix={arXiv},
      primaryClass={cs.CL},
      url={https://arxiv.org/abs/2402.11111}, 
}

@misc{owoicho2023exploitingsimulateduserfeedback,
      title={Exploiting Simulated User Feedback for Conversational Search: Ranking, Rewriting, and Beyond}, 
      author={Paul Owoicho and Ivan Sekulić and Mohammad Aliannejadi and Jeffrey Dalton and Fabio Crestani},
      year={2023},
      eprint={2304.13874},
      archivePrefix={arXiv},
      primaryClass={cs.IR},
      doi={https://doi.org/10.1145/3539618.3591683},
      url={https://arxiv.org/abs/2304.13874}, 
}

@misc{terragni2023incontextlearningusersimulators,
      title={In-Context Learning User Simulators for Task-Oriented Dialog Systems}, 
      author={Silvia Terragni and Modestas Filipavicius and Nghia Khau and Bruna Guedes and André Manso and Roland Mathis},
      year={2023},
      eprint={2306.00774},
      archivePrefix={arXiv},
      primaryClass={cs.CL},
      url={https://arxiv.org/abs/2306.00774}, 
}

@misc{zheng2023judgingllmasajudgemtbenchchatbot,
      title={Judging LLM-as-a-Judge with MT-Bench and Chatbot Arena}, 
      author={Lianmin Zheng and Wei-Lin Chiang and Ying Sheng and Siyuan Zhuang and Zhanghao Wu and Yonghao Zhuang and Zi Lin and Zhuohan Li and Dacheng Li and Eric P. Xing and Hao Zhang and Joseph E. Gonzalez and Ion Stoica},
      year={2023},
      eprint={2306.05685},
      archivePrefix={arXiv},
      primaryClass={cs.CL},
      url={https://arxiv.org/abs/2306.05685}, 
}

@misc{tang2024maia2unifiedmodelhumanai,
      title={Maia-2: A Unified Model for Human-AI Alignment in Chess}, 
      author={Zhenwei Tang and Difan Jiao and Reid McIlroy-Young and Jon Kleinberg and Siddhartha Sen and Ashton Anderson},
      year={2024},
      eprint={2409.20553},
      archivePrefix={arXiv},
      primaryClass={cs.AI},
      url={https://arxiv.org/abs/2409.20553}, 
}

@misc{yang2025qwen3technicalreport,
      title={Qwen3 Technical Report}, 
      author={An Yang and Anfeng Li and Baosong Yang and Beichen Zhang and Binyuan Hui and Bo Zheng and Bowen Yu and Chang Gao and Chengen Huang and Chenxu Lv and Chujie Zheng and Dayiheng Liu and Fan Zhou and Fei Huang and Feng Hu and Hao Ge and Haoran Wei and Huan Lin and Jialong Tang and Jian Yang and Jianhong Tu and Jianwei Zhang and Jianxin Yang and Jiaxi Yang and Jing Zhou and Jingren Zhou and Junyang Lin and Kai Dang and Keqin Bao and Kexin Yang and Le Yu and Lianghao Deng and Mei Li and Mingfeng Xue and Mingze Li and Pei Zhang and Peng Wang and Qin Zhu and Rui Men and Ruize Gao and Shixuan Liu and Shuang Luo and Tianhao Li and Tianyi Tang and Wenbiao Yin and Xingzhang Ren and Xinyu Wang and Xinyu Zhang and Xuancheng Ren and Yang Fan and Yang Su and Yichang Zhang and Yinger Zhang and Yu Wan and Yuqiong Liu and Zekun Wang and Zeyu Cui and Zhenru Zhang and Zhipeng Zhou and Zihan Qiu},
      year={2025},
      eprint={2505.09388},
      archivePrefix={arXiv},
      primaryClass={cs.CL},
      url={https://arxiv.org/abs/2505.09388},
}

@misc{bai2025qwen3vltechnicalreport,
      title={Qwen3-VL Technical Report},
      author={Shuai Bai and Yuxuan Cai and Ruizhe Chen and Keqin Chen and Xionghui Chen and Zesen Cheng and Lianghao Deng and Wei Ding and Chang Gao and Chunjiang Ge and Wenbin Ge and Zhifang Guo and Qidong Huang and Jie Huang and Fei Huang and Binyuan Hui and Shutong Jiang and Zhaohai Li and Mingsheng Li and Mei Li and Kaixin Li and Zicheng Lin and Junyang Lin and Xuejing Liu and Jiawei Liu and Chenglong Liu and Yang Liu and Dayiheng Liu and Shixuan Liu and Dunjie Lu and Ruilin Luo and Chenxu Lv and Rui Men and Lingchen Meng and Xuancheng Ren and Xingzhang Ren and Sibo Song and Yuchong Sun and Jun Tang and Jianhong Tu and Jianqiang Wan and Peng Wang and Pengfei Wang and Qiuyue Wang and Yuxuan Wang and Tianbao Xie and Yiheng Xu and Haiyang Xu and Jin Xu and Zhibo Yang and Mingkun Yang and Jianxin Yang and An Yang and Bowen Yu and Fei Zhang and Hang Zhang and Xi Zhang and Bo Zheng and Humen Zhong and Jingren Zhou and Fan Zhou and Jing Zhou and Yuanzhi Zhu and Ke Zhu},
      year={2025},
      eprint={2511.21631},
      archivePrefix={arXiv},
      primaryClass={cs.CV},
      url={https://arxiv.org/abs/2511.21631},
}

@misc{worden2026foundationalassisteducationaldatasetfoundational,
      title={FoundationalASSIST: An Educational Dataset for Foundational Knowledge Tracing and Pedagogical Grounding of LLMs}, 
      author={Eamon Worden and Cristina Heffernan and Neil Heffernan and Shashank Sonkar},
      year={2026},
      eprint={2602.00070},
      archivePrefix={arXiv},
      primaryClass={cs.CY},
      url={https://arxiv.org/abs/2602.00070}, 
}

@inproceedings{geertzen2013automatic,
  title={Automatic linguistic annotation of large scale L2 databases: The EF-Cambridge Open Language Database (EFCAMDAT)},
  author={Geertzen, Jeroen and Alexopoulou, Theodora and Korhonen, Anna and others},
  booktitle={Proceedings of the 31st Second Language Research Forum. Somerville, MA: Cascadilla Proceedings Project},
  pages={240--254},
  year={2013}
}

@article{Shatz_2020, title={Refining and modifying the EFCAMDAT: Lessons from creating a new corpus from an existing large-scale English learner language database}, volume={6}, ISSN={2215-1486}, url={http://dx.doi.org/10.1075/ijlcr.20009.sha}, DOI={10.1075/ijlcr.20009.sha}, number={2}, journal={International Journal of Learner Corpus Research}, publisher={John Benjamins Publishing Company}, author={Shatz, Itamar}, year={2020}, month=Dec, pages={220–236} }

@article{Oksuz_2025, title={The influence of L1 typology on the acquisition of the L2 English article: A large-scale corpus study}, volume={42}, ISSN={1477-0326}, url={http://dx.doi.org/10.1177/02676583251395876}, DOI={10.1177/02676583251395876}, number={2}, journal={Second Language Research}, publisher={SAGE Publications}, author={Öksüz, Doğuş and Alexopoulou, Theodora and Derkach, Kateryna and Tsimpli, Ianthi Maria}, year={2025}, month=Dec, pages={281–311} }

@misc{lichess,
  title        = {{Lichess} open database: standard rated games (May 2025 snapshot)},
  author       = {{Lichess}},
  year         = {2025},
  howpublished = {\url{https://database.lichess.org/standard/lichess_db_standard_rated_2025-05.pgn.zst}},
  note         = {Database portal: \url{https://database.lichess.org/}. Released under Creative Commons CC0 1.0.}
}

@misc{stockfish,
  title        = {Stockfish},
  author       = {{The Stockfish developers}},
  year         = {2025},
  howpublished = {\url{https://stockfishchess.org/}},
  note         = {Source repository: \url{https://github.com/official-stockfish/Stockfish}.
                  Recommended citation per project CITATION.cff at
                  \url{https://github.com/official-stockfish/Stockfish/blob/master/CITATION.cff}.}
}

@article{Zhao_2025, title={SWIFT: A Scalable Lightweight Infrastructure for Fine-Tuning}, volume={39}, ISSN={2159-5399}, url={http://dx.doi.org/10.1609/aaai.v39i28.35383}, DOI={10.1609/aaai.v39i28.35383}, number={28}, journal={Proceedings of the AAAI Conference on Artificial Intelligence}, publisher={Association for the Advancement of Artificial Intelligence (AAAI)}, author={Zhao, Yuze and Huang, Jintao and Hu, Jinghan and Wang, Xingjun and Mao, Yunlin and Zhang, Daoze and Jiang, Zeyinzi and Wu, Zhikai and Ai, Baole and Wang, Ang and Zhou, Wenmeng and Chen, Yingda}, year={2025}, month=Apr, pages={29733–29735} }

@misc{sheng2024hybridflowflexibleefficientrlhf,
      title={HybridFlow: A Flexible and Efficient RLHF Framework}, 
      author={Guangming Sheng and Chi Zhang and Zilingfeng Ye and Xibin Wu and Wang Zhang and Ru Zhang and Yanghua Peng and Haibin Lin and Chuan Wu},
      year={2024},
      eprint={2409.19256},
      archivePrefix={arXiv},
      primaryClass={cs.LG},
      doi={https://doi.org/10.1145/3689031.3696075},
      url={https://arxiv.org/abs/2409.19256}, 
}

@inproceedings{Zhang_2017, series={WWW ’17}, title={Dynamic Key-Value Memory Networks for Knowledge Tracing}, url={http://dx.doi.org/10.1145/3038912.3052580}, DOI={10.1145/3038912.3052580}, booktitle={Proceedings of the 26th International Conference on World Wide Web}, publisher={International World Wide Web Conferences Steering Committee}, author={Zhang, Jiani and Shi, Xingjian and King, Irwin and Yeung, Dit-Yan}, year={2017}, month=Apr, pages={765–774}, collection={WWW ’17} }

@misc{yuan2026validstudentsimulationlarge,
      title={Towards Valid Student Simulation with Large Language Models}, 
      author={Zhihao Yuan and Yunze Xiao and Ming Li and Weihao Xuan and Richard Tong and Mona Diab and Tom Mitchell},
      year={2026},
      eprint={2601.05473},
      archivePrefix={arXiv},
      primaryClass={cs.CL},
      url={https://arxiv.org/abs/2601.05473}, 
}

@misc{scarlatos2026simulatedstudentstutoringdialogues,
      title={Simulated Students in Tutoring Dialogues: Substance or Illusion?}, 
      author={Alexander Scarlatos and Jaewook Lee and Simon Woodhead and Andrew Lan},
      year={2026},
      eprint={2601.04025},
      archivePrefix={arXiv},
      primaryClass={cs.CL},
      url={https://arxiv.org/abs/2601.04025}, 
}

@inproceedings{Wu_2025, title={Embracing Imperfection: Simulating Students with Diverse Cognitive Levels Using LLM-based Agents}, url={http://dx.doi.org/10.18653/v1/2025.acl-long.488}, DOI={10.18653/v1/2025.acl-long.488}, booktitle={Proceedings of the 63rd Annual Meeting of the Association for Computational Linguistics (Volume 1: Long Papers)}, publisher={Association for Computational Linguistics}, author={Wu, Tao and Chen, Jingyuan and Lin, Wang and Li, Mengze and Zhu, Yumeng and Li, Ang and Kuang, Kun and Wu, Fei}, year={2025}, pages={9887–9908} }

@inproceedings{Martynova_2025, title={Can LLMs Effectively Simulate Human Learners? Teachers’ Insights from Tutoring LLM Students}, url={http://dx.doi.org/10.18653/v1/2025.bea-1.8}, DOI={10.18653/v1/2025.bea-1.8}, booktitle={Proceedings of the 20th Workshop on Innovative Use of NLP for Building Educational Applications (BEA 2025)}, publisher={Association for Computational Linguistics}, author={Martynova, Daria and Macina, Jakub and Daheim, Nico and Yalcin, Nilay and Zhang, Xiaoyu and Sachan, Mrinmaya}, year={2025}, pages={100–117} }

@misc{mannekote2024llmsreliablysimulatehuman,
      title={Can LLMs Reliably Simulate Human Learner Actions? A Simulation Authoring Framework for Open-Ended Learning Environments}, 
      author={Amogh Mannekote and Adam Davies and Jina Kang and Kristy Elizabeth Boyer},
      year={2024},
      eprint={2410.02110},
      archivePrefix={arXiv},
      primaryClass={cs.AI},
      url={https://arxiv.org/abs/2410.02110}, 
}

@misc{macina2023mathdialdialoguetutoringdataset,
      title={MathDial: A Dialogue Tutoring Dataset with Rich Pedagogical Properties Grounded in Math Reasoning Problems}, 
      author={Jakub Macina and Nico Daheim and Sankalan Pal Chowdhury and Tanmay Sinha and Manu Kapur and Iryna Gurevych and Mrinmaya Sachan},
      year={2023},
      eprint={2305.14536},
      archivePrefix={arXiv},
      primaryClass={cs.CL},
      url={https://arxiv.org/abs/2305.14536}, 
}

@misc{xu2024eduagentgenerativestudentagents,
      title={EduAgent: Generative Student Agents in Learning}, 
      author={Songlin Xu and Xinyu Zhang and Lianhui Qin},
      year={2024},
      eprint={2404.07963},
      archivePrefix={arXiv},
      primaryClass={cs.CY},
      url={https://arxiv.org/abs/2404.07963}, 
}

@misc{zhang2024simulatingclassroomeducationllmempowered,
      title={Simulating Classroom Education with LLM-Empowered Agents}, 
      author={Zheyuan Zhang and Daniel Zhang-Li and Jifan Yu and Linlu Gong and Jinchang Zhou and Zhanxin Hao and Jianxiao Jiang and Jie Cao and Huiqin Liu and Zhiyuan Liu and Lei Hou and Juanzi Li},
      year={2024},
      eprint={2406.19226},
      archivePrefix={arXiv},
      primaryClass={cs.CL},
      url={https://arxiv.org/abs/2406.19226}, 
}

@misc{gao2026agent4edugeneratinglearnerresponse,
      title={Agent4Edu: Generating Learner Response Data by Generative Agents for Intelligent Education Systems}, 
      author={Weibo Gao and Qi Liu and Linan Yue and Fangzhou Yao and Rui Lv and Zheng Zhang and Hao Wang and Zhenya Huang},
      year={2026},
      eprint={2501.10332},
      archivePrefix={arXiv},
      primaryClass={cs.CY},
      url={https://arxiv.org/abs/2501.10332}, 
}

@misc{xu2025classroomsimulacrabuildingcontextual,
      title={Classroom Simulacra: Building Contextual Student Generative Agents in Online Education for Learning Behavioral Simulation}, 
      author={Songlin Xu and Hao-Ning Wen and Hongyi Pan and Dallas Dominguez and Dongyin Hu and Xinyu Zhang},
      year={2025},
      eprint={2502.02780},
      archivePrefix={arXiv},
      primaryClass={cs.HC},
      doi={https://doi.org/10.1145/3706598.3713773},
      url={https://arxiv.org/abs/2502.02780}, 
}

@misc{louie2024roleplaydohenablingdomainexpertscreate,
      title={Roleplay-doh: Enabling Domain-Experts to Create LLM-simulated Patients via Eliciting and Adhering to Principles}, 
      author={Ryan Louie and Ananjan Nandi and William Fang and Cheng Chang and Emma Brunskill and Diyi Yang},
      year={2024},
      eprint={2407.00870},
      archivePrefix={arXiv},
      primaryClass={cs.CL},
      url={https://arxiv.org/abs/2407.00870}, 
}

@article{Markel_2023, title={GPTeach: Interactive TA Training with GPT-based Students}, url={http://dx.doi.org/10.35542/osf.io/r23bu}, DOI={10.35542/osf.io/r23bu}, publisher={Center for Open Science}, author={Markel, Julia Mae and Opferman, Steven G. and Landay, James A. and Piech, Chris}, year={2023}, month=Feb }

@misc{openai2024gpt4o,
      title={{GPT-4o System Card}},
      author={{OpenAI}},
      year={2024},
      month={aug},
      howpublished={\url{https://openai.com/index/gpt-4o-system-card/}},
}

@misc{openai2026gpt54,
      title={{GPT-5.4 Thinking System Card}},
      author={{OpenAI}},
      year={2026},
      month={mar},
      howpublished={\url{https://openai.com/index/gpt-5-4-thinking-system-card/}},
}

@misc{lu2024generativestudentsusingllmsimulated,
      title={Generative Students: Using LLM-Simulated Student Profiles to Support Question Item Evaluation}, 
      author={Xinyi Lu and Xu Wang},
      year={2024},
      eprint={2405.11591},
      archivePrefix={arXiv},
      primaryClass={cs.HC},
      doi={https://doi.org/10.1145/3657604.3662031},
      url={https://arxiv.org/abs/2405.11591}, 
}

@misc{peng2024quantifyingoptimizingglobalfaithfulness,
      title={Quantifying and Optimizing Global Faithfulness in Persona-driven Role-playing}, 
      author={Letian Peng and Jingbo Shang},
      year={2024},
      eprint={2405.07726},
      archivePrefix={arXiv},
      primaryClass={cs.CL},
      url={https://arxiv.org/abs/2405.07726}, 
}

@misc{samuel2025personagymevaluatingpersonaagents,
      title={PersonaGym: Evaluating Persona Agents and LLMs}, 
      author={Vinay Samuel and Henry Peng Zou and Yue Zhou and Shreyas Chaudhari and Ashwin Kalyan and Tanmay Rajpurohit and Ameet Deshpande and Karthik Narasimhan and Vishvak Murahari},
      year={2025},
      eprint={2407.18416},
      archivePrefix={arXiv},
      primaryClass={cs.CL},
      url={https://arxiv.org/abs/2407.18416}, 
}

@misc{wang2024incharacterevaluatingpersonalityfidelity,
      title={InCharacter: Evaluating Personality Fidelity in Role-Playing Agents through Psychological Interviews}, 
      author={Xintao Wang and Yunze Xiao and Jen-tse Huang and Siyu Yuan and Rui Xu and Haoran Guo and Quan Tu and Yaying Fei and Ziang Leng and Wei Wang and Jiangjie Chen and Cheng Li and Yanghua Xiao},
      year={2024},
      eprint={2310.17976},
      archivePrefix={arXiv},
      primaryClass={cs.CL},
      url={https://arxiv.org/abs/2310.17976}, 
}

@misc{scarlatos2025trainingllmbasedtutorsimprove,
      title={Training LLM-based Tutors to Improve Student Learning Outcomes in Dialogues}, 
      author={Alexander Scarlatos and Naiming Liu and Jaewook Lee and Richard Baraniuk and Andrew Lan},
      year={2025},
      eprint={2503.06424},
      archivePrefix={arXiv},
      primaryClass={cs.CL},
      doi={https://doi.org/10.1007/978-3-031-98414-3_18},
      url={https://arxiv.org/abs/2503.06424}, 
}

@misc{dinucujianu2025problemsolvingteachingproblemsolvingaligning,
      title={From Problem-Solving to Teaching Problem-Solving: Aligning LLMs with Pedagogy using Reinforcement Learning}, 
      author={David Dinucu-Jianu and Jakub Macina and Nico Daheim and Ido Hakimi and Iryna Gurevych and Mrinmaya Sachan},
      year={2025},
      eprint={2505.15607},
      archivePrefix={arXiv},
      primaryClass={cs.CL},
      url={https://arxiv.org/abs/2505.15607}, 
}

@misc{macina2025mathtutorbenchbenchmarkmeasuringopenended,
      title={MathTutorBench: A Benchmark for Measuring Open-ended Pedagogical Capabilities of LLM Tutors}, 
      author={Jakub Macina and Nico Daheim and Ido Hakimi and Manu Kapur and Iryna Gurevych and Mrinmaya Sachan},
      year={2025},
      eprint={2502.18940},
      archivePrefix={arXiv},
      primaryClass={cs.CL},
      url={https://arxiv.org/abs/2502.18940}, 
}

@misc{maurya2025unifyingaitutorevaluation,
      title={Unifying AI Tutor Evaluation: An Evaluation Taxonomy for Pedagogical Ability Assessment of LLM-Powered AI Tutors}, 
      author={Kaushal Kumar Maurya and KV Aditya Srivatsa and Kseniia Petukhova and Ekaterina Kochmar},
      year={2025},
      eprint={2412.09416},
      archivePrefix={arXiv},
      primaryClass={cs.CL},
      url={https://arxiv.org/abs/2412.09416}, 
}

@misc{wang2025tutorcopilothumanaiapproach,
      title={Tutor CoPilot: A Human-AI Approach for Scaling Real-Time Expertise}, 
      author={Rose E. Wang and Ana T. Ribeiro and Carly D. Robinson and Susanna Loeb and Dora Demszky},
      year={2025},
      eprint={2410.03017},
      archivePrefix={arXiv},
      primaryClass={cs.CL},
      url={https://arxiv.org/abs/2410.03017}, 
}

@misc{wang2024bridgingnoviceexpertgapmodels,
      title={Bridging the Novice-Expert Gap via Models of Decision-Making: A Case Study on Remediating Math Mistakes}, 
      author={Rose E. Wang and Qingyang Zhang and Carly Robinson and Susanna Loeb and Dorottya Demszky},
      year={2024},
      eprint={2310.10648},
      archivePrefix={arXiv},
      primaryClass={cs.CL},
      url={https://arxiv.org/abs/2310.10648}, 
}

@misc{learnlmteam2025learnlmimprovinggeminilearning,
      title={LearnLM: Improving Gemini for Learning}, 
      author={LearnLM Team and Abhinit Modi and Aditya Srikanth Veerubhotla and Aliya Rysbek and Andrea Huber and Brett Wiltshire and Brian Veprek and Daniel Gillick and Daniel Kasenberg and Derek Ahmed and Irina Jurenka and James Cohan and Jennifer She and Julia Wilkowski and Kaiz Alarakyia and Kevin R. McKee and Lisa Wang and Markus Kunesch and Mike Schaekermann and Miruna Pîslar and Nikhil Joshi and Parsa Mahmoudieh and Paul Jhun and Sara Wiltberger and Shakir Mohamed and Shashank Agarwal and Shubham Milind Phal and Sun Jae Lee and Theofilos Strinopoulos and Wei-Jen Ko and Amy Wang and Ankit Anand and Avishkar Bhoopchand and Dan Wild and Divya Pandya and Filip Bar and Garth Graham and Holger Winnemoeller and Mahvish Nagda and Prateek Kolhar and Renee Schneider and Shaojian Zhu and Stephanie Chan and Steve Yadlowsky and Viknesh Sounderajah and Yannis Assael},
      year={2025},
      eprint={2412.16429},
      archivePrefix={arXiv},
      primaryClass={cs.CY},
      url={https://arxiv.org/abs/2412.16429}, 
}

@inproceedings{Wang_2024, title={RoleLLM: Benchmarking, Eliciting, and Enhancing Role-Playing Abilities of Large Language Models}, url={http://dx.doi.org/10.18653/v1/2024.findings-acl.878}, DOI={10.18653/v1/2024.findings-acl.878}, booktitle={Findings of the Association for Computational Linguistics ACL 2024}, publisher={Association for Computational Linguistics}, author={Wang, Noah and Peng, Z.y. and Que, Haoran and Liu, Jiaheng and Zhou, Wangchunshu and Wu, Yuhan and Guo, Hongcheng and Gan, Ruitong and Ni, Zehao and Yang, Jian and Zhang, Man and Zhang, Zhaoxiang and Ouyang, Wanli and Xu, Ke and Huang, Wenhao and Fu, Jie and Peng, Junran}, year={2024}, pages={14743–14777} }

@inproceedings{Scarlatos_2025, series={LAK 2025}, title={Exploring Knowledge Tracing in Tutor-Student Dialogues using LLMs}, url={http://dx.doi.org/10.1145/3706468.3706501}, DOI={10.1145/3706468.3706501}, booktitle={Proceedings of the 15th International Learning Analytics and Knowledge Conference}, publisher={ACM}, author={Scarlatos, Alexander and Baker, Ryan S. and Lan, Andrew}, year={2025}, month=Mar, pages={249–259}, collection={LAK 2025} }

@inproceedings{Stasaski_2020, title={CIMA: A Large Open Access Dialogue Dataset for Tutoring}, url={http://dx.doi.org/10.18653/v1/2020.bea-1.5}, DOI={10.18653/v1/2020.bea-1.5}, booktitle={Proceedings of the Fifteenth Workshop on Innovative Use of NLP for Building Educational Applications}, publisher={Association for Computational Linguistics}, author={Stasaski, Katherine and Kao, Kimberly and Hearst, Marti A.}, year={2020}, pages={52–64} }

@book{Lord_2012, title={Applications of Item Response Theory To Practical Testing Problems}, ISBN={9781136557248}, url={http://dx.doi.org/10.4324/9780203056615}, DOI={10.4324/9780203056615}, publisher={Routledge}, author={Lord, F. M.}, year={2012}, month=Nov }

@misc{gusev2025pingpongbenchmarkroleplayinglanguage,
      title={PingPong: A Benchmark for Role-Playing Language Models with User Emulation and Multi-Model Evaluation}, 
      author={Ilya Gusev},
      year={2025},
      eprint={2409.06820},
      archivePrefix={arXiv},
      primaryClass={cs.CL},
      url={https://arxiv.org/abs/2409.06820}, 
}

@inproceedings{Choi_2020, series={L@S ’20}, title={Towards an Appropriate Query, Key, and Value Computation for Knowledge Tracing}, url={http://dx.doi.org/10.1145/3386527.3405945}, DOI={10.1145/3386527.3405945}, booktitle={Proceedings of the Seventh ACM Conference on Learning @ Scale}, publisher={ACM}, author={Choi, Youngduck and Lee, Youngnam and Cho, Junghyun and Baek, Jineon and Kim, Byungsoo and Cha, Yeongmin and Shin, Dongmin and Bae, Chan and Heo, Jaewe}, year={2020}, month=Aug, pages={341–344}, collection={L@S ’20} }

@inproceedings{Tu_2024, title={CharacterEval: A Chinese Benchmark for Role-Playing Conversational Agent Evaluation}, url={http://dx.doi.org/10.18653/v1/2024.acl-long.638}, DOI={10.18653/v1/2024.acl-long.638}, booktitle={Proceedings of the 62nd Annual Meeting of the Association for Computational Linguistics (Volume 1: Long Papers)}, publisher={Association for Computational Linguistics}, author={Tu, Quan and Fan, Shilong and Tian, Zihang and Shen, Tianhao and Shang, Shuo and Gao, Xin and Yan, Rui}, year={2024}, pages={11836–11850} }

@article{liu2024socraticlm,
  title={SocraticLM: exploring socratic personalized teaching with large language models},
  author={Liu, Jiayu and Huang, Zhenya and Xiao, Tong and Sha, Jing and Wu, Jinze and Liu, Qi and Wang, Shijin and Chen, Enhong},
  journal={Advances in Neural Information Processing Systems},
  volume={37},
  pages={85693--85721},
  year={2024}
}

@inproceedings{Park_2023, series={UIST ’23}, title={Generative Agents: Interactive Simulacra of Human Behavior}, url={http://dx.doi.org/10.1145/3586183.3606763}, DOI={10.1145/3586183.3606763}, booktitle={Proceedings of the 36th Annual ACM Symposium on User Interface Software and Technology}, publisher={ACM}, author={Park, Joon Sung and O’Brien, Joseph and Cai, Carrie Jun and Morris, Meredith Ringel and Liang, Percy and Bernstein, Michael S.}, year={2023}, month=Oct, pages={1–22}, collection={UIST ’23} }

@inbook{Cen_2006, title={Learning Factors Analysis – A General Method for Cognitive Model Evaluation and Improvement}, ISBN={9783540351603}, ISSN={1611-3349}, url={http://dx.doi.org/10.1007/11774303_17}, DOI={10.1007/11774303_17}, booktitle={Intelligent Tutoring Systems}, publisher={Springer Berlin Heidelberg}, author={Cen, Hao and Koedinger, Kenneth and Junker, Brian}, year={2006}, pages={164–175} }

@article{Reckase_1985, title={The Difficulty of Test Items That Measure More Than One Ability}, volume={9}, ISSN={1552-3497}, url={http://dx.doi.org/10.1177/014662168500900409}, DOI={10.1177/014662168500900409}, number={4}, journal={Applied Psychological Measurement}, publisher={SAGE Publications}, author={Reckase, Mark D.}, year={1985}, month=Dec, pages={401–412} }

@inproceedings{Liu_2024, title={Personality-aware Student Simulation for Conversational Intelligent Tutoring Systems}, url={http://dx.doi.org/10.18653/v1/2024.emnlp-main.37}, DOI={10.18653/v1/2024.emnlp-main.37}, booktitle={Proceedings of the 2024 Conference on Empirical Methods in Natural Language Processing}, publisher={Association for Computational Linguistics}, author={Liu, Zhengyuan and Yin, Stella Xin and Lin, Geyu and Chen, Nancy F.}, year={2024}, pages={626–642} }

@article{Nye_2014, title={AutoTutor and Family: A Review of 17 Years of Natural Language Tutoring}, volume={24}, ISSN={1560-4292}, url={http://dx.doi.org/10.1007/s40593-014-0029-5}, DOI={10.1007/s40593-014-0029-5}, number={4}, journal={International Journal of Artificial Intelligence in Education}, publisher={Elsevier BV}, author={Nye, Benjamin D. and Graesser, Arthur C. and Hu, Xiangen}, year={2014}, month=Dec, pages={427–469} }

@inbook{Pavlik_Philip_I__2009, title={Performance Factors Analysis \&amp;ndash; A New Alternative to Knowledge Tracing}, ISSN={0922-6389}, url={http://dx.doi.org/10.3233/978-1-60750-028-5-531}, DOI={10.3233/978-1-60750-028-5-531}, booktitle={Artificial Intelligence in Education}, publisher={IOS Press}, author={Pavlik Philip I. and Cen Hao and Koedinger Kenneth R.}, year={2009} }

@inproceedings{Shao_2023, title={Character-LLM: A Trainable Agent for Role-Playing}, url={http://dx.doi.org/10.18653/v1/2023.emnlp-main.814}, DOI={10.18653/v1/2023.emnlp-main.814}, booktitle={Proceedings of the 2023 Conference on Empirical Methods in Natural Language Processing}, publisher={Association for Computational Linguistics}, author={Shao, Yunfan and Li, Linyang and Dai, Junqi and Qiu, Xipeng}, year={2023}, pages={13153–13187} }

@inproceedings{Jiang_2024, title={PersonaLLM: Investigating the Ability of Large Language Models to Express Personality Traits}, url={http://dx.doi.org/10.18653/v1/2024.findings-naacl.229}, DOI={10.18653/v1/2024.findings-naacl.229}, booktitle={Findings of the Association for Computational Linguistics: NAACL 2024}, publisher={Association for Computational Linguistics}, author={Jiang, Hang and Zhang, Xiajie and Cao, Xubo and Breazeal, Cynthia and Roy, Deb and Kabbara, Jad}, year={2024}, pages={3605–3627} }

@inproceedings{Liu_2022, title={Open-ended Knowledge Tracing for Computer Science Education}, url={http://dx.doi.org/10.18653/v1/2022.emnlp-main.254}, DOI={10.18653/v1/2022.emnlp-main.254}, booktitle={Proceedings of the 2022 Conference on Empirical Methods in Natural Language Processing}, publisher={Association for Computational Linguistics}, author={Liu, Naiming and Wang, Zichao and Baraniuk, Richard and Lan, Andrew}, year={2022}, pages={3849–3862} }

@misc{pandey2019selfattentivemodelknowledgetracing,
      title={A Self-Attentive model for Knowledge Tracing}, 
      author={Shalini Pandey and George Karypis},
      year={2019},
      eprint={1907.06837},
      archivePrefix={arXiv},
      primaryClass={cs.LG},
      url={https://arxiv.org/abs/1907.06837}, 
}

@misc{park2026llmagentsgroundedselfreports,
      title={LLM Agents Grounded in Self-Reports Enable General-Purpose Simulation of Individuals}, 
      author={Joon Sung Park and Carolyn Q. Zou and Jonne Kamphorst and Niles Egan and Aaron Shaw and Benjamin Mako Hill and Carrie Cai and Meredith Ringel Morris and Percy Liang and Robb Willer and Michael S. Bernstein},
      year={2026},
      eprint={2411.10109},
      archivePrefix={arXiv},
      primaryClass={cs.AI},
      url={https://arxiv.org/abs/2411.10109}, 
}

@inproceedings{Liu_2023, series={WWW ’23}, title={Enhancing Deep Knowledge Tracing with Auxiliary Tasks}, url={http://dx.doi.org/10.1145/3543507.3583866}, DOI={10.1145/3543507.3583866}, booktitle={Proceedings of the ACM Web Conference 2023}, publisher={ACM}, author={Liu, Zitao and Liu, Qiongqiong and Chen, Jiahao and Huang, Shuyan and Gao, Boyu and Luo, Weiqi and Weng, Jian}, year={2023}, month=Apr, pages={4178–4187}, collection={WWW ’23} }

@misc{luo2024duetsimbuildingusersimulator,
      title={DuetSim: Building User Simulator with Dual Large Language Models for Task-Oriented Dialogues}, 
      author={Xiang Luo and Zhiwen Tang and Jin Wang and Xuejie Zhang},
      year={2024},
      eprint={2405.13028},
      archivePrefix={arXiv},
      primaryClass={cs.CL},
      url={https://arxiv.org/abs/2405.13028}, 
}

@inproceedings{Sonkar_2023, title={CLASS: A Design Framework for Building Intelligent Tutoring Systems Based on Learning Science principles}, url={http://dx.doi.org/10.18653/v1/2023.findings-emnlp.130}, DOI={10.18653/v1/2023.findings-emnlp.130}, booktitle={Findings of the Association for Computational Linguistics: EMNLP 2023}, publisher={Association for Computational Linguistics}, author={Sonkar, Shashank and Liu, Naiming and Mallick, Debshila and Baraniuk, Richard}, year={2023}, pages={1941–1961} }

@article{Cheng_2025, title={Uncertainty-aware Knowledge Tracing}, volume={39}, ISSN={2159-5399}, url={http://dx.doi.org/10.1609/aaai.v39i27.35007}, DOI={10.1609/aaai.v39i27.35007}, number={27}, journal={Proceedings of the AAAI Conference on Artificial Intelligence}, publisher={Association for the Advancement of Artificial Intelligence (AAAI)}, author={Cheng, Weihua and Du, Hanwen and Li, Chunxiao and Ni, Ersheng and Tan, Liangdi and Xu, Tianqi and Ni, Yongxin}, year={2025}, month=Apr, pages={27905–27913} }

@misc{lee2024languagemodelknowledgetracing,
      title={Language Model Can Do Knowledge Tracing: Simple but Effective Method to Integrate Language Model and Knowledge Tracing Task}, 
      author={Unggi Lee and Jiyeong Bae and Dohee Kim and Sookbun Lee and Jaekwon Park and Taekyung Ahn and Gunho Lee and Damji Stratton and Hyeoncheol Kim},
      year={2024},
      eprint={2406.02893},
      archivePrefix={arXiv},
      primaryClass={cs.CL},
      url={https://arxiv.org/abs/2406.02893}, 
}

@inbook{Ghosh_2021, title={Option Tracing: Beyond Correctness Analysis in Knowledge Tracing}, ISBN={9783030782924}, ISSN={1611-3349}, url={http://dx.doi.org/10.1007/978-3-030-78292-4_12}, DOI={10.1007/978-3-030-78292-4_12}, booktitle={Artificial Intelligence in Education}, publisher={Springer International Publishing}, author={Ghosh, Aritra and Raspat, Jay and Lan, Andrew}, year={2021}, pages={137–149} }

@misc{shao2024deepseekmathpushinglimitsmathematical,
      title={DeepSeekMath: Pushing the Limits of Mathematical Reasoning in Open Language Models},
      author={Zhihong Shao and Peiyi Wang and Qihao Zhu and Runxin Xu and Junxiao Song and Xiao Bi and Haowei Zhang and Mingchuan Zhang and Y. K. Li and Y. Wu and Daya Guo},
      year={2024},
      eprint={2402.03300},
      archivePrefix={arXiv},
      primaryClass={cs.CL},
      url={https://arxiv.org/abs/2402.03300},
}

@inproceedings{Schein_2002, series={SIGIR02}, title={Methods and metrics for cold-start recommendations}, url={http://dx.doi.org/10.1145/564376.564421}, DOI={10.1145/564376.564421}, booktitle={Proceedings of the 25th annual international ACM SIGIR conference on Research and development in information retrieval}, publisher={ACM}, author={Schein, Andrew I. and Popescul, Alexandrin and Ungar, Lyle H. and Pennock, David M.}, year={2002}, month=Aug, pages={253–260}, collection={SIGIR02} }

@misc{lee2019melumetalearneduserpreference,
      title={MeLU: Meta-Learned User Preference Estimator for Cold-Start Recommendation}, 
      author={Hoyeop Lee and Jinbae Im and Seongwon Jang and Hyunsouk Cho and Sehee Chung},
      year={2019},
      eprint={1908.00413},
      archivePrefix={arXiv},
      primaryClass={cs.IR},
      url={https://arxiv.org/abs/1908.00413}, 
}

@misc{bhattacharjee2025coldstartproblemexperimental,
      title={Cold Start Problem: An Experimental Study of Knowledge Tracing Models with New Students}, 
      author={Indronil Bhattacharjee and Christabel Wayllace},
      year={2025},
      eprint={2505.21517},
      archivePrefix={arXiv},
      primaryClass={cs.CY},
      doi={https://doi.org/10.1007/978-3-031-98459-4_30},
      url={https://arxiv.org/abs/2505.21517}, 
}

@misc{li2017usersimulatortaskcompletiondialogues,
      title={A User Simulator for Task-Completion Dialogues}, 
      author={Xiujun Li and Zachary C. Lipton and Bhuwan Dhingra and Lihong Li and Jianfeng Gao and Yun-Nung Chen},
      year={2017},
      eprint={1612.05688},
      archivePrefix={arXiv},
      primaryClass={cs.LG},
      url={https://arxiv.org/abs/1612.05688}, 
}

@misc{dodge2020finetuningpretrainedlanguagemodels,
      title={Fine-Tuning Pretrained Language Models: Weight Initializations, Data Orders, and Early Stopping}, 
      author={Jesse Dodge and Gabriel Ilharco and Roy Schwartz and Ali Farhadi and Hannaneh Hajishirzi and Noah Smith},
      year={2020},
      eprint={2002.06305},
      archivePrefix={arXiv},
      primaryClass={cs.CL},
      url={https://arxiv.org/abs/2002.06305}, 
}

@misc{finn2017modelagnosticmetalearningfastadaptation,
      title={Model-Agnostic Meta-Learning for Fast Adaptation of Deep Networks}, 
      author={Chelsea Finn and Pieter Abbeel and Sergey Levine},
      year={2017},
      eprint={1703.03400},
      archivePrefix={arXiv},
      primaryClass={cs.LG},
      url={https://arxiv.org/abs/1703.03400}, 
}

@misc{yang2026plugmemtaskagnosticpluginmemory,
      title={PlugMem: A Task-Agnostic Plugin Memory Module for LLM Agents},
      author={Ke Yang and Zixi Chen and Xuan He and Jize Jiang and Michel Galley and Chenglong Wang and Jianfeng Gao and Jiawei Han and ChengXiang Zhai},
      year={2026},
      eprint={2603.03296},
      archivePrefix={arXiv},
      primaryClass={cs.CL},
      url={https://arxiv.org/abs/2603.03296},
}

@article{Koedinger_2007, title={Exploring the Assistance Dilemma in Experiments with Cognitive Tutors}, volume={19}, ISSN={1573-336X}, url={http://dx.doi.org/10.1007/s10648-007-9049-0}, DOI={10.1007/s10648-007-9049-0}, number={3}, journal={Educational Psychology Review}, publisher={Springer Science and Business Media LLC}, author={Koedinger, Kenneth R. and Aleven, Vincent}, year={2007}, month=July, pages={239–264} }

@article{Wood_1976, title={THE ROLE OF TUTORING IN PROBLEM SOLVING}, volume={17}, ISSN={1469-7610}, url={http://dx.doi.org/10.1111/j.1469-7610.1976.tb00381.x}, DOI={10.1111/j.1469-7610.1976.tb00381.x}, number={2}, journal={Journal of Child Psychology and Psychiatry}, publisher={Wiley}, author={Wood, David and Bruner, Jerome S. and Ross, Gail}, year={1976}, month=Apr, pages={89–100} }

@book{VYGOTSKY_1980, title={Mind in Society: Development of Higher Psychological Processes}, ISBN={9780674576285}, url={http://dx.doi.org/10.2307/j.ctvjf9vz4}, DOI={10.2307/j.ctvjf9vz4}, publisher={Harvard University Press}, author={VYGOTSKY, L. S.}, year={1980}, month=Oct }

@article{Grigorenko_1998, title={Dynamic testing.}, volume={124}, ISSN={0033-2909}, url={http://dx.doi.org/10.1037/0033-2909.124.1.75}, DOI={10.1037/0033-2909.124.1.75}, number={1}, journal={Psychological Bulletin}, publisher={American Psychological Association (APA)}, author={Grigorenko, Elena L. and Sternberg, Robert J.}, year={1998}, month=July, pages={75–111} }

\newpage
\appendix

\section{Related Work}
\label{app:related_work}

This appendix expands Section~\ref{sec:related_work} with a per-paper survey along the same three lines: cognitive-state and behavior-prediction models (\cref{app:related_work_state}), LLM-prompted student simulators (\cref{app:related_work_llm}), and tutor-side evaluation and optimization (\cref{app:related_work_tutoring}).

\subsection{Cognitive-State and Behavior-Prediction Models}
\label{app:related_work_state}

\paragraph{Knowledge tracing.}
Knowledge tracing infers a student's latent cognitive state from a sequence of observed responses and predicts future responses. Bayesian Knowledge Tracing maintains a per-skill Bernoulli mastery variable updated after each attempt with four parameters (prior, learn, slip, guess) \citep{Corbett_1995}; Deep Knowledge Tracing replaces hand-specified updates with a recurrent network over the interaction history \citep{piech2015deepknowledgetracing}; attentive variants condition on items and recent context for longer-range and item-similarity effects \citep{ghosh2020contextawareattentiveknowledgetracing}. Related state representations include key-value memory and transformer KT, auxiliary-task DKT, Item Response Theory and multidimensional IRT, and performance- and learning-factor analysis \citep{Zhang_2017, pandey2019selfattentivemodelknowledgetracing, Choi_2020, Liu_2023, Lord_2012, Reckase_1985, Pavlik_Philip_I__2009, Cen_2006}. Across the family the state is a compact latent updated from labeled response outcomes (correct/incorrect, sometimes latency, partial credit, or hints used), and the prediction target is the next item's outcome; the observation channel never includes free-form tutor guidance.

\paragraph{LLM-based knowledge tracing.}
From 2024 onward, question text and dialogue enter the KT stack. Integrating pretrained language-model representations of question text improves cold-start generalization over item-id baselines \citep{lee2024languagemodelknowledgetracing}; extending KT to tutor-student dialogues with per-turn skill and correctness annotation beats classical KT especially under limited data \citep{Scarlatos_2025}; and uncertainty-aware variants add stochastic-embedding uncertainty \citep{Cheng_2025}. The dialogue-KT entry is the closest published prior art for our $\mathcal{R}$ axis, being the only KT variant that ingests tutor language as input, yet all of these predict answer correctness from one shared model rather than generating a specific student's responses from a per-student one.

\paragraph{From correctness prediction to response generation.}
Option Tracing predicts which distractor a student selects rather than only whether they answer correctly \citep{Ghosh_2021}; Open-Ended KT extends the idea to free-form responses in programming tasks, predicting the exact response rather than its correctness label, and is the canonical precursor of \textsc{StudentSim}'s response-generation objective \citep{Liu_2022}.

\paragraph{Human behavior cloning in chess.}
A parallel strand fits behavioral predictors directly to logs of human play. Maia trains a separate move-prediction network for each rating band and reproduces characteristic human mistakes at the target skill level, best predicting the blunders human players make \citep{mcilroyyoung2020aligningsuperhumanaihuman}; Maia2 extends it into a single skill-conditioned network with stronger prediction and broader rating coverage \citep{tang2024maia2unifiedmodelhumanai}. Maia2 is closer to population-level behavior cloning at a target skill than to per-skill state tracking, but shares the same conditioning limit: a small structured descriptor (rating, history, sometimes time control), no free-form text, and no per-individual fit.

\paragraph{Where the gap shows up.}
This whole line updates state only from labeled outcomes, has no architectural route by which a paragraph of explanation, a hint, or a worked example becomes an update to the latent, and is fit population-wide. Under our metrics it clears $\mathcal{F}$ at or above untrained-LLM baselines (e.g., on chess, Maia2 is the strongest $\mathcal{F}$ baseline; Section~\ref{sec:fidelity_results}), but its $\mathcal{R}$ collapses to a near-zero, zero-instruction floor because guidance text has no input pathway (Section~\ref{sec:guidance_results}).

\subsection{LLM-Prompted Student Simulators}
\label{app:related_work_llm}

\paragraph{User and student simulation.}
A second strand prompts a large language model to role-play a target user: build a textual description of the target (preferences, knowledge state, persona), prepend it to the conversation, and let the model continue conditioned on it. This recipe is used to evaluate conversational and task-oriented dialogue agents by replaying simulated user turns \citep{owoicho2023exploitingsimulateduserfeedback, terragni2023incontextlearningusersimulators}, and recurs as persona prompting, profile-conditioned simulation, synthetic-user generation for offline evaluation, and interactive goal-directed simulation, sometimes constrained with in-context exemplars or dual-LLM coordination \citep{luo2024duetsimbuildingusersimulator}. A parallel line measures whether such role-plays stay consistent with a described character, using the resulting scores as benchmarks, as training data, or as rewards for fine-tuning persona-following \citep{peng2024quantifyingoptimizingglobalfaithfulness, samuel2025personagymevaluatingpersonaagents, wang2024incharacterevaluatingpersonalityfidelity, Jiang_2024, Wang_2024, Shao_2023, Tu_2024, gusev2025pingpongbenchmarkroleplayinglanguage}.

\paragraph{Critical observation.}
The simulator is a frozen LLM with a prompt, pretrained to generate plausible language rather than to map a description of a learner onto the behavioral consequences of that cognitive state. Two failure modes follow: the simulator solves problems beyond the described student's level, or its surface mimicry overshoots and misses problems the student would solve. Both appear in our chess data, where GPT-5.4 reaches $\mathcal{F}=0.23$ and GPT-4o $\mathcal{F}=0.22$, below state-tracking baselines, even as they attain higher $\mathcal{R}$ than state-tracking models can (Section~\ref{sec:fidelity_results}). We classify any system whose output is a student-style response (an answer, an essay, a confusion message) as a student simulator, and reserve ``judge'' or ``LLM-as-judge'' for systems whose output is a scalar score or rubric verdict \citep{zheng2023judgingllmasajudgemtbenchchatbot}; the proxy-reward and LLM-as-judge line lives in \cref{app:related_work_tutoring}.

\subsubsection{Student-Simulation Validity}
\label{app:related_work_validity}
The validity of LLM-based student simulation has emerged as an open question. The \emph{competence paradox} holds that a broadly capable LLM asked to behave as a partially knowledgeable learner produces fluent but unreliable student text, because its pretrained reasoning and helpfulness habits override the competence description; this line reframes simulator validity as constrained generation under an explicit epistemic-state specification, with state evolution under instruction as a required axis rather than an emergent side effect of prompting \citep{yuan2026validstudentsimulationlarge}. Empirical work measures simulated-student quality directly and reports systematic mismatches with real students in solve rate and error type \citep{scarlatos2026simulatedstudentstutoringdialogues}, and other work proposes architectural changes such as cognitive prototype modules or controlled imperfect reasoning to elicit student-like errors instead of correcting them away \citep{Wu_2025}. Teacher-facing studies add that instructors find LLM students overly attentive, emotionally flat, linguistically too sophisticated, and inconsistent across turns \citep{Martynova_2025}. Our $\mathcal{F}$ axis targets exactly this per-individual fidelity gap: each simulator is trained against a real learner's behavioral data and scored per student on held-out records that learner produced. $\mathcal{R}$, by contrast, scores whether the simulator produces the canonical corrected response under tutor guidance (Appendix~\ref{app:design_g_semantics}), which is what a tutor RL loop rewards: guidance that moves the student to the right answer. How closely $\mathcal{R}$ tracks an individual student's literal post-guidance behavior is a complementary, counterfactual question that controlled studies with real students under matched guidance can address, a natural extension of the metric.

\subsubsection{Educational Student Simulators}
Educational simulators instantiate the prompting recipe in several ways: half-synthetic tutoring dialogues from teacher--LLM-student pairing or crowdworker role-play \citep{macina2023mathdialdialoguetutoringdataset, Stasaski_2020}; multi-agent classrooms and generative learner agents with reflective memory, class-control mechanisms, or longitudinal real-student datasets \citep{zhang2024simulatingclassroomeducationllmempowered, gao2026agent4edugeneratinglearnerresponse, xu2025classroomsimulacrabuildingcontextual}; profile- and personality-conditioned simulators tying behavior to cognitive-ability and personality traits \citep{Liu_2024}; and application-driven simulators for item-quality evaluation, teaching-assistant training, and learner-data generation \citep{lu2024generativestudentsusingllmsimulated, Markel_2023, xu2024eduagentgenerativestudentagents}. A per-individual sub-line specifies agents per target rather than per group: patient-simulation elicits per-target natural-language behavioral principles for a prompted LLM, the closest per-individual specification methodology even though it is prompt-based and non-educational \citep{louie2024roleplaydohenablingdomainexpertscreate}, and social-science work constructs per-individual agents from a person's self-reports, builds generative agents with retrieval-based memory, or equips LLM agents with task-agnostic plugin memory modules, using prompting, retrieval, and external memory rather than weight specialization and without evaluating state evolution under instruction \citep{park2026llmagentsgroundedselfreports, Park_2023, yang2026plugmemtaskagnosticpluginmemory}. These approaches vary along granularity (persona, profile, group, or individual), conditioning interface (prompt, prompt-plus-retrieval, or weight-level fine-tuning), and evaluation target (stylistic plausibility, a downstream task, or response-level fidelity to a specific learner); \textsc{StudentSim} is per-individual, weight-level, and response-level with the additional $\mathcal{R}$ guidance-response axis, a combination we are not aware of in prior educational work.

\subsection{Tutor-Side Evaluation and Optimization}
\label{app:related_work_tutoring}

\paragraph{Tutor models, judges, and benchmarks.}
The third line studies the AI tutor side of the loop. Tutor-specialized models supervise on expert-curated or synthetic tutoring dialogues \citep{chevalier2024languagemodelssciencetutors}; proxy-reward and LLM-as-judge methods supply scalar feedback by querying a generic LLM against a quality rubric, with the target being rubric quality or human preference rather than a specific student's response \citep{zheng2023judgingllmasajudgemtbenchchatbot}. Benchmark work moves from single-rubric judges toward multi-dimensional pedagogical evaluation and documents a tension between a tutor model's subject expertise and its pedagogical skill \citep{macina2025mathtutorbenchbenchmarkmeasuringopenended, maurya2025unifyingaitutorevaluation}. Classroom-scale studies test tutor-assist and tutor-as-replacement settings against real students and report learning gains, in some cases with particular benefit to under-served populations \citep{wang2025tutorcopilothumanaiapproach}, and LLM-tutor instruction-tuning reports parallel pedagogical-quality gains under expert preference evaluation \citep{learnlmteam2025learnlmimprovinggeminilearning}. A separate line conditions tutor output on a student descriptor or a structured pedagogical strategy (a profile, a recent error, an instructional style, or step-by-step scaffolding) \citep{Sonkar_2023, Nye_2014, liu2024socraticlm}, extending the intelligent tutoring systems (ITS) lineage exemplified by Squirrel~AI, ALEKS, and Carnegie Learning; \textsc{StudentSim} differs from ITS-era student models in handling open-ended natural-language guidance and training per individual rather than fitting a small skill graph with hand-coded production rules.

\paragraph{Simulator-grounded rewards.}
The closest cluster to our downstream setup uses a knowledge-tracing model or a prompted LLM student as the reward signal in a tutor optimization loop rather than a generic rubric. One pattern scores candidate tutor utterances by a knowledge-tracing model's predicted post-turn correctness, combines that with a rubric judge, and trains the tutor via direct preference optimization \citep{scarlatos2025trainingllmbasedtutorsimprove}; another runs a full online RL loop in which the tutor and a prompted LLM student converse for multiple turns and the tutor is rewarded by the simulated student's post-dialogue solve rate plus a pedagogy judge, optimized via GRPO \citep{dinucujianu2025problemsolvingteachingproblemsolvingaligning}. What recurs across this cluster is that the reward is grounded in a knowledge-tracing model that predicts only answer correctness or a prompted LLM student, rather than a simulator trained on real student behavioral data; where the student is a prompted LLM, the model that judges its simulated learner is typically the same generic LLM, so the reward is only as well-grounded as that LLM's persona-following, which the validity literature in \cref{app:related_work_validity} documents to be unstable. Section~\ref{sec:tutor_rl} supplies the missing pathway: a frozen \textsc{StudentSim}, trained on real learner data and calibrated on $\mathcal{F}$ and $\mathcal{R}$, is plugged into a tutor RL loop as the reward source, with the reward depending on whether the simulated student's response improves under tutor guidance ($r = \text{score}(A_\text{rev}) - \text{score}(A_\text{prev})$). Because the signal is grounded in real-learner-derived behavior rather than a prompted persona or generic rubric, expert raters judge the resulting tutor as more accurate, better-guided, and more personalized than tutors rewarded by a prompted-simulator reward or no reinforcement learning (\cref{sec:tutor_rl}).

\section{Per-Benchmark Data and Tasks}
\label{app:benchmark_details}

This appendix documents the source corpus, raw record fields, training-instance formats, and corpus-level statistics for each of the three domains the \textsc{StudentSim} framework instantiates. Each subsection follows the same five-block structure: data source, raw record, single-turn record format $S_i = \{(x, m)\}$, multi-turn record format $T_i = \{(x, m, \tau, m^*)\}$, and a corpus-statistics tabular. Symbols $S_i$ and $T_i$ are defined in Section~\ref{sec:problem_formulation}. Worked examples below show real fields with identifiers anonymized as Player~A (chess) or Student~A (second-language English writing and mathematics).

\subsection{The \textsc{StudentSimEval} Evaluation Protocol}
\label{app:studentsimeval_protocol}

\textsc{StudentSimEval} fixes, once and independently of any method, both the population of evaluated students and the per-student held-out records that every method is scored on. It evaluates a frozen roster of $30$ chess players, $15$ L2 learners, and $15$ math students ($60$ in total). For each student it fixes two held-out sets, disjoint from all training data: a single-turn set $S_i$ that scores behavioral fidelity $\mathcal{F}$, and a multi-turn set $T_i$ that scores guidance responsiveness $\mathcal{R}$. The split is chronological: each student's earlier records form the training pool and their later records are held out for $S_i$ and $T_i$, so every score reflects prediction of a student's later behavior from their earlier records. Every method, our trained simulators and the GPT-4o, GPT-5.4, Maia2, and base-Qwen baselines alike, is fit on the same per-student training records and scored on these same held-out sets. The held-out sets are materialized once as frozen split files and reused verbatim, so they are a property of the protocol, not of any method; a new method is evaluated by running it on the same files. The per-student held-out sizes are the ``Held-out'' columns of \cref{tab:training_records}: $|S_i| = 5{,}000$ and $|T_i| = 4{,}000$ in chess; $|S_i| = 26$ and $|T_i| = 40$ in L2; $|S_i| = 66$ in math, with $|T_i|$ ranging over $21$ to $99$. The multi-turn set is split evenly across each domain's guidance modes: four in chess (error remediation, comparative, strategic, Socratic), two in L2 (point-based, rule-based), and three in math (error remediation, Socratic, conceptual).

The protocol equalizes the number of held-out records per student wherever the data allows, so that no student dominates a population mean. Where a domain has more records than the budget needs, a cap gives every student the same count; where equalizing would leave too little to estimate the metric, the protocol keeps every available record instead. Chess uses an absolute cap of $5{,}000$ single-turn positions and $1{,}000$ multi-turn records per guidance mode, subsampled so every player contributes the same count from a larger available pool. For the scarcer single-turn data in L2 and math, we equalize to the smallest student's held-out size, giving $|S_i| = 26$ (L2) and $|S_i| = 66$ (math) for every student, and L2 caps its multi-turn set at $20$ records per correction mode. Math multi-turn records are the case where equalizing would leave too little to estimate $\mathcal{R}$, since a multi-turn record exists only for a problem the student first answered incorrectly, so the protocol keeps every eligible wrong-answer problem, fanned across the three guidance modes ($879$ records in total, $293$ per mode). Training draws a separate capped subset of each student's records, so the held-out portion is reserved and never enters any training stage.

\subsection{Chess}
\label{app:benchmark_chess}

\paragraph{Data source.}
The chess records come from the public Lichess game-database export at \texttt{https://database.lichess.org/} \citep{lichess}, specifically the May 2025 monthly snapshot of standard-time-control human games. Lichess is an open-source online chess platform on which every published game is between two real human players; no game in the export is engine-vs-engine or otherwise synthetic. The platform also publishes Stockfish position evaluations for many games, and our pipeline uses these where available rather than running Stockfish from scratch. The export is released under CC0, which permits unrestricted research, redistribution, and modification.

\paragraph{Raw record.}
A Lichess game is a sequence of half-moves recorded in PGN, plus per-game metadata (the two players' Lichess handles, their ELO ratings at the time of the game, time control, opening ECO code, and final outcome). For each half-move we extract the position before the move in FEN notation, the player's actual move in UCI, and engine-derived quality features for that move: the engine top-$k$ move list with their centipawn evaluations, the player's centipawn loss for the played move, a binary \texttt{is\_best\_move} flag from the engine ranking, the index of the played move within the engine top-$k$ list, and the move-index within the game (used downstream to bucket positions by phase). The fields kept in each per-position record are therefore: \texttt{player\_handle}, \texttt{player\_elo}, \texttt{time\_control}, \texttt{game\_outcome}, \texttt{move\_idx}, \texttt{fen\_before}, \texttt{player\_uci}, \texttt{engine\_top\_k}, \texttt{cp\_loss}, \texttt{is\_best\_move}, and \texttt{player\_move\_rank}. We discard pre-move clock times, evaluation comments embedded in the PGN, and game-chat fields. Positions are bucketed by phase into very-early opening (\texttt{move\_idx} $< 10$), opening (\texttt{move\_idx} $\in [10, 19]$), middlegame ($[20, 39]$), and endgame ($\ge 40$).

\paragraph{Single-turn record format $S_i = \{(x, m)\}$.}
The single-turn input $x$ is a templated player-context prompt followed by the FEN of a position and a request for the next move in UCI. The player-context block is fully derivable from the player's prior games on Lichess and contains: a basic profile (recent win rate, best-move accuracy, average centipawn loss), a coarse style label (offensive / balanced / defensive) plus offensive-move and defensive-move rates, per-piece usage rates, recent-history features (last three game results, current best-move streak by player and opponent, recent captures gained and lost), an opening-move-tendency line for early-game positions (\texttt{move\_idx} $\le 10$), and a top-$5$ Maia2-derived candidate-move list with per-move probabilities. For the closed-source LLM baselines we omit the Maia2-derived candidate-move list, so their scores reflect only their own student modeling and are not influenced by Maia2's predictions. The response $m$ is the player's actual move on the held-out position, in UCI. A real held-out record from Player~A illustrates the format:
\begin{lstlisting}[caption={A single-turn record for Player~A},label={lst:chess_single_turn}]
x:
  Player context: Recent win rate: 54.6%; Best-move accuracy: 32.84%;
  Avg centipawn loss: 215.74; Style: offensive;
  Piece usage: bishop 14.00%, knight 19.68%, queen 17.38%;
  Last 3 game results: loss, loss, loss; ...
  The current position in FEN notation is:
  2kr1b1r/pp2pppp/5nb1/8/2BN4/4N3/PPPB1PPP/R3K2R b KQ - 0 12
  You are playing as Black. Respond in UCI format.

m:
  d8d4
\end{lstlisting}

\paragraph{Multi-turn record format $T_i = \{(x, m, \tau, m^*)\}$.}
The multi-turn record reuses the same $x$ and the same player move $m$, then appends two additional fields. Multi-turn records are drawn from positions outside the very-early opening (\texttt{move\_idx} $\ge 10$), because an opening move reflects a player's repertoire and preparation, where there is no single move to correct toward. Tutor guidance $\tau$ is a natural-language tutor message in one of four pedagogical styles (\emph{error remediation}, \emph{comparative}, \emph{strategic}, \emph{Socratic}), generated by an LLM tutor (GPT-4o or GPT-5.1, selected per record by whether the player's move fell outside the engine's top-$k$ list) conditioned on the FEN, the player's actual (often non-optimal) move, the engine best move, and the engine continuations for both. Each style is realized by a fixed prompt template that constrains the tutor's framing: error-remediation pinpoints the mistake and states the correction, comparative contrasts the played move with the engine's top alternatives, strategic foregrounds long-term plan considerations, and Socratic prompts the player with one or two leading questions. The reference move $m^*$ is the engine best move at fixed depth on the same FEN. A real strategic-style record from Player~A:
\begin{lstlisting}[caption={A multi-turn record for Player~A, strategic style},label={lst:chess_multi_turn}]
x:
  Player context (as above)
  + FEN 7k/5R2/2pr2p1/2p4p/P5p1/1B1P4/2P3PP/6K1 w - - 0 30
  You are playing as White.

m:
  h2h3

tau (strategic):
  "In this position, the strategic focus should be on maximizing piece
  activity and pawn structure. h2h3 does not address the immediate
  strategic opportunities; consider creating a passed pawn with a4a5,
  activating the bishop with b3c4, or improving the rook via f7f8 ..."

m*:
  a4a5
\end{lstlisting}

\begin{figure}[h]
  \centering
  \includegraphics[width=0.32\textwidth]{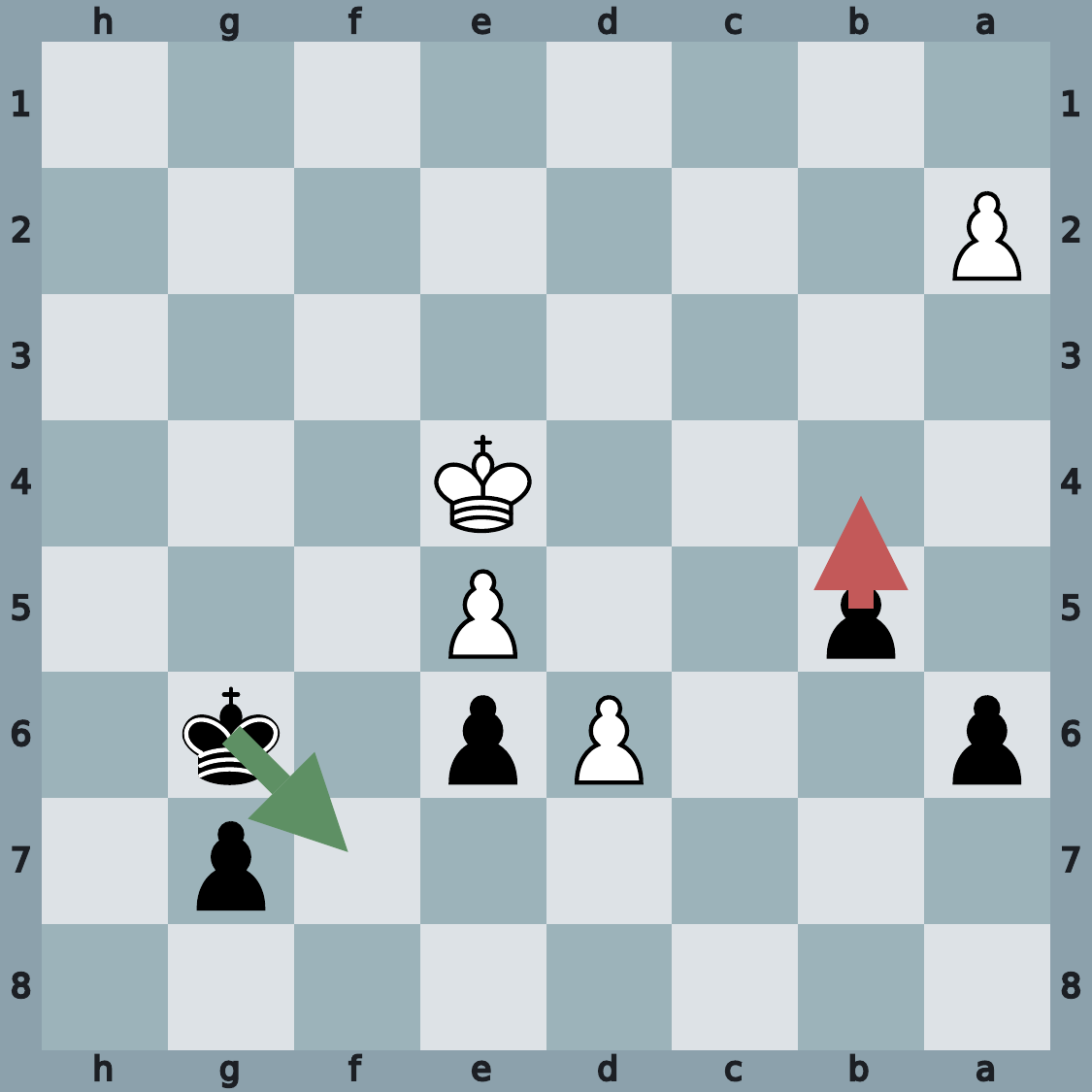}
  \caption{\textbf{The position behind the four-style example.} Black to move, FEN \texttt{8/6p1/p2Pp1k1/1p2P3/4K3/8/P7/8 b - - 0 41}. Red marks the move the player actually made (\texttt{b5b4}); green marks the engine best move $m^*$ (\texttt{g6f7}).}
  \label{fig:fourmode_board}
\end{figure}

The four styles differ in how much of the correction they hand over. The record below shows all four on a single king-and-pawn endgame from Player~B (\cref{fig:fourmode_board}), who pushed a flank pawn with \texttt{b5b4} and lost a winning position (engine evaluation $+425$ to $-656$ centipawns); the engine best move is the king activation \texttt{g6f7}.

\begin{lstlisting}[caption={The same position under all four guidance styles, Player~B},label={lst:chess_four_modes}]
x:
  Player context (as above)
  + FEN 8/6p1/p2Pp1k1/1p2P3/4K3/8/P7/8 b - - 0 41
  You are playing as Black.

==========================================================================
m:
  b5b4

==========================================================================
tau (error remediation):
  "Your move b4 initiates an unnecessary pawn push that loses critical tempo and
  worsens your position significantly. After b4, White plays d7, forcing your king to
  retreat to h7, and quickly promotes their pawn with d8=Q. This sequence leaves Black
  materially and positionally worse, as White's queen dominates the board and Black's
  pawns are left defenseless. Instead, the best move is Kf7. By activating your king
  with Kf7, you challenge White's king on e4 and prepare to hinder White's advancing
  pawns, starting with their central d-pawn. After Kf7, if White responds with Ke3, you
  can play b4 to activate your passed pawn while keeping your king near the action on
  d4-e4. This keeps White's king under pressure and delays their pawn promotion
  significantly. Effective king positioning is critical in endgames, and Kf7 sets up a
  much stronger resistance plan. To improve in positions like this, prioritize using
  your king actively in endgames to control key squares and block pawn advances. Avoid
  pushing pawns recklessly, as that can waste time and leave your king unable to
  respond to threats. Look for ways to hinder your opponent's plans and play precise
  moves that maximize your king's influence on the board."

==========================================================================
tau (comparative):
  "Let's analyze three candidate moves: your move b4, Kf7, and Kh6. Your move, b4, aims
  to activate the pawn structure on the queenside and potentially disrupt White's pawn
  advancement. The continuation b4 d7 Kh7 d8=Q a5 follows. This move allows White to
  push the d7 pawn, creating a strong threat with the promotion to a queen on d8. The
  resulting position places White in a commanding position due to the powerful queen
  while Black's pawn play hasn't resulted in any significant counter-threats. Next,
  consider Kf7. This move shifts the Black king toward greater activity, attempting to
  control key squares and potentially disrupt White's advancement on the d-file. The
  continuation shows Kf7 Ke3 b4 Kd4 Ke8. Here, the Black king aims to support the
  defense of the back rank and control pivotal central squares, limiting White's
  capability to convert the d-pawn effectively. This active king positioning can lead
  to improved chances down the line. Lastly, Kh6 serves as a more defensive option. Its
  continuation is Kh6 d7 g5 d8=Q g4. This move attempts to prepare for an active pawn
  push on the kingside while avoiding immediate threats from the advancing d-pawn.
  However, promoting a queen on d8 remains a formidable option for White, which Kh6
  barely manages to counter. Comparing these approaches, piece activity and king
  mobility, seen in Kf7, prove more decisive in confronting White's pawn strength on
  the d-file than launching pawns themselves in b4 or the tentative king safety of
  Kh6."

==========================================================================
tau (strategic):
  "In this position, the key strategic themes are the advancement of passed pawns and
  king activity. White has a dangerous passed pawn on d6, and Black's priority should
  be neutralizing this threat. The student's move b4 advances a flank pawn, but misses
  immediate central concerns. In endgames, especially pawn races, it's crucial to
  ensure that the opponent's most advanced pawns are blocked or captured before
  advancing your own. Consider moves that bring the king closer to White's advanced
  pawns, like Kf7. Such moves help restrain the opponent's pawn while activating your
  king. The strategic principle is that in pawn races, central pawn control and king
  activation will often dictate the balance of power. Moves aimed at mobilizing the
  king to confront the central passed pawn ensure that you stand ready to handle the
  advancing threat efficiently. In a similar situation, remember to focus on
  neutralizing central threats first before considering peripheral pawn advances."

==========================================================================
tau (Socratic):
  "1. What happens to the balance of the position after b4? Consider the consequences
  for both sides' pawn structures. 2. How does White respond if Black tries b4? How
  effective is this in stopping the d-pawn? 3. What role does the Black king play in
  this position? Can it impact the pawn promotion directly? 4. Compare the move b4 with
  a potential move that increases king activity. How do these moves change the pressure
  on White's advance? 5. By considering moves other than b4, how can Black use king
  movement to create counterplay or directly block White's threats?"

==========================================================================
m*:
  g6f7
\end{lstlisting}

\paragraph{Corpus statistics.}
The May 2025 standard-time-control export covers tens of millions of games drawn from a multi-million-player pool on Lichess. The Stage-1, Stage-2, and per-student held-out counts that draw from this pool are summarized in \cref{tab:training_records} of Section~\ref{sec:experiment_setup}; the Stage-2 pool is nested inside the Stage-1 pool. Players are sampled to span a broad ELO band rather than a narrow rating slice, with a minimum game-count threshold so that each sampled player has enough recorded games to support both pooled training and held-out evaluation. The multi-turn evaluation set is restricted to non-best-move positions outside the very-early-opening band.

\begin{table}[h]
\centering
\small
\caption{Chess corpus statistics. Stage-1 / Stage-2 training counts and per-student held-out evaluation counts ($\mathcal{F}$ test, $\mathcal{R}$ test) are reported in \cref{tab:training_records}.}\label{tab:benchmark_chess_stats}
\begin{tabular}{@{}p{0.45\linewidth} p{0.5\linewidth}@{}}
\toprule
\textbf{\boldmath Field} & \textbf{\boldmath Value} \\
\midrule
Source corpus & Lichess May~2025 standard-time-control export, CC0 \\
Engine analysis & Stockfish, top-$4$ candidate moves, depth adapted to game phase ($8$ in the very-early opening up to $15$ in the endgame) \\
Per-record response space & legal moves at the position, in UCI \\
Guidance modes & error remediation, comparative, strategic, Socratic ($4$ total) \\
\bottomrule
\end{tabular}
\end{table}

\subsection{Second-Language English Writing}
\label{app:benchmark_l2}

\paragraph{Data source.}
The L2 records come from EFCAMDAT, the EF--Cambridge Open Language Database \citep{geertzen2013automatic}, a large open-access corpus of English essays written by real human L2 English students (real students learning English as a second language) on the Education First online-school platform. We use the second release, which contains roughly $1.18$M essays from roughly $174{,}000$ students across the platform's 16 proficiency levels, which are aligned with the Common European Framework of Reference for Languages (CEFR), the six-level scale running from A1 for a beginner to C2 for a near-native writer. Two derived sub-corpora are relevant: the \emph{cleaned subcorpus} \citep{Shatz_2020} restricts to students from the eleven most-represented nationalities and supplies normalized texts up to level 15, and the \emph{cleaned error-coded subcorpus} \citep{Oksuz_2025} adds part-of-speech tagging and per-token error-category annotations on top of the cleaned subset. Access requires an academic affiliation and a signed user agreement administered by the Cambridge research lab. No essays in our pipeline are synthetic.

\paragraph{Raw record.}
Each EFCAMDAT essay has: a student identifier (the field is named \texttt{learner\_id} in the source corpus), the student's nationality and current EFCAMDAT level, which the profile block also renders as its approximate CEFR equivalent (levels $1$--$15$ mapped onto A1 through C2), the lesson topic and unit, the required key words from the lesson, the raw essay text, the error-coded version of the essay (per-token tags by category and sub-category), the canonical corrected version of the essay, the platform's grade for the essay, and lightweight metadata (essay length, time spent). At training-instance construction time we keep \texttt{learner\_id}, \texttt{nationality}, \texttt{efcamdat\_level} (and derived CEFR), \texttt{topic\_id}, \texttt{topic\_title}, \texttt{level\_unit}, \texttt{key\_words}, \texttt{raw\_text}, \texttt{corrected\_text}, the per-token error-category annotations from the error-coded subcorpus, and the platform-assigned grade; we discard time-spent and other UI-side fields. EFCAMDAT marks each error in an essay with three pieces of information: the text the student wrote, the correction for it, and one of fifteen category labels (spelling, word choice, article, preposition, verb tense, plural, agreement, word order, and so on). A student who writes \emph{I go to school yesterday} would have one error recorded, with original text \texttt{go}, correction \texttt{went}, and label \texttt{verb tense}. We keep all three fields. Counting the labels over a learner's earlier essays produces the error-frequency line of the student-profile block, which for one learner on our roster reads \texttt{Common error patterns: spelling(37), word choice(30), extra word (delete)(28), missing word(23), article(17)}, and a few of the original-to-correction pairs are listed under it as concrete examples. Each single marked error also becomes one multi-turn record, where the tutor turn points at that original text and the student's target reply is the correction. The seven LanguageTool issue-type buckets used by the fidelity metric (misspelling, grammar, typographical, style, uncategorized, whitespace, inconsistency) are not raw fields: they are computed downstream from \texttt{raw\_text} via the LanguageTool open-source checker, and their formal role in the metric is defined in Appendix~\ref{app:metric_definitions}. 

\paragraph{Single-turn record format $S_i = \{(x, m)\}$.}
The single-turn input $x$ is a student-profile-conditioned essay prompt: the student's nationality and CEFR level, an essay-history summary (essays written so far, average length, average platform grade), the top error categories the student makes most often with concrete examples of recent corrections, a short list of recent essays on similar topics with their grades, and the new lesson task (topic title, level-unit pair, required key words). The response $m$ is the student's actual raw essay, with their characteristic errors preserved. EFCAMDAT is released to researchers on request and is not ours to redistribute, so we give the record schema instead of a verbatim record; the rendered prompt retains EFCAMDAT's source-side ``Learner'' field labels:
\begin{lstlisting}[caption={Single-turn record schema (L2). EFCAMDAT is released to researchers on request, so we show the schema in place of a verbatim record.},label={lst:l2_single_turn}]
x:
  Learner profile: nationality; current EFCAMDAT level (with CEFR equivalent);
  essays written so far, average length, average platform grade;
  common error patterns with counts;
  recent essays (most recent last), each with its lesson topic and grade
  Now write a response to this lesson task: topic; level and unit; key words

m:
  the learner's own essay for that lesson task
\end{lstlisting}

\paragraph{Multi-turn record format $T_i = \{(x, m, \tau, m^*)\}$.}
The multi-turn record reuses the same $x$ and the same student essay $m$, and then appends a tutor turn $\tau$ and a canonical corrected fragment $m^*$. The tutor message $\tau$ is generated in one of two surface-form templates: \emph{point-based}, which states that an error of a given kind is present in the highlighted span and asks the student to fix it, and \emph{rule-based}, which states the underlying linguistic rule first and then asks the student to apply it to the highlighted span. Only the surface phrasing of the two style templates is generated; the highlighted span and the target correction are drawn directly from EFCAMDAT's own error-coded annotations. The reference $m^*$ is the correction EFCAMDAT records for that span. For errors of the delete type, their target is the literal token \texttt{delete}.  The rule-based schema continues the same record:
\begin{lstlisting}[caption={Multi-turn record schema (L2), shown as a schema for the same reason.},label={lst:l2_multi_turn}]
tau (rule-based, verb tense):
  the grammatical rule that governs the flagged span, followed by an excerpt
  of the learner's essay with the offending fragment marked, and a request to
  apply the rule to that fragment

m*:
  the teacher's corrected fragment
\end{lstlisting}

\paragraph{Corpus statistics.}
We build on the error-coded subcorpus of EFCAMDAT \citep{Oksuz_2025}, which is the release that carries the span-level teacher corrections our multi-turn records need. It holds $620{,}206$ essays from $116{,}312$ learners. Per-learner volume is low and right-skewed: a learner writes $5.3$ essays on average and $3$ at the median, $33.3\%$ of learners have a single essay, and $71.4\%$ have five or fewer. Learners are ranked by recorded volume and the most active are taken for the Stage-1 and Stage-2 pools, so that every selected learner supports both training and a frozen held-out split; the Stage-2 learners therefore sit above the corpus median, contributing on the order of $90$ essays each. The Stage-1, Stage-2, and per-student held-out counts that draw from this corpus are summarized in \cref{tab:training_records} of Section~\ref{sec:experiment_setup}; the Stage-2 pool is nested inside the Stage-1 pool. Students are sampled to span the level range and the eleven most-represented nationalities, with a minimum essay-count threshold to support held-out evaluation. The seven LanguageTool issue-type buckets used by the fidelity metric (misspelling, grammar, typographical, style, uncategorized, whitespace, inconsistency) are computed downstream and not directly retained as raw fields.

Although the per-student single-turn held-out budget on L2 ($S = 26$ in \cref{tab:training_records}) is smaller in essay count than the per-student budgets for chess or mathematics, the per-student estimator of $\mathcal{F}$ is driven by error events rather than by essays. Each held-out essay typically contains on the order of $5$--$20$ LanguageTool-flagged events, so the $26$ essays per student aggregate to roughly $10^2$--$10^3$ per-100-word error events across the seven LanguageTool buckets, and the per-student error-density vector that $\mathcal{F}$ scores is estimated from that event-level total rather than from the $26$ essay-level outcomes. The per-bucket density is also a continuous quantity over the seven buckets, not a binary correctness outcome, which lowers the variance of the per-student estimator at this scale. The formal definition of $\mathcal{F}$ for L2 and the per-bucket aggregation are given in Appendix~\ref{app:metric_definitions}.

\begin{table}[h]
\centering
\small
\caption{Second-language English writing corpus statistics. Stage-1 / Stage-2 training counts and per-student held-out evaluation counts ($\mathcal{F}$ test, $\mathcal{R}$ test) are reported in \cref{tab:training_records}. Shaded rows are the subcorpus we use and the learners we train on; the two unshaded scale rows describe the full release for context.}\label{tab:benchmark_l2_stats}
\begin{tabular}{@{}p{0.45\linewidth} p{0.5\linewidth}@{}}
\toprule
\textbf{\boldmath Field} & \textbf{\boldmath Value} \\
\midrule
Source corpus & EFCAMDAT, error-coded subcorpus \citep{Oksuz_2025} \\
Full second release, all essays & $\sim\!1{,}180{,}310$ \\
Full second release, all students & $\sim\!174{,}743$ \\
\rowcolor{bluegray!20} Error-coded subcorpus, essays & $620{,}206$ \\
\rowcolor{bluegray!20} Error-coded subcorpus, learners & $116{,}312$ \\
\rowcolor{bluegray!20} Essays per learner in the error-coded subcorpus & $5.3$ mean, $3$ median, $141$ max \\
\rowcolor{bluegray!20} Essays per learner, the $15$ Stage-2 learners & $92.4$ mean, $87$ median, $84$--$141$ range \\
LanguageTool issue-type buckets (fidelity metric) & misspelling, grammar, typographical, style, uncategorized, whitespace, inconsistency (7 total) \\
\bottomrule
\end{tabular}
\end{table}

\subsection{Mathematics}
\label{app:benchmark_math}

\paragraph{Data source.}
The mathematics records come from the FoundationalASSIST dataset \citep{worden2026foundationalassisteducationaldatasetfoundational}, a release of student--problem interactions from the ASSISTments online-tutoring platform aligned to the Illustrative Mathematics curriculum and, through it, to the Common Core State Standards, the United States grade-by-grade specification of what mathematics a student should master. Each problem in the release carries the standard codes it exercises, which is where the skill labels used below come from. Every interaction is by a real human student; no responses are synthetic. The dataset is hosted on Hugging Face under a gating mechanism that requires accepting a Responsible Use Agreement before download. FA-specific processing notes (Excel-date corruption fix, LLM answer-key audit, free-text vs.\ multiple-choice filter) are summarized below.

\paragraph{Raw record.}
Each FA interaction has a hashed student identifier, a problem identifier, the problem body (HTML with embedded MathML), the platform's claimed-correct answer key, the answer-type (Numeric, Algebraic Expression, Exact Match, Exact Fraction, Numeric Expression, Ordering, Drop Down, Multiple Choice, Check All That Apply), the student's free-text answer, a binary \texttt{discrete\_score}, and timestamps; supplementary tables map problem to skill (224 unique skills) and skill to Common Core standard. The fields kept in each interaction record are: \texttt{interaction\_id} (the unique row key, since $\sim\!5\%$ of student--problem pairs occur on multiple days), \texttt{user\_id}, \texttt{problem\_id}, \texttt{skill}, \texttt{problem\_body} (cleaned to plain text with paragraph breaks preserved and inline MathML rendered as \texttt{numerator/denominator}), \texttt{claimed\_answer} (after the date-remap and audit steps below), \texttt{answer\_type}, \texttt{student\_free\_text}, and \texttt{discrete\_score}. Three FA-specific quirks are handled before downstream record construction: \textit{i)} a subset of $149$ problems whose \texttt{Fill-in Answers} field was Excel-coerced from fractions to dates (\texttt{5/6} stored as \texttt{6-May}, \texttt{3/10} as \texttt{10-Mar}, \texttt{8/15} as \texttt{15-Aug}) is repaired deterministically by a date-pattern remap; \textit{ii)} every remaining \texttt{(problem\_id, claimed\_answer)} pair, the deterministically repaired ones excepted, is audited by GPT-4o, which solves the problem independently and returns \texttt{AGREE} or \texttt{DISAGREE} against the platform key, counting equivalent surface forms as agreement, and any pair whose verdict is not \texttt{AGREE} is dropped, which removes keys carrying arithmetic errors and keys for problems whose figures were lost in text extraction; and \textit{iii)} we keep the problems the platform scores as free text (Numeric, Algebraic Expression, Exact Match, Exact Fraction, Numeric Expression, Ordering, Drop Down), since those are the problems that record what other students wrote, which is where our distractors come from.

\paragraph{Single-turn record format $S_i = \{(x, m)\}$.}
The single-turn input $x$ is a student-profile prompt followed by a single problem in a four-way multiple-choice form. The student-profile block contains the pseudonym, overall and recent accuracy, per-skill performance summary (correct / attempted, with percentages), and the most recent attempts on the current problem's skill in compact form (correct flag, problem stem, student's free-text response). The multiple-choice form is constructed deterministically from the underlying free-text record: this student's own recorded answer is one option, and the other options are the answers other students most often gave on the same \texttt{problem\_id}, after a normalization step that collapses equivalent surface forms and removes any that coincide with this student's answer. The options are shuffled by a fixed per-record seed and labelled $A$ through $D$, and the prompt asks the simulator to pick a single letter, so $m$ is the letter carrying this student's own answer. A problem on which other students gave too few distinct answers to fill the options is dropped. FoundationalASSIST is distributed under a Responsible Use Agreement, so we give the record schema in place of a verbatim record:
\begin{lstlisting}[caption={Single-turn record schema (math). FoundationalASSIST is distributed under a Responsible Use Agreement, so we show the schema in place of a verbatim record. Student names are pseudonyms.},label={lst:math_single_turn}]
x:
  Student profile: pseudonym; overall and recent accuracy;
  per-skill performance (correct / attempted); most recent attempts on the skill
  Now solve this problem: [skill] problem text
  Multiple-choice options: A) ...; B) ...; C) ...; D) ...
  Pick the letter (A-D). Respond with only the letter.

m:
  the letter carrying the student's own recorded answer
\end{lstlisting}

\paragraph{Multi-turn record format $T_i = \{(x, m, \tau, m^*)\}$.}
The multi-turn record extends the single-turn form with a tutor turn $\tau$ and a reference letter $m^*$. Tutor guidance $\tau$ is a natural-language tutor message in one of three styles (\emph{error remediation}, \emph{Socratic}, \emph{conceptual}), generated by an LLM tutor (default model GPT-5.4). The reference $m^*$ is the letter carrying the correct answer. A Socratic record:
\begin{lstlisting}[caption={Multi-turn record schema (math), shown as a schema for the same reason.},label={lst:math_multi_turn}]
x:
  student-profile block (as above) plus the four-way multiple-choice form

m:
  the letter carrying the student's own recorded answer

tau (Socratic):
  one or two guiding questions about the step the student got wrong

m*:
  the letter carrying the canonical correct answer
\end{lstlisting}

\paragraph{Corpus statistics.}
The FA release covers $5{,}000$ students contributing $215$--$421$ interactions each ($350.7$ on average, $349$ at the median), for $1{,}753{,}384$ interactions over $3{,}368$ unique problems and $224$ unique skills. Answers are first-attempt correct on $61.6\%$ of interactions, so a student answers $129.7$ distinct problems incorrectly on average. Students are ranked by recorded volume on the same rule as L2; the $15$ Stage-2 students average $404.9$ recorded interactions each, of which $122.7$ problems are answered incorrectly and can therefore carry a correction turn. The Stage-1, Stage-2, and per-student held-out counts that draw from this corpus, after the date-remap, audit, and free-text filter described above, are summarized in \cref{tab:training_records} of Section~\ref{sec:experiment_setup}; the Stage-2 pool is nested inside the Stage-1 pool. The multi-turn held-out test set is restricted to interactions on which the student answered incorrectly.

\begin{table}[h]
\centering
\small
\caption{Mathematics corpus statistics. Stage-1 / Stage-2 training counts and per-student held-out evaluation counts ($\mathcal{F}$ test, $\mathcal{R}$ test) are reported in \cref{tab:training_records}. The shaded row is the population our simulators are trained on; the rows above it describe the release it is drawn from.}\label{tab:benchmark_math_stats}
\begin{tabular}{@{}p{0.45\linewidth} p{0.5\linewidth}@{}}
\toprule
\textbf{\boldmath Field} & \textbf{\boldmath Value} \\
\midrule
Source corpus & FoundationalASSIST (ASSISTments) \\
Curriculum & Illustrative Mathematics, Common Core aligned \\
Total students & $5{,}000$ \\
Total interactions & $1{,}753{,}384$ \\
Unique problems / unique skills & $3{,}368$ / $224$ \\
Interactions per student & $350.7$ mean, $349$ median, $215$--$421$ range \\
\rowcolor{bluegray!20} Interactions per student, the $15$ Stage-2 students & $404.9$ mean, of which $122.7$ problems answered incorrectly \\
Per-record response space & four-way MC over $\{A, B, C, D\}$ \\
\bottomrule
\end{tabular}
\end{table}

\section{Per-Benchmark Metric Definitions}\label{app:metric_definitions}

Section~\ref{sec:problem_formulation} states $\mathcal{F}$ and $\mathcal{R}$ in symbols. This appendix gives the per-benchmark formula, the operational recipe used at evaluation, and the principled reason that operationalization is the right one for that benchmark's response space. The same symbols are used throughout: $\pi_i$ (real student $i$), $M_i$ (their simulator), $S_i = \{(x, m)\}$ (held-out single-turn records), $T_i = \{(x, m, \tau, m^*)\}$ (held-out multi-turn records). The population aggregation
\begin{equation}\label{eq:pop_agg}
\mathcal{F} = \frac{1}{N}\sum_{i=1}^{N} \mathcal{F}_i, \qquad \mathcal{R} = \frac{1}{N}\sum_{i=1}^{N} \mathcal{R}_i,
\end{equation}
is the same in every benchmark; only the per-instance and per-student definitions change to track the response space.

The single principle running through the six choices below is that each $\mathcal{F}_i$ scores how closely the simulator's response on a held-out input matches what student $\pi_i$ actually recorded, and each $\mathcal{R}_i$ scores how closely the simulator's response after reading guidance $\tau$ matches the canonical corrected response $m^*$. Where a per-instance score in $[0,1]$ is well-defined from a single recorded observation, $\mathcal{F}_i$ is the mean of that score. Where the recorded supervision is categorical with an empirical candidate set, $\mathcal{F}_i$ is top-1 accuracy on that categorical set. Where the canonical corrected response is uniquely defined, $\mathcal{R}_i$ is exact match against it.

\subsection{Chess Fidelity: Top-1 Move Accuracy}\label{subsec:chess_f_def}

\paragraph{Definition.}
\begin{equation}\label{eq:chess_f}
\mathcal{F}_i \;=\; \frac{1}{|S_i|} \sum_{(x, m) \in S_i} \mathbb{1}\!\left[\,\arg\max_{m' \in \mathcal{L}(x)} P_{M_i}(m' \mid x) \;=\; m\,\right],
\end{equation}
where $\mathcal{L}(x)$ is the set of legal moves at position $x$ and $m$ is the move student $\pi_i$ actually played.

\paragraph{Computation at evaluation.}
\begin{enumerate}
  \item For each player $i$, $S_i$ is the held-out set of single-turn records (board position, recorded move). The held-out games are disjoint from training games (Appendix~\ref{app:benchmark_details}).
  \item For each $(x, m) \in S_i$, decode the simulator $M_i$'s top-1 next move on $x$ under greedy decoding; call it $\hat m$.
  \item Per-instance score is $\mathbb{1}[\hat m = m] \in \{0, 1\}$.
  \item $\mathcal{F}_i$ is the mean per-instance score over $S_i$; $\mathcal{F}$ aggregates by Equation~\eqref{eq:pop_agg}.
\end{enumerate}

\paragraph{Why this choice.}
The chess move space at any position is finite (typically tens of legal moves), and the player's record contains exactly one chosen move per position; there is no recorded distribution to compare against. Top-1 accuracy uses only what is recorded and is therefore well-posed from one observation per position. Several plausible alternatives answer a different question or require supervision that the benchmark does not record: \textit{i)} a move-distribution divergence (for example KL or JSD against the player's empirical distribution at $x$) requires repeated visits to the same position by the same player to estimate any target distribution, and Lichess records do not supply that; \textit{ii)} top-$k$ inclusion accuracy requires choosing $k$, and any $k > 1$ admits simulators that hedge across the legal-move set rather than commit to the player's move. Centipawn loss relative to the player's move measures how badly the simulator plays in absolute terms, not how closely it tracks the specific player; the latter is what $\mathcal{F}$ is for.

\subsection{Chess Guidance Responsiveness: Top-1 Against the Engine-Best Move}\label{subsec:chess_g_def}

\paragraph{Definition.}
\begin{equation}\label{eq:chess_g}
\mathcal{R}_i \;=\; \frac{1}{|T_i|} \sum_{(x, m, \tau, m^*) \in T_i} \mathbb{1}\!\left[\,\arg\max_{m' \in \mathcal{L}(x)} P_{M_i}(m' \mid x, m, \tau) \;=\; m^*\,\right],
\end{equation}
where $T_i$ contains held-out positions on which $\pi_i$ played a non-optimal move $m$, $\tau$ is the tutor guidance addressing that mistake, and $m^*$ is the engine-recommended move (Stockfish at fixed depth, treated as the canonical correction).

\paragraph{Computation at evaluation.}
\begin{enumerate}
  \item For each player $i$, $T_i$ is the held-out set of multi-turn records on positions where $\pi_i$ played a non-optimal move; the four guidance modes (error remediation, comparative, strategic, Socratic) are evenly represented.
  \item For each record, the simulator receives the three-message conversation prefix $(x, m, \tau)$ and is queried for its corrected move; we read its top-1 prediction $\hat m^*$ under greedy decoding.
  \item Per-instance score is $\mathbb{1}[\hat m^* = m^*] \in \{0, 1\}$.
  \item $\mathcal{R}_i$ is the mean per-instance score over $T_i$; $\mathcal{R}$ aggregates by Equation~\eqref{eq:pop_agg}.
\end{enumerate}

\paragraph{Why this choice.}
After receiving guidance $\tau$, the simulator's response is again a single legal move and the canonical correction $m^*$ is the engine-recommended move, so top-1 against $m^*$ is the direct comparison in the response space. The most plausible alternatives shift the measurement target away from guidance-following: \textit{i)} centipawn-loss reduction credits any move that happens to be objectively strong, even when $\tau$ explicitly steers elsewhere; \textit{ii)} improvement over $m$ is automatic for any near-random change away from a known blunder; \textit{iii)} top-$k$ inclusion admits a hedging simulator on every record whose $m^*$ is anywhere in its $k$-best list. Top-1 against $m^*$ matches the strict question $\mathcal{R}$ is meant to answer, namely whether the simulator updates to the move that the guidance is steering toward.

\subsection{L2 Fidelity: Error-Pattern Density Match}\label{subsec:l2_f_def}

\paragraph{Definition.}
For a held-out essay record $(x, m) \in S_i$, let $\hat m$ be the simulator's generated essay on prompt $x$, let $\rho_e(t)$ denote the density of LanguageTool\footnote{\url{https://languagetool.org}} issue type $e$ on essay text $t$ (errors of type $e$ per $100$ words), and let $\rho_{\text{tot}}(t) = \sum_{e \in \mathcal{C}} \rho_e(t)$ denote the total error density on $t$. Define the per-record fidelity score
\begin{equation}\label{eq:l2_f}
\mathcal{F}_i^{(x, m)} \;=\; \tfrac{1}{2}\, d\!\left(\rho_{\text{tot}}(\hat m), \, \rho_{\text{tot}}(m)\right) \;+\; \tfrac{1}{2 \, |\mathcal{C}|} \sum_{e \in \mathcal{C}} d\!\left(\rho_e(\hat m), \, \rho_e(m)\right), \qquad d(a, b) = 1 - \tfrac{|a - b|}{\max(a, b)},
\end{equation}
where $d(a, b) \in [0, 1]$ (with $d(0, 0) := 1$) is the density-similarity kernel and $\mathcal{C}$ is the set of seven LanguageTool issue types: misspelling, grammar, typographical, style, uncategorized, whitespace, inconsistency. Then $\mathcal{F}_i$ is the mean of $\mathcal{F}_i^{(x, m)}$ over $S_i$.

\paragraph{Computation at evaluation.}
\begin{enumerate}
  \item For each learner $i$, $S_i$ is the held-out set of essay prompts paired with the learner's recorded essay $m$; $\hat m$ is the simulator's free-form essay generated on $x$.
  \item Run LanguageTool on $\hat m$ and on $m$; bin issues by issue type to obtain per-category counts and a total count, divide by essay length to get densities $\rho_e$ and $\rho_{\text{tot}}$.
  \item Compute $\mathcal{F}_i^{(x, m)}$ via Equation~\eqref{eq:l2_f}.
  \item $\mathcal{F}_i$ is the mean of $\mathcal{F}_i^{(x, m)}$ over $S_i$; $\mathcal{F}$ aggregates by Equation~\eqref{eq:pop_agg}.
\end{enumerate}

\paragraph{Why this choice.}
L2 essays are free-form text, so essay-level surface match is not a fidelity signal: no two essays the same learner writes are identical at the surface, and a strong baseline that produces a clean fluent essay would score well on lexical similarity even though the recorded learner essay is full of mistakes. The signal that the metric needs to reward is the learner's overall error rate and issue-type profile. Equation~\eqref{eq:l2_f} operationalizes that signal with two terms guarding distinct failure modes: the total-density term rules out a ``too clean'' simulator that has the right error categories at near-zero rates, and the per-category mean rules out a simulator that hits the right total volume with a category profile that does not match the learner. The two-term split is symmetric and equally weighted; we do not tune the $0.5/0.5$ split. The seven LanguageTool issue types are the issue-type taxonomy LanguageTool itself emits; we do not collapse or relabel them. Surface-form alternatives (BLEU, ROUGE, chrF) reward fluency and topic-keyword overlap, both of which a strong general-purpose model satisfies independent of any learner profile; under those alternatives a simulator that ignores the learner's typical errors can score higher than one that reproduces them, which can invert the fidelity ranking on this corpus.

\subsection{L2 Guidance Responsiveness: Canonical-Correction Exact Match}\label{subsec:l2_g_def}

\paragraph{Definition.}
For each record $(x, m, \tau, m^*) \in T_i$, let $\hat m^*$ be the simulator's corrected fragment after reading $(x, m, \tau)$, and let $n(\cdot)$ denote the normalization that strips leading and trailing whitespace, lowercases, and collapses the parenthesized edit marker `(delete)' to the literal token `delete'. Then
\begin{equation}\label{eq:l2_g}
\mathcal{R}_i \;=\; \frac{1}{|T_i|} \sum_{(x, m, \tau, m^*) \in T_i} \mathbb{1}\!\left[\, n(\hat m^*) \;=\; n(m^*)\,\right].
\end{equation}

\paragraph{Computation at evaluation.}
\begin{enumerate}
  \item For each learner $i$, $T_i$ is the held-out set of multi-turn correction records, spanning the two correction modes (point-based, rule-based).
  \item For each record, the simulator receives the three-message conversation prefix $(x, m, \tau)$ and is queried for the corrected fragment; we read its free-form output $\hat m^*$.
  \item Per-instance score is $\mathbb{1}[n(\hat m^*) = n(m^*)] \in \{0, 1\}$.
  \item $\mathcal{R}_i$ is the mean per-instance score over $T_i$; $\mathcal{R}$ aggregates by Equation~\eqref{eq:pop_agg}.
\end{enumerate}

\paragraph{Why this choice.}
A tutor correction in the L2 corpus names a specific canonical corrected fragment $m^*$, and the question $\mathcal{R}$ is asking is whether the simulator produces that fragment after reading $\tau$. Normalized exact match against $m^*$ is the closest operationalization of that match under the corpus's annotation scheme; the normalization $n(\cdot)$ removes only whitespace, casing, and the syntactic edit marker, so it does not absorb genuine semantic differences. Softer alternatives are more permissive than the corpus's correction criterion: \textit{i)} BLEU or chrF on the corrected fragment scores a simulator that produces the original fragment lightly edited as nearly equivalent to one that produces the canonical correction, which mixes ``made the requested edit'' with ``made some edit lexically close to the original''; \textit{ii)} sentence-level embedding similarity rewards staying on topic. Exact match also aligns with the corpus tutor-correction protocol, which records a single accepted corrected fragment per instance.

\subsection{Math Fidelity: Top-1 Accuracy on a Four-Way Multiple-Choice Form}\label{subsec:math_f_def}

\paragraph{Definition.}
Each held-out problem $x$ for student $i$ is recast as a four-way multiple-choice instance whose option set $\mathcal{A}(x) = \{A, B, C, D\}$ contains \textit{i)} the literal answer $m$ that student $\pi_i$ recorded on this problem (placed at the letter $\ell^* \in \{A, B, C, D\}$ by a deterministic shuffle keyed on $x$) and \textit{ii)} three distractor letters, each holding an alternative answer drawn from the most common answers given by other students on the same problem (after case-insensitive string deduplication against the student's recorded answer, and excluding the 200 Stage-1 and 15 Stage-2 students themselves). The student's recorded answer is included whether it is canonically correct or wrong; on the 990-record math evaluation set $70.6\%$ of records have a canonically correct recorded answer and $29.4\%$ have a wrong one, and the simulator is asked to predict that specific letter in both cases. The per-record fidelity is top-1 accuracy of the simulator on this multiple-choice prompt,
\begin{equation}\label{eq:math_f_acc}
\mathcal{F}_i^{(x, m)} \;=\; \mathbb{1}\!\left[\,\arg\max_{\ell \in \mathcal{A}(x)} P_{M_i}(\ell \mid x) \;=\; \ell^*\,\right],
\end{equation}
and $\mathcal{F}_i$ is the mean of Equation~\eqref{eq:math_f_acc} over $S_i$. When the empirical distractor pool for a problem holds fewer than three eligible answers after deduplication, $|\mathcal{A}(x)|$ falls to $3$ for that record (we never pad with synthetic distractors); $K = 3$ occurs in $5$ of the $990$ records in the released math evaluation set.

\paragraph{Computation at evaluation.}
\begin{enumerate}
  \item For each student $i$, $S_i$ is the held-out set of math problems that the student recorded an answer to; each problem is converted to the four-way multiple-choice form once, with the per-record letter mapping $\ell^*$ fixed by a deterministic shuffle keyed on the problem id.
  \item For each record, run the simulator on the four-way multiple-choice prompt and decode greedily for a single letter token.
  \item Per-instance score is $\mathbb{1}[\hat\ell = \ell^*] \in \{0, 1\}$, where $\hat\ell$ is the simulator's decoded letter.
  \item $\mathcal{F}_i$ is the mean per-instance score over $S_i$; $\mathcal{F}$ aggregates by Equation~\eqref{eq:pop_agg}.
\end{enumerate}

\paragraph{Why this choice.}
In math, the fidelity target is whether the simulator predicts the specific answer that this student would record on this problem, given a candidate set that contains the answers students actually give on the problem. Top-1 accuracy on a four-way multiple-choice form is the direct operationalization of that target on a categorical response space. Three operational details motivate the construction. First, scoring a free-text math answer directly as raw text would make per-record likelihoods or string comparisons incomparable across simulators because they would depend on tokenization of the numerical string, on the model's prior over arbitrary number tokens, and on string-normalization choices; recasting the problem as a four-way form removes those confounds and pins the comparison to a controlled categorical set. Second, the candidate set is constructed from real student behavior on the same problem: the student's recorded answer plus the three most common other-student answers. This makes the comparison conditional on an empirical confusion set for the problem, rather than on random distractors or arbitrary string perturbations, and a correct letter prediction therefore reflects that the simulator distinguishes this student's specific answer from the answers other students actually give on the same problem. Third, top-1 accuracy is symmetric across simulators that expose logits and closed-source APIs that do not: greedy single-token decoding of a letter is supported uniformly, so every model is scored on the same observable quantity without depending on top-$k$ logprob caps (GPT-5.4 exposes only top-$5$). We use $K = 4$ rather than larger $K$ to keep the four letters distinct and to keep the empirical distractor pool well-populated on every problem; in the small fraction of records where fewer than three distinct other-student answers exist after deduplication, the record is kept with $K = 3$ rather than padded with synthetic options.

\subsection{Math Guidance Responsiveness: Top-1 Accuracy on a Four-Way Post-Guidance Multiple-Choice Form}\label{subsec:math_g_def}

\paragraph{Definition.}
For each record $(x, m, \tau, m^*) \in T_i$, the simulator reads a three-message conversation prefix consisting of \textit{i)} a problem prompt $x$ recast in four-way multiple-choice form whose option set contains the student's recorded answer $m$ and three other-student-answer distractors on the same problem, \textit{ii)} the letter $\ell_m$ that maps to $m$ under that record's deterministic shuffle, and \textit{iii)} the tutor correction $\tau$ delivered as natural-language prose, and then decodes \textit{iv)} the simulator's post-guidance letter prediction. The post-guidance candidate set $\mathcal{A}(x, \tau)$ reuses the same four-letter shuffle as $\mathcal{A}(x)$, with the canonical corrected answer $m^*$ occupying its letter $\ell^{m^*}$ in that shuffle. The per-record score is top-1 accuracy of the simulator's letter at the post-guidance turn,
\begin{equation}\label{eq:math_g}
\mathcal{R}_i \;=\; \frac{1}{|T_i|} \sum_{(x, m, \tau, m^*) \in T_i} \mathbb{1}\!\left[\,\arg\max_{\ell \in \mathcal{A}(x, \tau)} P_{M_i}(\ell \mid x, m, \tau) \;=\; \ell^{m^*}\,\right].
\end{equation}

\paragraph{Computation at evaluation.}
\begin{enumerate}
  \item For each student $i$, $T_i$ is the held-out set of multi-turn records on problems where $\pi_i$ initially produced a wrong answer $m$ and a tutor delivered a correction $\tau$; the three guidance modes (error remediation, Socratic, conceptual) are evenly represented.
  \item For each record, a single deterministic shuffle keyed on the problem id places the student's recorded answer $m$ at letter $\ell_m$ and the canonical corrected answer $m^*$ at letter $\ell^{m^*}$; the same shuffle is used at both the student turn and the post-guidance turn.
  \item The simulator receives the three-message prefix $(x_{\text{MC}}, \ell_m, \tau)$, where $\tau$ is the tutor correction delivered as natural-language prose, and decodes greedily for a single letter token $\hat\ell$.
  \item Per-instance score is $\mathbb{1}[\hat\ell = \ell^{m^*}] \in \{0, 1\}$.
  \item $\mathcal{R}_i$ is the mean per-instance score over $T_i$; $\mathcal{R}$ aggregates by Equation~\eqref{eq:pop_agg}.
\end{enumerate}

\paragraph{Why this choice.}
After reading a tutor correction that addresses the student's wrong answer, the question $\mathcal{R}$ asks is whether the simulator now arrives at the canonical corrected answer. We measure this on the same categorical four-way form used by $\mathcal{F}$ for two reasons. First, training and evaluation share the same form: the simulator is trained on multiple-choice records at both turns, and decoding a letter at evaluation matches the training distribution rather than crossing a free-text / categorical boundary that would penalize the simulator for an artifact of formatting. Second, the candidate set is constructed from the same empirical other-student answer pool used at the student turn, so the comparison is held against the answers students do actually give on this problem; the canonical corrected answer competes against three real alternative forms rather than against random distractors, and the metric rewards a simulator that resolves to the corrected form among realistic competitors. We considered free-text exact match against $m^*$ and rejected it because string normalization absorbs or fails to absorb form differences (e.g., $0.5$ versus $.5$ versus $1/2$) inconsistently across simulators, and that surface-form noise dominates the per-record signal under exact match; the four-way form with empirical distractors converts surface-form variation into the candidate set itself and lets the metric score whether the simulator selects the corrected canonical letter among the realistic alternatives.

\paragraph{Reproducibility note: bf16 nondeterminism on small samples.}
On the open-source per-student simulator (LoRA over a bf16 Qwen3-4B-Instruct base), greedy single-token decoding on borderline-logit records can flip across runs because of nondeterministic bf16 accumulation. The effect is per-record and averages out at scale: on a $315$-record cap-$7$ subsample of the math evaluation set the same checkpoint scored $\mathcal{R} = 0.565$ on a first pass and $\mathcal{R} = 0.952$ on a re-run, while on the full $879$-record no-cap evaluation set the population mean is stable across re-runs at the headline value reported in \cref{tab:guidance_main}. The closed-source baselines (GPT-4o, GPT-5.4) are queried with API temperature $0$ and are much less sensitive to within-checkpoint jitter. For faithful comparison the headline math number is reported on the full no-cap $879$-record set; we recommend reproducing on the same set rather than on a downsample.

\subsection{Per-Instruction-Mode $\mathcal{R}$ Breakdown}\label{subsec:r_per_mode}

Each domain's multi-turn evaluation set covers multiple tutor guidance modes (Section~\ref{sec:problem_formulation}). The tables below break the population-mean $\mathcal{R}$ from \cref{tab:guidance_main} down by mode within each domain. Per-mode aggregation is by mode within each per-student $T_i$ and then averaged equally across students within the mode. Per-mode totals are reported only when the held-out evaluation records partition evenly across modes within each $T_i$; in math the per-student $T_i$ ranges from $21$ to $99$ records jagged across students but the modes are evenly represented per student (counts of $293$ per mode in the $879$-record total).

\begin{table}[h]
\centering
\small
\caption{\textbf{Chess: $\mathcal{R}$ by guidance mode} across the four chess tutor modes. Population mean across the $30$ chess players' held-out multi-turn records, equal records per mode within each player; subscripts are standard deviations across the three training seeds of \cref{tab:results_std_seed}. Maia2 reads only the board position and the player rating, so the four modes present it with the same input and its $\mathcal{R}$ does not decompose by mode.}
\label{tab:r_per_mode_chess}
\begin{tabular}{l r r r r}
\toprule
\textbf{\boldmath mode} & \textbf{\boldmath Maia2} & \textbf{\boldmath GPT-4o} & \textbf{\boldmath GPT-5.4} & \textbf{\textsc{StudentSim}} \\
\midrule
error remediation & N/A & $0.9550$ & $0.9652$ & $\boldsymbol{0.9959_{\pm 0.0003}}$ \\
comparative       & N/A & $0.8778$ & $0.7696$ & $\boldsymbol{0.9581_{\pm 0.0017}}$ \\
strategic         & N/A & $0.7075$ & $0.6223$ & $\boldsymbol{0.9076_{\pm 0.0013}}$ \\
Socratic          & N/A & $0.5217$ & $0.5174$ & $\boldsymbol{0.7649_{\pm 0.0012}}$ \\
\bottomrule
\end{tabular}
\end{table}

\begin{table}[h]
\centering
\small
\caption{\textbf{L2: $\mathcal{R}$ by guidance mode} across the two L2 tutor modes (point-based, rule-based). Population mean across the 15 L2 learners' held-out multi-turn records, $20$ records per mode within each learner.}
\label{tab:r_per_mode_l2}
\begin{tabular}{l r r r r}
\toprule
\textbf{\boldmath mode} & \textbf{\boldmath base (Qwen3-4B)} & \textbf{\boldmath GPT-4o} & \textbf{\boldmath GPT-5.4} & \textbf{\textsc{StudentSim}} \\
\midrule
point-based & $0.0233$ & $0.3667$ & $0.5767$ & $\boldsymbol{0.6533_{\pm 0.0145}}$ \\
rule-based  & $0.0167$ & $0.4100$ & $0.6133$ & $\boldsymbol{0.6300_{\pm 0.0557}}$ \\
\bottomrule
\end{tabular}
\end{table}

\begin{table}[h]
\centering
\small
\caption{\textbf{Math: $\mathcal{R}$ by guidance mode} across the three math tutor modes (error remediation, Socratic, conceptual). Population mean across the 15 math students' held-out multi-turn records, $293$ records per mode in the $879$-record total (modes evenly represented within each per-student $T_i$).}
\label{tab:r_per_mode_math}
\begin{tabular}{l r r r r}
\toprule
\textbf{\boldmath mode} & \textbf{\boldmath base (Qwen3-4B)} & \textbf{\boldmath GPT-4o} & \textbf{\boldmath GPT-5.4} & \textbf{\textsc{StudentSim}} \\
\midrule
error remediation & $0.819$ & $0.877$ & $0.809$ & $\boldsymbol{0.9909_{\pm 0.0020}}$ \\
Socratic          & $0.481$ & $0.560$ & $0.655$ & $\boldsymbol{0.8783_{\pm 0.0232}}$ \\
conceptual        & $0.539$ & $0.645$ & $0.666$ & $\boldsymbol{0.8851_{\pm 0.0104}}$ \\
\bottomrule
\end{tabular}
\end{table}

Across all three domains, \textsc{StudentSim} outperforms the strongest baseline on every guidance mode, including the hardest, most indirect ones, for example chess Socratic and the math Socratic and conceptual settings. In these modes the tutor steers the student with leading questions or conceptual hints without stating which move or answer is correct, so the simulator has to infer the correction from the tutor's reasoning. That makes these modes the strictest test of guidance responsiveness.

\section{Training Pipeline Ablations and Parameter Sweeps}\label{app:ablation}

This appendix reports the chess-domain sweeps that set the default training recipe for the reference simulators, together with a robustness check on the pooled two-turn fraction. We run these sweeps only in chess. Chess is the most challenging of the three domains to retrain end-to-end (since LLMs are less familiar with playing chess than learning a second-language or doing mathematics), and it uses the same two training stages and the same guidance-conditioning mechanism as the other two domains. We then transfer the resulting recipe to second-language English writing and to mathematics without per-knob retuning. This appendix therefore establishes a default recipe, not a claim that each knob has the same optimum in every domain. We first ablate the main pipeline choice of cross-student pooling in Stage 1, then examine three recipe settings: \textit{i)} input modality (text vs.\ vision), \textit{ii)} the fraction of guidance data in pooled-training, and \textit{iii)} the composition of the guidance-mode mix.

We report two summaries. $\mathcal{F}$ on chess is the population mean of the per-student behavioral-fidelity score, measured on held-out positions on randomly selected 3 students. $\mathcal{R}$ on chess is the Turn-2 move accuracy on a held-out evaluation set of guided interactions. The board positions in this evaluation set are held out from both training stages, and the set covers all four chess guidance modes (error remediation, Socratic, strategic, comparative) with an equal number of positions per mode. The equal-count construction ensures that $\mathcal{R}$ reflects overall guidance responsiveness rather than capability on whichever mode happens to be most represented in evaluation, and it also makes per-mode breakdowns directly comparable. Both metrics are defined formally in Section~\ref{sec:fidelity_results} and Section~\ref{sec:guidance_results}. Each sweep is evaluated as a matched comparison within its own run configuration and student subset, so rows inside a table are directly comparable, while rows from different tables come from separate ablation runs.

\subsection{What Stage-1 Pooling Contributes}\label{app:ablation_cross_student_pooling}

This ablation asks what the pooled Stage-1 corpus contributes in the main pipeline, holding the optimization budget fixed. In the reference condition, Stage 1 trains on $100{,}000$ chess records pooled from $100$ players, giving $391$ optimizer steps at effective batch $256$. In the matched ablation, Stage 1 instead trains on one player's own $1{,}000$ records repeated $100$ times, which produces the same $391$ optimizer steps under the same effective batch, LoRA rank and scaling, learning rates, warmup, and seed. Stage 2 is identical in both conditions: the model is specialized on that player's own $1{,}000$ records for $12$ optimizer steps. This construction holds the update budget fixed and changes only the number of distinct students represented in Stage 1. We evaluate on three randomly sampled players.

\begin{table}[h]
\centering
\small
\caption{Ablation of cross-student pooling in chess Stage 1. The reference arm uses the default pooled Stage-1 corpus of $100{,}000$ records from $100$ players. The matched ablation replaces this with one player's own $1{,}000$ records repeated $100$ times, which preserves the Stage-1 optimizer budget ($391$ steps) and all training hyperparameters. Both arms then run the same Stage-2 specialization on that player's own $1{,}000$ records ($12$ steps). Results are averaged over three ramdomly sampled players.}
\label{tab:ablation_cross_student_pooling_chess}
\begin{tabular}{lcc}
\toprule
\textbf{\boldmath Stage-1 corpus} & \textbf{\boldmath $\mathcal{F}$} & \textbf{\boldmath $\mathcal{R}$} \\
\midrule
pooled across $100$ players \textit{(default)} & 0.5131 & 0.9003 \\
one player's own records, repeated & 0.4602 & 0.8276 \\
\bottomrule
\end{tabular}
\end{table}

Cross-student pooling is what the pooled stage contributes. In \cref{tab:ablation_cross_student_pooling_chess}, the reference arm reaches $\mathcal{F}=0.5131$ and $\mathcal{R}=0.9003$, while the repeated-single-student arm reaches $\mathcal{F}=0.4602$ and $\mathcal{R}=0.8276$. The same pattern holds for each player individually: fidelity drops from $0.4558$ to $0.3992$, from $0.4660$ to $0.4186$, and from $0.6174$ to $0.5628$, while responsiveness drops from $0.9093$ to $0.8363$, from $0.8928$ to $0.8238$, and from $0.8990$ to $0.8228$.

With Stage-1 and Stage-2 step counts, schedules, and data volume matched, the gain in the reference arm comes from the Stage-1 corpus spanning many students. Repeating one player's limited records for the same number of updates does not reproduce the effect of pooling across students.

\subsection{Input Modality: Text vs.\ Vision}\label{app:ablation_modality}

Chess is one of the three domains where a non-textual encoding of the student's working state is available: the board position can be rendered as an image and consumed by a vision-language base. We compared the text-only recipe (Qwen3-4B-Instruct, FEN plus features rendered as a string) against a matched vision-language recipe (Qwen3-VL-4B-Instruct, $400 \times 400$ board image plus the same text features), keeping every other hyperparameter fixed at the chess defaults.

\begin{table}[h]
\centering
\small
\caption{Modality sweep on chess. The board image adds $0.001$ to $\mathcal{F}$ and requires a vision tower in the training loop; the two arms are matched on effective batch size.}\label{tab:ablation_modality}
\begin{tabular}{lcc}
\toprule
\textbf{\boldmath Modality} & \textbf{\boldmath $\mathcal{F}$} & \textbf{\boldmath training cost (relative)} \\
\midrule
text-only (Qwen3-4B-Instruct) & 0.536 & $1.0 \times$ \\
vision-language (Qwen3-VL-4B-Instruct) & 0.537 & $\approx 2.0 \times$ \\
\bottomrule
\end{tabular}
\end{table}

\cref{tab:ablation_modality} shows that the visual board does not improve $\mathcal{F}$ in any meaningful sense. The textual board encoding (FEN plus the candidate-move and history features described in Section~\ref{sec:method}) already specifies the full game state relevant to move selection, so the rendered image adds little new information beyond what the text encoding already carries. We therefore use the text-only recipe in every domain. The other two domains are already text-native, so we do not repeat the modality sweep there.

\subsection{Robustness to the Pooled Two-Turn Fraction}\label{app:ablation_mt_ratio}

The pooled-training mix contains both single-turn samples (board $\to$ move) and two-turn samples (board $\to$ move; AI tutor guidance $\to$ revised move). The mixing ratio $\rho$ controls the fraction of two-turn samples in the pooled stage. We fix $\rho = 0.2$ in all production runs, and the sweep below checks whether the paper's conclusions depend on that setting.

\begin{table}[h]
\centering
\small
\caption{Multi-turn mixing-ratio sweep on chess (sequential sampler, fixed Turn-2 suffix). Adding any nonzero amount of guidance data lifts $\mathcal{R}$ from $0.17$ to above $0.80$. From there $\mathcal{R}$ stays on a plateau, and $\mathcal{F}$ is unaffected by $\rho$ within $\pm 0.004$. The production setting $\rho=0.20$ sits inside the plateau.}\label{tab:ablation_mt_ratio}
\begin{tabular}{lccc}
\toprule
\textbf{\boldmath multi-turn ratio $\rho$} & \textbf{\boldmath $\mathcal{F}$} & \textbf{\boldmath $\mathcal{R}$} & \textbf{\boldmath $\Delta\mathcal{R}$ vs.\ $\rho=0$} \\
\midrule
$0.000$ (single-turn only) & 0.490 & 0.167 & n/a \\
$0.025$ & 0.489 & 0.826 & $+0.659$ \\
$0.05$  & 0.488 & 0.849 & $+0.682$ \\
$0.10$  & 0.489 & 0.816 & $+0.649$ \\
$\mathbf{0.20}$ \textit{(default)} & \textbf{0.487} & \textbf{0.859} & $\mathbf{+0.692}$ \\
$0.40$ & 0.491 & 0.878 & $+0.711$ \\
$0.80$ & 0.487 & 0.857 & $+0.690$ \\
\bottomrule
\end{tabular}
\end{table}

Two facts stand out. First, $\mathcal{F}$ is essentially invariant in $\rho$ across the full range, so fidelity is stable under the Stage-1 multi-turn mixing ratio and no fidelity result hinges on this setting. Second, $\mathcal{R}$ reaches its high-value regime as soon as any guidance data is present and stays there, so the guidance results at $\rho = 0.2$ sit well inside a plateau. That plateau is not automatic: without a fixed Turn-2 suffix that re-asks for the updated move, some runs at intermediate $\rho$ fail on error remediation by stopping after the reasoning trace instead of emitting a move. Pairing sequential sampling with the Turn-2 suffix removes that failure mode and produces the wide plateau in \cref{tab:ablation_mt_ratio}.

\subsection{Guidance-Mode Composition}\label{app:ablation_guidance_mix}

Chess uses four guidance modes (Section~\ref{sec:guidance_results}): error remediation (the tutor diagnoses the weakness in the student's move, then motivates the correct answer with concrete consequences or alternative candidates), strategic (the tutor states a plan such as ``activate the king-side rook''), comparative (the tutor contrasts other stronger candidate moves), and Socratic (the tutor asks a question that prompts the student to derive the move themselves, e.g.\ ``what is Black threatening on the next move?''). We include error remediation because direct correction is one end of the assistance dilemma \citep{Koedinger_2007}: when a student is stuck or repeatedly failing, the appropriate tutor action is to supply the move and the reason for it, and a faithful simulator population has to cover that end of the spectrum as well. We include the Socratic mode to represent guided discovery at the indirect end of the tutoring spectrum, where the tutor prompts the student to resolve the correction. Because the tutor does not state which move to play, this mode is demanding for the simulator, which must combine the student's characteristic style with the tutor's reasoning prompt. Mixing four modes during pooled-training is meant to prevent the simulator from collapsing onto whichever mode is easiest.

We swept the per-mode composition at fixed $\rho = 0.2$ in three ways: \textit{i)} \textit{leave-one-out}, dropping each mode in turn, \textit{ii)} \textit{single-mode training}, keeping only one mode, and \textit{iii)} \textit{Socratic weight scaling}, multiplying the Socratic mode's weight while keeping the others at $1$. \cref{tab:ablation_loo}, \cref{tab:ablation_only}, and~\cref{tab:ablation_weight} report $\mathcal{R}$ broken down by evaluation mode.

\begin{table}[h]
\centering
\small
\caption{Leave-one-out sweep on chess. Each row drops one guidance mode from the pooled-training mix and evaluates on all four modes. Three of the four removals \textit{improve} overall $\mathcal{R}$, including the per-mode score for the dropped mode itself, indicating that the modes are largely complementary at training time.}\label{tab:ablation_loo}
\begin{tabular}{lccccc}
\toprule
\textbf{\boldmath training mix (weights err:strat:comp:soc)} & \textbf{\boldmath $\mathcal{R}$} & \textbf{\boldmath error rem.} & \textbf{\boldmath strategic} & \textbf{\boldmath comparative} & \textbf{\boldmath Socratic} \\
\midrule
$1$:$1$:$1$:$1$ \textit{(uniform default)} & 0.836 & 0.962 & 0.880 & 0.890 & 0.611 \\
$0$:$1$:$1$:$1$ (drop error remediation) & 0.859 & 0.969 & 0.864 & 0.919 & 0.683 \\
$1$:$1$:$0$:$1$ (drop comparative) & 0.869 & 0.984 & 0.878 & 0.934 & 0.682 \\
$1$:$1$:$1$:$0$ (drop Socratic) & 0.849 & 0.986 & 0.884 & 0.903 & 0.621 \\
$1$:$0$:$1$:$1$ (drop strategic) & 0.820 & 0.926 & 0.854 & 0.846 & 0.654 \\
\bottomrule
\end{tabular}
\end{table}

\begin{table}[h]
\centering
\small
\caption{Single-mode training on chess. Each row trains the pooled-training mix on one guidance mode only and evaluates on all four. Even one-mode training reaches $\mathcal{R}\in[0.72, 0.80]$, against $0.836$ for the four-mode default. Comparative-only transfers best to the other three modes, and error-remediation-only transfers worst.}\label{tab:ablation_only}
\begin{tabular}{lccccc}
\toprule
\textbf{\boldmath training mode (single mode at full weight)} & \textbf{\boldmath $\mathcal{R}$} & \textbf{\boldmath error rem.} & \textbf{\boldmath strategic} & \textbf{\boldmath comparative} & \textbf{\boldmath Socratic} \\
\midrule
all four \textit{(uniform default, ref)} & 0.836 & 0.962 & 0.880 & 0.890 & 0.611 \\
comparative only & 0.798 & 0.926 & 0.850 & 0.865 & 0.552 \\
strategic only & 0.790 & 0.859 & 0.857 & 0.883 & 0.562 \\
Socratic only & 0.769 & 0.864 & 0.816 & 0.878 & 0.519 \\
error remediation only & 0.720 & 0.898 & 0.751 & 0.781 & 0.449 \\
\bottomrule
\end{tabular}
\end{table}

\begin{table}[h]
\centering
\small
\caption{Weight-scaling sweep on chess: scaling the Socratic mode while keeping the other three at $1$. Multiplying the Socratic weight raises the Socratic per-mode score from $0.611$ to $0.670$ ($+5.9$\,pp at $4\times$). The other three modes shift by a few points at the strongest setting, and overall $\mathcal{R}$ is close to flat.}\label{tab:ablation_weight}
\begin{tabular}{lccccc}
\toprule
\textbf{\boldmath training mix (weights err:strat:comp:soc)} & \textbf{\boldmath $\mathcal{R}$} & \textbf{\boldmath error rem.} & \textbf{\boldmath strategic} & \textbf{\boldmath comparative} & \textbf{\boldmath Socratic} \\
\midrule
$1$:$1$:$1$:$1$ \textit{(uniform default)} & 0.836 & 0.962 & 0.880 & 0.890 & 0.611 \\
$1$:$1$:$1$:$2$ (boost Socratic $2\times$) & 0.842 & 0.972 & 0.875 & 0.884 & 0.635 \\
$1$:$1$:$1$:$4$ (boost Socratic $4\times$) & 0.841 & 0.924 & 0.864 & 0.904 & 0.670 \\
\bottomrule
\end{tabular}
\end{table}

Three findings come out of this chess composition sweep.

\textit{Modes partly substitute for one another during pooled-training.} In \cref{tab:ablation_loo}, removing error remediation, comparative, or Socratic does not lower that mode's evaluation score, and in two of those three cases the score on the dropped mode actually rises. The most extreme case is comparative: removing it from training raises the comparative evaluation score from $0.890$ to $0.934$. The four guidance modes share enough underlying structure (extracting a chess move that is consistent with both the student's prior and the AI tutor's pedagogical signal) that training on three of them transfers to the fourth. Strategic is the one exception: its removal lowers overall $\mathcal{R}$, which suggests that strategic carries pedagogical signal the other three modes do not fully cover.

\textit{A model trained on one mode still answers the other three.} \cref{tab:ablation_only} pushes the substitutability claim further: training the pooled-training mix on a single guidance mode and evaluating on all four still gives $\mathcal{R} \in [0.72, 0.80]$, compared with $0.836$ for the four-mode default. The transfer is asymmetric. Comparative-only training transfers best (overall $\mathcal{R} = 0.798$), consistent with comparative being the most redundant mode in the leave-one-out sweep. Error-remediation-only training transfers worst (overall $\mathcal{R} = 0.720$). This mode delivers a full diagnosis of the mistake together with the correction and its rationale, and naming the move comes with that direct-correction role; a model trained on it alone can settle for copying the named move, which does not carry over to the other three modes, where the tutor does not state which move to play. The Socratic per-mode score also stays low ($\leq 0.56$) in every single-mode row, so Socratic capability is the hardest piece to acquire by transfer alone and benefits most from being explicitly represented in the training mix.

\textit{Per-mode capability is controllable through training-data weights.} In \cref{tab:ablation_weight}, scaling the Socratic weight scales the Socratic per-mode score in a graded way, with the other three modes shifting by a few points at the strongest setting. The per-mode weighting handle is useful when a downstream experiment needs a different per-mode profile: a simulator that is especially responsive to Socratic guidance, or that under-responds to it to model a student who ignores reasoning prompts, can be produced by adjusting the pooled-training mix rather than by adding new training procedures or new objectives. The reference simulators reported in the main paper use the uniform $1{:}1{:}1{:}1$ default; the weight-scaling result is recorded here only to document that the per-mode profile can be adjusted predictably.

\section{Training and Experimental Details}\label{app:training_details}

This appendix records the hyperparameters and infrastructure that produced every checkpoint and metric reported in Section~\ref{sec:fidelity_results}, Section~\ref{sec:guidance_results}, and Section~\ref{sec:tutor_rl}. Data sources, training-instance formats, and corpus statistics are in Appendix~\ref{app:benchmark_details}; metric definitions are in Appendix~\ref{app:metric_definitions}; sweep results that motivated the recipe are in Appendix~\ref{app:ablation}. A reader who reproduces the source corpora and follows the values in the tables below should be able to retrain the full reference family and rerun every evaluation reported in the main paper.

The base model in all three domains is Qwen3-4B-Instruct, specifically the \texttt{Qwen3-4B-Instruct-2507} release \citep{yang2025qwen3technicalreport}. Training is in \texttt{bf16} on a single $8 \times $ A100 (40\,GB) node using ms-swift \citep{Zhao_2025}. Both stages train LoRA adapters \citep{hu2021loralowrankadaptationlarge}; Stage 1 trains a fresh LoRA on the pooled-training corpus, and Stage 2 loads the Stage-1 adapter and continues training that adapter on one student's records with a smaller learning rate. Stage 2 therefore inherits the Stage-1 LoRA architecture (rank, $\alpha$, dropout, target modules) in all three domains. The single random seed is the ms-swift trainer default (\texttt{42}) for optimizer state and dataloader shuffling, with a separate seed (\texttt{65}) for the data sampler that constructs the per-domain training-instance pool.

\paragraph{Two-stage training and shared design.}
The two-stage design separates a Stage-1 \textit{pooled training} phase that pools records across a large student set from a Stage-2 \textit{specialization} phase that adapts the resulting LoRA adapter to one student's records. The two stages share the base model (Qwen3-4B-Instruct), LoRA rank, $\alpha$, dropout, target modules, optimizer (AdamW with $\beta = (0.9, 0.95)$), bf16 precision with gradient checkpointing, weight decay, gradient-norm clip, max sequence length, loss masking, trainer seed, and data-sampler seed; they differ only in learning rate, schedule, warmup length, and the per-domain values of corpus size, effective batch, and optimization length. Within each stage we keep a single shared configuration across chess, L2, and math; the per-domain values that differ (\cref{tab:training_stage1_domain} and~\cref{tab:training_stage2_domain}) are dictated by per-domain dataset scale, not tuned for any single benchmark. Every deviation from the shared configuration is enumerated in Section~\ref{app:training_stage2}.

\subsection{Stage 1 Hyperparameters}\label{app:training_stage1}

Stage 1 trains one domain-specific LoRA on the pooled-training corpus described in Appendix~\ref{app:benchmark_details}, mixing single-turn and multi-turn records at a fixed multi-turn ratio of $\rho = 0.20$ in all three domains; Appendix~\ref{app:ablation_mt_ratio} shows that the results do not depend on this setting. \cref{tab:training_stage1_shared} lists the settings shared across chess, second-language English writing (L2), and mathematics (math); \cref{tab:training_stage1_domain} lists the per-domain corpus size and resulting optimization length.

\begin{table}[h]
\centering
\small
\caption{Shared pooled-training-stage hyperparameters across chess, L2, and math.}
\label{tab:training_stage1_shared}
\begin{tabular}{@{}ll@{}}
\toprule
\textbf{\boldmath Setting} & \textbf{\boldmath Value} \\
\midrule
Base model & Qwen3-4B-Instruct-2507 \citep{yang2025qwen3technicalreport} \\
Training framework & ms-swift \citep{Zhao_2025} \\
Precision & bf16 with gradient checkpointing \\
Hardware & 1 node, $8 \times$ A100 (40\,GB) \\
Adapter type & LoRA \citep{hu2021loralowrankadaptationlarge} \\
LoRA rank $r$ & $128$ \\
LoRA $\alpha$ & $256$ \\
LoRA dropout & $0.05$ \\
Target modules & all linear projections (q, k, v, o, gate, up, down) \\
Optimizer & AdamW \\
Learning rate & $1\!\times\!10^{-4}$ \\
Schedule & cosine, $100$ warmup steps \\
Weight decay & $0.1$ \\
Gradient-norm clip & $1.0$ \\
Per-device batch & $4$ \\
Gradient accumulation & $8$ \\
Effective batch & $256$ \\
Max sequence length & $4096$ \\
Epochs & $1.0$ (L2 uses $3.0$) \\
Multi-turn ratio $\rho$ & $0.20$ \\
Loss masking & ignore tokens inside empty \texttt{<think></think>} \\
Corpus-construction sampler seed & $65$ \\
\bottomrule
\end{tabular}
\end{table}

\begin{table}[h]
\centering
\small
\caption{Per-domain pooled-training-stage corpus size and realized optimization length. Corpus sizes match \cref{tab:training_records} in Section~\ref{sec:experiment_setup}; the multi-turn fraction within each corpus is the shared $\rho = 0.20$ from \cref{tab:training_stage1_shared}. Wall-clock times are rough estimates.}
\label{tab:training_stage1_domain}
\begin{tabular}{@{}lrrr@{}}
\toprule
\textbf{\boldmath Domain} & \textbf{\boldmath training instances} & \textbf{\boldmath total grad steps} & \textbf{\boldmath wall-clock time} \\
\midrule
chess & $100{,}000$ & $391$ & ${\sim}\,75$ min \\
L2    & $7{,}800$   & $93$  & ${\sim}\,25$ min \\
math  & $23{,}400$  & $91$  & ${\sim}\,30$ min \\
\bottomrule
\end{tabular}
\end{table}

In all runs, LoRA is applied to all linear projections (\texttt{q\_proj}, \texttt{k\_proj}, \texttt{v\_proj}, \texttt{o\_proj}, \texttt{gate\_proj}, \texttt{up\_proj}, \texttt{down\_proj}). Loss masking ignores tokens inside empty \texttt{<think></think>} spans. The chess and math Stage-1 grad-step counts in \cref{tab:training_stage1_domain} are induced directly by the shared single-epoch setting; L2 Stage 1 is the one deviation: its pooled corpus is small enough that a single epoch gives too few optimizer steps, so it trains for $3$ epochs, which is why its row differs.

\subsection{Stage 2 Hyperparameters}\label{app:training_stage2}

In the per-student specialization stage, each run loads the Stage-1 adapter via the ms-swift \texttt{--adapters} interface and continues training the same LoRA on one student's records. The LoRA architecture, target modules, precision, max sequence length, loss masking, weight decay, gradient-norm clip, and seeds are inherited from Stage 1. \cref{tab:training_stage2_shared} lists the shared Stage-2 hyperparameters; \cref{tab:training_stage2_domain} lists the per-domain values that remain specific to each domain.

\begin{table}[h]
\centering
\small
\caption{Shared per-student-specialization-stage hyperparameters.}
\label{tab:training_stage2_shared}
\begin{tabular}{@{}ll@{}}
\toprule
\textbf{\boldmath Setting} & \textbf{\boldmath Value} \\
\midrule
Hardware & 1 node, $8 \times$ A100 (40\,GB) \\
Initialization & load Stage-1 adapter via ms-swift \texttt{--adapters} (no re-init) \\
Optimizer & AdamW (\texttt{adamw\_torch\_fused}), fresh state \\
Learning rate & $5\!\times\!10^{-5}$ \\
Schedule & cosine, $30$ warmup steps \\
\bottomrule
\end{tabular}
\end{table}

\begin{table}[h]
\centering
\small
\caption{Per-domain per-student-specialization-stage settings. ``Per-student instances'' is the training-set size used for one student's specialization run; for a given domain this size is constant across the per-student adapters in our reference family. ``Total steps'' is the number of optimizer steps in one student's specialization run. The chess and math effective batch of $256$ is realized as per-device batch $4 \times $ gradient accumulation $8 \times 8$ GPUs; the L2 effective batch of $8$ is per-device batch $1 \times $ no accumulation $\times 8$ GPUs.}
\label{tab:training_stage2_domain}
\begin{tabular}{@{}lrrrrrr@{}}
\toprule
\textbf{\boldmath Domain} & \textbf{\boldmath students} & \textbf{\boldmath per-student instances} & \textbf{\boldmath Stage-2 $\rho$} & \textbf{\boldmath epochs} & \textbf{\boldmath effective batch} & \textbf{\boldmath total steps} \\
\midrule
chess & $30$ & $1{,}000$ & $0.20$ & $3.0$ & $256$ & $12$ \\
L2    & $15$ & $73$       & $0.20$ & $3.0$ & $8$   & ${\sim}\,30$ \\
math  & $15$ & $153$      & $0.20$ & $3.0$ & $256$ & $3$ \\
\bottomrule
\end{tabular}
\end{table}

The Stage-2 learning rate is half of Stage 1 and the warmup is shorter ($30$ vs.\ $100$) because Stage 2 starts from the Stage-1 adapter rather than from the base model, and per-student datasets are about two orders of magnitude smaller than the pooled corpus in all three domains. Chess and math keep the Stage-1 effective batch ($256$); L2 uses an effective batch of $8$ (per-device batch $1$, no gradient accumulation, $8$ GPUs), which at $73$ per-student records and $3$ epochs yields approximately $30$ optimizer steps, sufficient to specialize the warm-started Stage-1 LoRA to each learner's error profile. L2 Stage 2 mixes single-turn essay records and multi-turn correction records at $\rho = 0.2$ ($15$ multi-turn plus $58$ single-turn per student), so the per-student adapter continues to receive multi-turn signal beyond what Stage 1 supplied; the L2 guidance-responsiveness numbers in Section~\ref{sec:guidance_results} reflect both stages' multi-turn training.

The saved Stage-2 adapter for every per-student simulator carries four LoRA fields ($r = 128$, $\alpha = 256$, dropout $= 0.05$, target modules $=$ all linear projections) matching \cref{tab:training_stage1_shared}, confirming inheritance from Stage 1 across all three domains.

\paragraph{Per-domain deviations and rationale.}\label{subsec:per_domain_deviations}
Chess Stage 1 and Stage 2 follow the shared defaults (1 epoch Stage 1, 3 epochs Stage 2, effective batch 256, decoding budget 32 tokens, repetition penalty 1.0). L2 and math each deviate in a small number of places summarized below; each deviation is forced by the per-domain dataset scale.

\begin{itemize}[leftmargin=*]
\item \textbf{L2 Stage 1 epochs: 3 (default 1).} At the L2 pooled-corpus size of $7{,}800$ instances and effective batch $256$, one epoch yields only $31$ optimizer steps. Three epochs bring the total to $93$ steps, comparable in magnitude to chess ($391$) and math ($91$) at a single epoch.
\item \textbf{L2 Stage 2 effective batch: 8 (default 256),} realized as per-device batch $1 \times $ no gradient accumulation $\times 8$ GPUs. With only $73$ records per learner, batch $256$ does not produce even one full optimizer step per epoch; batch $8$ gives approximately $30$ steps over the $3$-epoch run.
\item \textbf{L2 decoding budget: \texttt{max\_new\_tokens} $256$ (default $32$), \texttt{repetition\_penalty} $1.1$ (default $1.0$).} L2 outputs are full essays of roughly $50$ to $150$ words; the larger token budget accommodates the longer outputs, and a small repetition penalty prevents greedy-decoding loops past the natural essay end.
\item \textbf{Math Stage 2 optimization length: $3$ optimizer steps total} (chess $12$, L2 $\approx 30$). At $153$ per-student instances (\cref{tab:training_stage2_domain}), the shared effective batch $256$ over $3$ epochs yields exactly $3$ steps, the maximum the data supports without departing from the shared batch size. Because each student contributes few records, the Stage-2 optimization horizon is short in all three domains, and within this regime the trainer seed has little effect on run-to-run variance. For example, in math the observed std across runs (\cref{tab:results_std_seed}) is dominated by the data-sampler seed.
\end{itemize}

\subsection{Evaluation Protocol}\label{app:training_evaluation}

Decoding for both behavioral fidelity (single-turn, on $S_i$) and guidance responsiveness (multi-turn, on $T_i$) is greedy ($T = 0$) under the per-domain budgets in \cref{tab:training_eval_decoding}. The chat-template flag \texttt{enable\_thinking} is pinned to \texttt{False} at decode time. Without that flag, Qwen3's chat template defaults to opening a \texttt{<think>} block that is only closed if the model emits the matched closing tag within \texttt{max\_new\_tokens}; under the short-answer budgets used for chess and math the model otherwise stays inside the reasoning block and never produces the answer token. The flag is set identically at training time (Stage 1, Stage 2, and the tutor RL rollouts in Appendix~\ref{app:training_rl}) so train and inference agree. The same decoding budgets are applied to the trained simulators and to the prompted closed-source baselines unless a metric requires token log-probabilities rather than free-form decoding.

\begin{table}[h]
\centering
\small
\caption{Per-domain decoding settings at evaluation. All decodes are greedy ($T = 0$), and the token budget applies to fidelity and guidance evaluation alike.}
\label{tab:training_eval_decoding}
\begin{tabular}{@{}lll@{}}
\toprule
\textbf{\boldmath Domain} & \textbf{\boldmath \texttt{max\_new\_tokens}} & \textbf{\boldmath \texttt{repetition\_penalty}}\\
\midrule
chess & $32$ & $1.0$ \\
L2    & $256$ & $1.1$ \\
math  & $32$  & $1.0$ \\
\bottomrule
\end{tabular}
\end{table}

The L2 \texttt{max\_new\_tokens} budget is larger because L2 outputs are full essays of roughly $50$ to $150$ words; chess outputs are a single move and math outputs a short numerical answer. The L2 repetition penalty of $1.1$ applies when decoding an essay, where pure-greedy decoding occasionally entered repeat loops past the natural essay end; the penalty removes that decoding artifact without changing the per-token error profile that L2 fidelity scores.

\paragraph{Held-out sets.}
Each per-student evaluation uses the frozen \textsc{StudentSimEval} held-out sets (Appendix~\ref{app:studentsimeval_protocol}), which follow the $(S_i, T_i)$ schema from Section~\ref{sec:problem_formulation}. The single-turn set $S_i$ is sampled from the same data stream as training but on disjoint records (different game positions in chess, different essays in L2, different problems in math; per-domain split protocols in Appendix~\ref{app:benchmark_details}). The multi-turn set $T_i$ is similarly disjoint. For the headline numbers in Section~\ref{sec:fidelity_results} and Section~\ref{sec:guidance_results}, $|S_i|$ is $5{,}000$ positions per chess player, $26$ essays per L2 student, and $66$ problems per math student (uniform); $|T_i|$ is $4{,}000$ multi-turn records per chess player ($1{,}000$ per guidance mode), $40$ multi-turn records per L2 student ($20$ per guidance mode), and jagged across math students with a per-student range of $21$ to $99$ (mean $59$, evenly split across the three math guidance modes within each student). Per-domain $|S_i|$ and $|T_i|$ counts are tabulated in Appendix~\ref{app:benchmark_details}.

\paragraph{Closed-source baseline scoring.}
GPT-4o and GPT-5.4 are queried through the Azure OpenAI API with greedy decoding ($T = 0$) and the same per-token budgets as \cref{tab:training_eval_decoding}. For math fidelity and math guidance responsiveness, which are top-1 letter accuracy on a four-way multiple-choice form (Appendix~\ref{subsec:math_f_def}, Appendix~\ref{subsec:math_g_def}), the baselines decode a single letter token and are scored by exact letter match.

\paragraph{Random seeds and number of runs.}
Headline numbers in Section~\ref{sec:fidelity_results} and Section~\ref{sec:guidance_results} are the cross-seed mean over $S = 3$ independent training runs per domain that vary both the trainer seed and the data-sampler seed; each run retrains the full pipeline (Stage 1 + Stage 2) from scratch and re-evaluates on the same held-out set. The run-to-run standard deviation across these seeds is reported in \cref{tab:results_std_seed} (Appendix~\ref{app:results_std}). Decoding is greedy with $T = 0$ throughout.

\subsection{Run-to-Run Standard Deviation}\label{app:results_std}

\cref{tab:fidelity_main} and \cref{tab:guidance_main} report population means only. \cref{tab:results_std_seed} reports the run-to-run standard deviation of the population means across multiple training runs with different random seeds (trainer seed and data-sampler seed both varied). This measures method-level variance: how much the headline number moves if we retrain from scratch with a different seed. The closed-source baselines and the naive Qwen3-4B-Instruct baseline are not trained, so there are no training seeds to vary and this retrain-from-scratch variance does not apply to them; the run-to-run column is therefore omitted for them, and only \textsc{StudentSim} carries a run-to-run std.

\begin{table}[h]
\centering
\small
\caption{\textbf{Run-to-run standard deviations for \textsc{StudentSim}.} Each cell is the \textsc{StudentSim} population mean for that domain and metric, with the across-seed standard deviation appended as a subscript. Computed from $S$ independent training runs with different trainer seeds and data-sampler seeds; each run retrains the full pipeline (Stage 1 + Stage 2) from scratch and re-evaluates on the same held-out set. The naive Qwen3-4B-Instruct baseline and the closed-source baselines (GPT-4o, GPT-5.4) are not trained, so there are no training seeds to vary and this table's retrain-from-scratch std does not apply to them; they are omitted.}
\label{tab:results_std_seed}
\setlength{\tabcolsep}{3pt}
\begin{tabular}{l c c c}
\toprule
\textbf{\boldmath Domain} & \textbf{\boldmath $S$ (\# seeds)} & \textbf{\textsc{StudentSim} $\mathcal{F}$} & \textbf{\textsc{StudentSim} $\mathcal{R}$} \\
\midrule
chess (30 players)  & $3$ & $\boldsymbol{0.5150_{\pm 0.0009}}$ & $\boldsymbol{0.9067_{\pm 0.0009}}$ \\
L2 (15 learners)    & $3$ & $\boldsymbol{0.5624_{\pm 0.0046}}$ & $\boldsymbol{0.6417_{\pm 0.0233}}$ \\
math (15 students)  & $3$ & $\boldsymbol{0.6384_{\pm 0.0044}}$ & $\boldsymbol{0.9181_{\pm 0.0099}}$ \\
\bottomrule
\end{tabular}
\end{table}

\subsection{Tutor RL: SFT Base, GRPO Setup, and Evaluation}\label{app:training_rl}

Section~\ref{sec:tutor_rl} reports a chess proof of concept that compares three tutor-RL rewards: the supervised-finetuned tutor with no RL, a frontier LLM (GPT-5.4) prompted as the student simulator, and our trained \textsc{StudentSim} with extensible heads. The two RL conditions start from the same SFT checkpoint and continue with GRPO \citep{shao2024deepseekmathpushinglimitsmathematical}. The implementation is verl \citep{sheng2024hybridflowflexibleefficientrlhf} on a single $8\times$ A100 (40\,GB) node, over a chess RL playground built from held-out positions for the $30$ chess players (Appendix~\ref{app:benchmark_details}). The conditions share the SFT base and every GRPO hyperparameter and differ only in the reward model queried per rollout. The two RL conditions are compared at their final checkpoint, after the $20$ GRPO steps in \cref{tab:training_rl_grpo}. This appendix documents the SFT base (\cref{tab:training_rl_sft}), the GRPO settings (\cref{tab:training_rl_grpo}), the reward definitions (\cref{tab:training_rl_reward}), and the expert human evaluation (\cref{tab:training_rl_human}).

\paragraph{SFT base (the no-RL condition, and the initialization for the two RL conditions).}
The tutor is Qwen3-VL-8B-Instruct supervised-finetuned with LoRA via ms-swift on rendered chess positions paired with reference tutor guidance; hyperparameters are in \cref{tab:training_rl_sft}. The SFT data recipe combines four ingredients that together cut factual errors on this multimodal task: \textit{i)} the board image is rendered with the side-to-move at the bottom, \textit{ii)} the position is grounded piece-by-square in the prompt text, \textit{iii)} reference-answer filters drop instances whose guidance contains an illegal (hallucinated) move, makes a piece-on-square claim that contradicts the board, or quotes engine metadata such as a numeric evaluation, and \textit{iv)} the prompt carries a color hint and a generic system message. We select the step-$704$ checkpoint, merge the LoRA adapter into the base weights, and use the merged checkpoint directly as the no-RL condition and as the policy initialization for the two RL conditions.

\begin{table}[h]
\centering
\small
\setlength{\tabcolsep}{4pt}
\caption{SFT hyperparameters for the tutor base (the no-RL condition, and the initialization for the two RL conditions).}
\label{tab:training_rl_sft}
\begin{tabular}{@{}lp{0.55\linewidth}@{}}
\toprule
\textbf{\boldmath Setting} & \textbf{\boldmath Value} \\
\midrule
Base model & Qwen3-VL-8B-Instruct \citep{bai2025qwen3vltechnicalreport} \\
Framework & ms-swift, LoRA \\
LoRA rank / $\alpha$ & $32$ / $64$ (all linear projections) \\
Optimizer / LR & AdamW / $1\!\times\!10^{-4}$ \\
Epochs & $1$ \\
Effective batch & $256$ \\
Max sequence length & $2048$ \\
Max image pixels & $448\times448$ \\
Selected checkpoint & step $704$, LoRA merged into base \\
\bottomrule
\end{tabular}
\end{table}

\paragraph{GRPO settings.}
The two RL conditions use the identical GRPO configuration in \cref{tab:training_rl_grpo} and differ only in the reward model.

\begin{table}[h]
\centering
\small
\setlength{\tabcolsep}{4pt}
\caption{Shared GRPO hyperparameters for the chess tutor proof of concept; the two RL conditions use these values and differ only in the reward model (\cref{tab:training_rl_reward}).}
\label{tab:training_rl_grpo}
\begin{tabular}{@{}lp{0.55\linewidth}@{}}
\toprule
\textbf{\boldmath Setting} & \textbf{\boldmath Value} \\
\midrule
RL framework & verl \citep{sheng2024hybridflowflexibleefficientrlhf} \\
Advantage estimator & GRPO (critic-free; no value model) \\
Group size (rollouts per prompt) & $4$ \\
KL & \texttt{use\_kl\_loss} True, \texttt{kl\_loss\_type} low-variance, coefficient $0.01$; \texttt{use\_kl\_in\_reward} False \\
Entropy coefficient & $0$ \\
Actor opt / LR & AdamW / $1\!\times\!10^{-6}$ \\
Train batch / mini-batch & $112$ / $112$ \\
Sharding & FSDP2 with parameter and optimizer offload \\
Rollout engine & vLLM, max model length $2048$ \\
Training length & $20$ steps \\
\bottomrule
\end{tabular}
\end{table}

\paragraph{Reward definitions.}
The three conditions differ only in how a rollout is scored (\cref{tab:training_rl_reward}). Both RL rewards build on a shared move-quality term: given the student's wrong move and the simulator's revised move after reading the tutor's guidance, the centipawn difference is capped at $\pm 1500$ to keep mate-changing moves from saturating the signal, then squashed through a $\tanh$ with a $500$-centipawn shoulder so that typical move improvements live in the linear regime; an illegal or missing move scores $-1.0$. Centipawn values come from a precomputed Stockfish depth-$15$ lookup over the playground positions. The conditions differ in who produces the revised move and in what shapes it: our \textsc{StudentSim} reward uses our trained \textsc{StudentSim} as the simulator and multiplies the move-quality base by a style gate and a perception gate, both read off the frozen simulator backbone, whereas the GPT-5.4-simulator reward uses GPT-5.4 prompted as the student.

\begin{table}[h]
\centering
\small
\setlength{\tabcolsep}{4pt}
\caption{Reward definitions for the three conditions. The move-quality term (cp $=$ Stockfish centipawns) is shared by the two RL rewards. Our \textsc{StudentSim} reward multiplies the move-quality base by a style gate and a perception gate; the $2.0$ and $3.0$ are the gates' $\alpha$ decay rates (defined below the table).}
\label{tab:training_rl_reward}
\begin{tabular}{@{}lp{0.66\linewidth}@{}}
\toprule
\textbf{\boldmath Condition} & \textbf{\boldmath Reward} \\
\midrule
No RL & none (SFT base, no reinforcement learning) \\
GPT-5.4 simulator & with GPT-5.4 prompted as the student to produce the revised move; one frontier-model API call per rollout \\
Ours & move-quality (our \textsc{StudentSim} simulator) $\times$ style gate $\times$ perception gate, one scalar per rollout \\
\midrule
move-quality term & $\tanh\!\bigl(\mathrm{clip}(q_\text{post} - q_\text{wrong},\, \pm 1500) / 500\bigr)$, units in cp \\
illegal / missing move & $-1.0$ \\
quality lookup & precomputed Stockfish table at depth $15$ on playground positions \\
\textsc{StudentSim} simulator & chess Stage-1 LoRA on Qwen3-4B-Instruct-2507 (\cref{tab:training_stage1_shared}), served locally \\
style gate & $\exp\!\bigl(-2.0\,(1 - p_\text{pref})\bigr)$; $p_\text{pref}$ is the style head's probability of the preferred style \\
perception gate & $\exp\!\bigl(-3.0\textstyle\sum_i w_i p_i\bigr)$; $p_i$ are the six board-error probabilities, $w_i$ their per-class F1 scores normalized to sum to one \\
\bottomrule
\end{tabular}
\end{table}

\paragraph{Reward heads.}
The style and perception heads are two linear probes on the same frozen chess Stage-1 \textsc{StudentSim} simulator that produces the student's revised move: only these two heads are trained, while the simulator's weights and its move prediction are inherited unchanged from Stage 1, so the entire reward is computed from one on-node model. Both heads read the pooled hidden states over the tutor's explanation. The style head classifies the explanation into one of four teaching styles (error remediation, Socratic, strategic, comparative), distilled from a TF-IDF logistic-regression style classifier; its gate rewards adherence to the run's preferred style, set to Socratic in the reported runs to match the Socratic-leaning student used in the human-study personalization phase. The perception head produces six independent probabilities for board-grounding errors in the explanation (wrong square, wrong piece, wrong color, hallucinated piece, wrong capture, illegal move), trained against a \texttt{python-chess} rule verifier and GPT-5.4 labels; its gate penalizes explanations that misdescribe the position, weighting the six error types by their per-class F1. Both gates lie in $(0, 1]$ and multiply the move-quality base (\cref{tab:training_rl_reward}), so an on-style, board-accurate explanation leaves the base intact while an off-style or board-misdescribing one discounts it.

\paragraph{GPU layout and serving.}
GRPO uses $7$ trainer GPUs (FSDP2 with parameter and optimizer offload) and $1$ GPU that hosts the reward model. For our \textsc{StudentSim} reward, that GPU serves the local \textsc{StudentSim} simulator together with the style and perception heads, so the entire reward is computed on-node with no external dependency. For the GPT-5.4-simulator reward, the reward instead issues a GPT-5.4 API call on every rollout to obtain the simulated student's revised move; this frontier-model dependency, absent from our reward, is the practical cost of the GPT-5.4-simulator reward at training time. The no-RL baseline runs no RL and needs no reward GPU.

\paragraph{Expert Human Study.}
The evaluators are competitive chess players screened by an internal rating assessment; their ratings run up to $2200$, and the three rated $2000$ or above form an expert subset that we report separately (\cref{tab:training_rl_human}). They rate each tutor's response, blind to condition and presentation order, in three phases. Phase 1 (accuracy): annotators mark each sentence of the tutor's guidance for factual correctness, and accuracy is the fraction of responses with no sentence flagged as a severe, actively misleading error. Phase 2 (teaching quality) is a $1$--$5$ rating of guidance quality. Phase 3 (personalization) is a $1$--$5$ rating in the context of a Socratic-leaning student. The study collected $74$ finalized annotations from these evaluators; \cref{tab:training_rl_human} reports all annotations, a cleaner triply-annotated ($k{=}3$) subset, and the $2000$+ expert subset.

\begin{table}[h]
\centering
\small
\setlength{\tabcolsep}{6pt}
\caption{\textbf{Human evaluation of the AI chess tutors.} \emph{Accuracy} is the percentage of responses with no severe (actively misleading) factual error; \emph{Guidance} and \emph{Personalization} are $1$--$5$ ratings. ``Global'' is all annotations; ``$k{=}3$'' is the triply-annotated subset; ``Experts $\geq 2000$'' restricts to the three evaluators rated $2000$ or above. Each annotation rates every condition, so $n$ is the number of annotations behind each row of a view: $74$ annotations over $30$ positions from $8$ evaluators (Global), $48$ over the $16$ positions that three evaluators each annotated ($k{=}3$), and $30$ from the three highest-rated evaluators (Experts $\geq 2000$). Our \textsc{StudentSim} reward ranks first on all three axes in all three views.}
\label{tab:training_rl_human}
\begin{tabular}{@{}llrrr@{}}
\toprule
\textbf{\boldmath View} & \textbf{\boldmath Condition} & \textbf{\boldmath Accuracy (\%)} & \textbf{\boldmath Guidance ($1$--$5$, $\uparrow$)} & \textbf{\boldmath Personalization ($1$--$5$, $\uparrow$)} \\
\midrule
\multirow{3}{*}{\shortstack[l]{Global\\$(n{=}74)$}}
 & No RL           & $75.7$ & $2.99$ & $2.80$ \\
 & GPT-5.4 reward  & $71.6$ & $3.08$ & $2.42$ \\
 & Ours            & $\boldsymbol{90.5}$ & $\boldsymbol{3.31}$ & $\boldsymbol{3.93}$ \\
\midrule
\multirow{3}{*}{\shortstack[l]{$k{=}3$\\$(n{=}48)$}}
 & No RL           & $75.0$ & $3.10$ & $2.65$ \\
 & GPT-5.4 reward  & $70.8$ & $3.21$ & $2.33$ \\
 & Ours            & $\boldsymbol{91.7}$ & $\boldsymbol{3.44}$ & $\boldsymbol{4.00}$ \\
\midrule
\multirow{3}{*}{\shortstack[l]{Experts $\geq 2000$\\$(n{=}30)$}}
 & No RL           & $70.0$ & $3.03$ & $2.83$ \\
 & GPT-5.4 reward  & $56.7$ & $2.97$ & $2.37$ \\
 & Ours            & $\boldsymbol{93.3}$ & $\boldsymbol{3.17}$ & $\boldsymbol{3.70}$ \\
\bottomrule
\end{tabular}
\end{table}

\subsection{Compute Resources}\label{app:training_compute}

All training was performed on a single $8 \times $ A100-40GB node. The runs reported in the main paper consumed approximately $390$ A100-40GB GPU-hours in total: roughly $150$ for Stage-1 pooled training across the three domains, $20$ for Stage-2 per-student specialization, $170$ for the chess tutor SFT and the GRPO runs behind the proof of concept, and $50$ for evaluation and decoding (including prompted-baseline inference for GPT-4o and GPT-5.4). The full project, including ablations on subsets of students within a single domain (the modality VL sweep and the reward-variant comparisons) and preliminary or failed runs, used approximately $1{,}000$ A100-40GB GPU-hours. All GPU-hour figures here are rough estimates from the run wall-clocks rather than exact accounting.

\section{Scope and Design Rationale}\label{app:design_rationale}

This appendix records the design rationale behind several scope and operationalization choices in the framework: what behavioral fidelity and guidance responsiveness measure, how the multi-turn corpora are constructed, why the per-domain $\mathcal{F}$ instantiations differ, how per-student data scale interacts with the two-stage pipeline, which baselines we evaluate against in each domain, and what the tutor RL proof of concept claims. Each subsection states the design choice as a thesis, gives the constraint or principle that motivates it, and points to the section in the main paper where the choice is exercised.

\subsection{What Guidance Responsiveness Measures}\label{app:design_g_semantics}

Educational measurement has long treated assisted performance as informative about the learner in its own right. Vygotsky's zone of proximal development locates achievement under guidance alongside independent achievement \citep{VYGOTSKY_1980}, and dynamic assessment operationalizes that idea by evaluating how a learner responds after mediation \citep{Grigorenko_1998}. Our multi-turn record $(x, m, \tau, m^*)$ has exactly this structure: a problem $x$, the learner's initial response $m$, a tutor's mediating turn $\tau$, and the criterion response $m^*$ that the guidance is steering toward. In this framework, behavioral fidelity $\mathcal{F}$ captures the learner's independent starting point, while guidance responsiveness $\mathcal{R}$ captures the learner's assisted performance. Prior student simulators reproduce isolated responses; carrying an instructional exchange forward through answer, guidance, and revision is the capability an interactive tutor loop depends on.

This grounding also fixes the target of $\mathcal{R}$. Dynamic assessment scores post-mediation performance against a criterion response \citep{Grigorenko_1998}, because the construct of interest is whether the learner can move in the direction opened by the help. We therefore score whether the simulator emits the canonical resolved response $m^*$ on each record $(x, m, \tau, m^*)$ after reading the tutor guidance $\tau$. The target $m^*$ is well-defined per benchmark: the engine best move in chess, the correct option in mathematics, and the canonical written correction in second-language English writing. Section~\ref{sec:problem_formulation} gives the formal definition.

This criterion-based target is also consistent with theories of contingent tutoring and scaffolding, where effective support is adapted to the learner and then withdrawn as the learner takes over \citep{Wood_1976}, and with the assistance dilemma, which holds that the appropriate form of help depends on how much the learner can productively do with support \citep{Koedinger_2007}. In that setting, responsiveness concerns whether a learner advances under a given intervention, and a per-student continuation target would entangle $\mathcal{R}$ with the student-specific response patterns already measured by $\mathcal{F}$. Collecting such continuations for each intervention at benchmark scale is also a data-collection cost few settings can absorb. Substituting real per-student continuations in future datasets remains compatible with the $\mathcal{F}$/$\mathcal{R}$ decomposition.

This is also exactly the signal the tutor optimization loop needs. In \textsc{StudentSim}, the simulator serves as the reward model in tutor RL (Section~\ref{sec:tutor_rl}), so the tutor must receive feedback about whether its guidance moved the student toward the response that guidance was intended to elicit. $\mathcal{R}$ measures the rate at which the simulator supplies that feedback correctly. A simulator with high $\mathcal{R}$ produces a reward whose sign and magnitude track the direction of tutor intent, enabling policy optimization; a simulator with low $\mathcal{R}$ breaks that link, so optimization cannot recover the missing instructional signal.

Together, $\mathcal{F}$ and $\mathcal{R}$ supply the two properties a tutor RL loop needs from a simulator: the interaction begins from the actual error this student would make ($\mathcal{F}$), and then evolves in a way that reflects how that student responds to guidance ($\mathcal{R}$). The fuller longitudinal dynamics a learner exhibits across many interactions extend beyond these two axes and remain the horizon this decomposition opens toward (Section~\ref{sec:conclusion}). Section~\ref{sec:tutor_rl} shows that simulators strong on both axes provide a stronger tutor-RL reward than a frontier-LLM simulator or no RL on held-out chess positions, which is the operational test of this measurement choice.

\subsection{Multi-Turn Corpus Construction and the Simulator-Tutor Decoupling}\label{app:design_multiturn_corpus}

Multi-turn $(x, m, \tau, m^*)$ tuples do not exist at scale in two of our three domains, so constructing them with a parameterized LLM tutor is the principled way to instantiate the $(S_i, T_i)$ schema of Section~\ref{sec:problem_formulation} and to control the guidance-style distribution at training time. The Lichess export \citep{lichess} pairs each board position with the player's move and an engine evaluation, but carries no tutor commentary attached to the player's error. FoundationalASSIST \citep{worden2026foundationalassisteducationaldatasetfoundational} pairs each problem with a student answer and a correctness flag, but carries no per-error tutor explanation. A corpus that exposes problem, error, and canonical correction does not, by itself, expose the natural-language guidance turn $\tau$ that the schema needs.

L2 uses real-teacher annotations directly. EFCAMDAT \citep{geertzen2013automatic,Shatz_2020,Oksuz_2025} contains span-level corrections written by real teachers, and we use those annotations as the canonical $m^*$ on every L2 multi-turn record. The natural-language tutor turn $\tau$ is rendered from each annotation through one of two style templates, point-based and rule-based, so the surface phrasing is consistent while the underlying span and correction stay human-authored. The framework instantiates against whatever guidance source each domain exposes: real-teacher annotations on L2, controllable LLM authorship on chess and math.

For chess and math, $\tau$ is generated by an LLM tutor conditioned on the position or problem, the student error, and the canonical correction, under a fixed style template per record. This authorship is a control mechanism. The templates parameterize four guidance styles in chess (error remediation, comparative, strategic, Socratic) and three in math (error remediation, Socratic, conceptual explanation), mixed uniformly at training time. The style set spans the range of instructional support a tutor chooses among, from stating the correction outright to prompting the student to derive it, the range over which effective tutoring adjusts help to the learner \citep{Wood_1976} and along which the assistance dilemma is posed \citep{Koedinger_2007}. That control is what enables the per-mode $\mathcal{R}$ breakdown in Section~\ref{sec:guidance_results} and the held-out Socratic case study, in which the simulator solves a position whose Socratic prompt it never saw at training time. A fixed corpus of opportunistic tutor traces would not afford a balanced per-style population.

\paragraph{Guidance Quality.} For every record $(x, m, \tau, m^*)$, the target $m^*$ is fixed by verified supervision before any tutor turn is written, and $\tau$ is generated from the analysis that establishes that target. In chess, $m^*$ is the Stockfish depth-$15$ best move, and the generator is given the engine's full position analysis: the ranked candidate moves with their centipawn evaluations, and the principal continuation of each candidate. The continuations are what let the guidance explain how the game unfolds after each choice. In math, $m^*$ is the platform answer key after an audit (Appendix~\ref{app:benchmark_details}), and the generator receives that audited key together with the student's recorded answer. In L2 writing, $m^*$ is the human teacher's span correction, and $\tau$ is templated directly from that annotation with no LLM generation step. Across all three domains, the LLM determines only the wording of the guidance; the endpoint it steers toward is set in advance by an engine or a human annotation and is the same supervision used to compute $\mathcal{R}$.

The generator, the simulator, and the prompted closed-source baseline are three separate systems. The LLM that authors $\tau$ during corpus construction is GPT-4o or GPT-5.1 in chess, selected per record by whether the student's move fell outside the engine's top-$k$ list (contrasting challenging positions and easy ones), and GPT-5.4 in math. The simulator is Qwen3-4B-Instruct \citep{yang2025qwen3technicalreport} with LoRA adapters; it never produces $\tau$ and only consumes $\tau$ to emit $m^*$. At evaluation the closed-source baseline is GPT-5.4, prompted under the same $\tau$ that the simulator reads, so on math the baseline shares a model family with the generator, an arrangement that works in the baseline's favor. Our simulator never authors $\tau$ at any point; it only consumes it.

\subsection{Per-Domain Operationalization of Behavioral Fidelity}\label{app:design_f_per_domain}

$\mathcal{F}$ is a generic concept: how closely the simulator's response on a held-out problem matches student $\pi_i$'s recorded response. The operational form is per-domain because the response space is per-domain. Chess $\mathcal{F}_i$ is top-1 accuracy against the recorded move. L2 $\mathcal{F}_i$ is per-record error-density and issue-type-profile match across seven LanguageTool categories between the simulator's generated essay and the recorded essay. Math $\mathcal{F}_i$ is top-1 accuracy on a four-way multiple-choice form: the target letter holds the student's recorded answer (correct or wrong) and the three distractor letters hold the most common other-student answers on the same problem. Appendix~\ref{app:metric_definitions} gives the full formulas.

Each operationalization is the well-posed match metric on its response space. Chess legal-move sets are finite (typically about thirty moves) and the recorded supervision is one chosen move per position; top-1 against that move is the only fidelity signal one observation can support. L2 essays are free-form text, and no two essays the same learner writes are identical at the surface; what survives as a per-learner signal is the rate and type-profile of mistakes the learner still makes, which is what per-record error-profile match captures. Math student answers are open-ended free-text strings, and raw-text likelihoods on them are incomparable across simulators because they depend on tokenization of the numerical answer; casting each problem into a four-way form with empirical other-student distractors turns the response space into a categorical one, on which top-1 accuracy is the natural fidelity score.

Cross-domain absolute comparison of $\mathcal{F}$ values is not the framework's unit of analysis. What we report per domain is the relative improvement of \textsc{StudentSim} over the strongest available baseline (\cref{tab:fidelity_main}, Section~\ref{sec:fidelity_results}), which is comparable across domains because every baseline in a domain is scored on the same per-domain $\mathcal{F}$. This is the convention that supports BLEU on machine translation alongside ROUGE on summarization: both rank systems within a benchmark, and neither gains from being coerced into the other's units.

A unified $\mathcal{F}$ is not the right design. Coercing all responses into a shared embedding space loses the domain-specific signal: chess top-1 uses a discrete legal-move list, L2 profiles a fixed seven-category taxonomy, math letter accuracy a four-way categorical decision. Reporting only relative improvement over a population prior discards the absolute scale that makes per-student claims interpretable within a domain. Reporting per-domain $\mathcal{F}$ and per-domain relative improvement together keeps both an absolute interpretation within each domain and a clean comparison across baselines.

\subsection{Per-Student Data Scale and the Population Prior}\label{app:design_per_student_scale}

Per-student data is bounded by the number of interactions any one learner actually generates, and obtaining more of them takes as long as that learner takes to produce them, which is why per-user sparsity recurs in every system that learns from interactions with real users \citep{Schein_2002, lee2019melumetalearneduserpreference, bhattacharjee2025coldstartproblemexperimental, li2017usersimulatortaskcompletiondialogues}. Training a simulator end to end on one student's records would overfit the handful that exist and would learn the domain's shared regularities over again for every student \citep{dodge2020finetuningpretrainedlanguagemodels, finn2017modelagnosticmetalearningfastadaptation}. The framework instead uses the data that exists at two levels of granularity: pooled domain records for what students share, and individual records to adapt that base to a particular learner.

The two-stage pipeline in Section~\ref{sec:method} turns that constraint into a training recipe. Stage 1 pools records across students within a domain, producing a population prior over the response space, the mistakes students commonly make, and the ways students revise after guidance. Stage 2 continues from that base on one student's own records, so each additional personal instance gives the simulator more evidence about that learner's characteristic behavior. The design therefore scales with whatever evidence a platform holds: the pooled stage supplies broad competence from the full domain, as supported by the matched chess ablation in Appendix~\ref{app:ablation_cross_student_pooling}, and the specialization stage sharpens the per-student fit as personal data accrues. A deployment needs only those two ingredients, and for online learning platforms that want to train student simulators to help each student grow can supply both, since a large population generates the pooled corpus while each learner contributes a handful of records.

A learner with no recorded interactions occupies the pooled-prior end of this range. The simulator is then the Stage-1 domain model. Once interactions are observed, Stage 2 specializes that prior toward the individual learner, so the framework needs no separate cold-start mechanism.

\subsection{Domain-Specific Baselines and the Strongest Available Reference}\label{app:design_baselines}

For each of the three domains, we evaluate against the strongest baseline that exists in that domain's literature. The asymmetry across domains in our baseline set reflects a real asymmetry in what per-student behavior models the respective communities have published. In brief, chess has a community-trained per-player move predictor that we can use directly; L2 and math do not, and in those domains the closed-source LLM with the student profile in context is the strongest available reference.

Chess has Maia2 \citep{tang2024maia2unifiedmodelhumanai}, which builds on Maia \citep{mcilroyyoung2020aligningsuperhumanaihuman}: a chess move-prediction model conditioned on player rating, trained on millions of recorded human games. Two domain conditions enable Maia2's existence: a fixed, narrow response space (legal moves), and an active community that has trained dedicated player-behavior engines. We use Maia2 directly as the chess state-tracking baseline, alongside the closed-source LLM baselines (GPT-4o and GPT-5.4) that are common to all three domains.

For L2 and math, no published model plays the analogous role of Maia2. Knowledge tracing exists for math problem traces in principle, and earlier work has applied BKT \citep{Corbett_1995}, DKT \citep{piech2015deepknowledgetracing}, and AKT \citep{ghosh2020contextawareattentiveknowledgetracing} to predict per-step correctness; however, those models predict binary correctness on the next problem rather than which specific (often wrong) free-text answer a particular student would emit, which is the fidelity target our framework operationalizes. They are therefore not a drop-in baseline for our $\mathcal{F}$. For L2 essay generation conditioned on per-learner history, no published artifact exists at all. In both domains, the closed-source LLM with the student profile in context is the strongest available reference, and that is what we evaluate against in Section~\ref{sec:fidelity_results} and Section~\ref{sec:guidance_results}.

\paragraph{Closed-source LLM prompt construction.}
The closed-source LLM baselines receive the same in-context task inputs as the trained simulator: the per-student profile block, the problem (a FEN position in chess, an essay prompt in L2, a multiple-choice problem in math), and, for the guidance-responsiveness records, the student's initial response followed by the tutor's guidance turn. In L2 and math the closed-source prompt is character-for-character identical to the trained simulator's. In chess it agrees on every profile statistic, the FEN presentation, the color-to-move line, and the guidance turn, with three deliberate differences. First, the closed-source prompt omits the ``likely next moves'' line, a Maia2-derived per-move probability list that appears in the trained simulator's chess profile; we hold this line out so the closed-source baseline is scored on the same textual profile the two share and not on Maia2's move distribution, which Maia2 already supplies as a separate baseline. Second, a short system message instructs the model to emit only a UCI move. Third, the move-request suffix asks for a bare UCI move with no accompanying reasoning. The second and third differences normalize the closed-source model's output so that its move is parseable and comparable to the trained simulator's greedy-decoded move; they carry no additional task information. The closed-source baseline therefore receives the same description of the student and the board position as the trained simulator, and Maia2's move predictions enter the comparison separately, through the trained simulator's profile line and through Maia2's own baseline.

The framework itself is benchmark-agnostic. The $(S_i, T_i)$ schema and the $\mathcal{F}$ and $\mathcal{R}$ metrics do not depend on which baselines are present in the comparison set. Any future per-domain student-behavior model published in the L2 or math literature can be added to \cref{tab:fidelity_main} and \cref{tab:guidance_main} without modifying the metrics or the held-out evaluation protocol. The framework is therefore designed to absorb stronger baselines as they become available.

\subsection{Validation Scope of the Tutor RL Proof of Concept}\label{app:design_poc_scope}

The Section~\ref{sec:tutor_rl} comparison is constructed so that the reported tutor quality is evaluated independently of the GRPO reward used during training. We assess the trained tutors through an expert human study in which competitive chess players rate blind, order-shuffled tutor responses on accuracy, guidance quality, and personalization. Evaluators do not see which system produced a response, and they never see the reward definition and the student simulator used during tutor training. This isolates the question of interest: whether a trained \textsc{StudentSim}, used as the tutor-RL reward, produces a better tutor than a frontier-LLM-as-student-simulator reward and than no reinforcement learning.

The study is reported in three views: all annotations, the triply-annotated subset, and the subset of evaluators rated $2000$ or above. These views test whether the ordering holds under stricter annotation agreement and stronger evaluator expertise, and all of them rest on judgments that are decoupled from the optimization signal that trained the tutors.

The proof of concept is scoped to simulator-informed AI tutor optimization, where the question is whether feedback from a trained student simulator is informative enough to improve tutor responses under a controlled comparison. Section~\ref{sec:tutor_rl} answers that by training tutors with different reward sources and evaluating their outputs with independent expert judgments. Real-student learning outcomes are the target of a deployment study, whose endpoint is student improvement over time under live use, and the simulator-feedback result established here is what motivates that next stage.

\paragraph{Why chess is the right scope for the proof of concept.}
Chess supplies a precise per-position notion of move quality through Stockfish centipawn evaluation at fixed depth, a signal that is independent of the simulator and of the tutor's wording and is tied directly to the game state. That makes chess a clean setting in which to run tutor RL against a well-specified reward while keeping a separate human evaluation of the final tutor outputs. Instantiating the same proof of concept in L2 writing or open-ended mathematics would first require per-domain reward functions over free-form student responses, such as reliable essay-quality scoring or robust verification of mathematical reasoning, each an open research program in its own right. Whether simulator feedback can improve a tutor is a question about the framework, and chess is the clearest domain in which to answer it under an evaluation protocol that stays independent of the training reward.

\end{document}